%% file: main_arxiv.tex
\documentclass[journal]{IEEEtran}

\usepackage[pdftex]{graphicx}

\usepackage{amsmath}
\usepackage{algorithmic}

\usepackage{times}
\usepackage{epsfig}
\usepackage{amssymb}
\usepackage{nicefrac}
\usepackage[ruled,vlined]{algorithm2e}
\usepackage{booktabs} %
\usepackage{overpic}
\usepackage{caption}
\usepackage{xcolor}
\usepackage{amsthm}
\usepackage{tikz}

\usepackage[noadjust]{cite}

\definecolor{red1}{rgb}{0.8431    0.1882    0.1529}
\definecolor{orange1}{rgb}{0.9882    0.5529    0.3490}
\definecolor{yellow1}{rgb}{0.9961    0.8784    0.5451}
\definecolor{blue1}{rgb}{0.3000    0.3000    0.8490}
\definecolor{green2}{rgb}{0.2    0.8    0.3}
\definecolor{green1}{rgb}{0.1020    0.5961    0.3137}
\definecolor{red2}{rgb}{0.8431    0.1882    0.1529}

\newcommand{\blue}[1]{{\color{blue}#1}}
\newcommand{\red}[1]{{\color{red}#1}}

\definecolor{aquamarine}{rgb}{0.5, 1.0, 0.83}
\definecolor{amber}{rgb}{1.0, 0.75, 0.0}
\definecolor{arsenic}{rgb}{0.15, 0.17, 0.19} %
\definecolor{coralred}{rgb}{1.0, 0.25, 0.25}

\input{md_math}

\usepackage[pagebackref=true,breaklinks=true,colorlinks,bookmarks=false]{hyperref}

\begin{document}
\title{Polyblur: Removing mild blur\\ by polynomial reblurring}

\author{%
Mauricio~Delbracio, %
Ignacio Garcia-Dorado,
Sungjoon Choi, 
Damien Kelly,
Peyman Milanfar\\[0.3em]
Google Research
}

\maketitle

\begin{abstract}
We present a highly efficient blind restoration method to remove mild blur in natural images.  Contrary to the mainstream, we focus on removing slight blur that is often present, damaging image quality and commonly generated by small out-of-focus, lens blur, or slight camera motion. The proposed algorithm first estimates image blur and then compensates for it by  combining multiple applications of the estimated blur in a principled way. 
To estimate blur we introduce a simple yet robust algorithm based on empirical observations about the distribution of the gradient in sharp natural images. Our experiments show that, in the context of mild blur, the proposed method outperforms traditional and modern blind deblurring methods and runs in a fraction of the time.  Our method can be used to blindly correct blur before applying off-the-shelf deep super-resolution methods leading to superior results than other highly complex and computationally demanding techniques. The proposed method estimates and removes mild blur from a 12MP image on a modern mobile phone in a fraction of a second.\end{abstract}


\begin{figure}
    \centering
    \footnotesize
     \begin{minipage}[c]{\linewidth}
     \begin{minipage}[c]{.35\linewidth}
     \begin{tikzpicture}
      \node[anchor=north west,inner sep=0] (image) at (0,0) {\includegraphics[clip,trim=160 0 700 20, width=\linewidth]{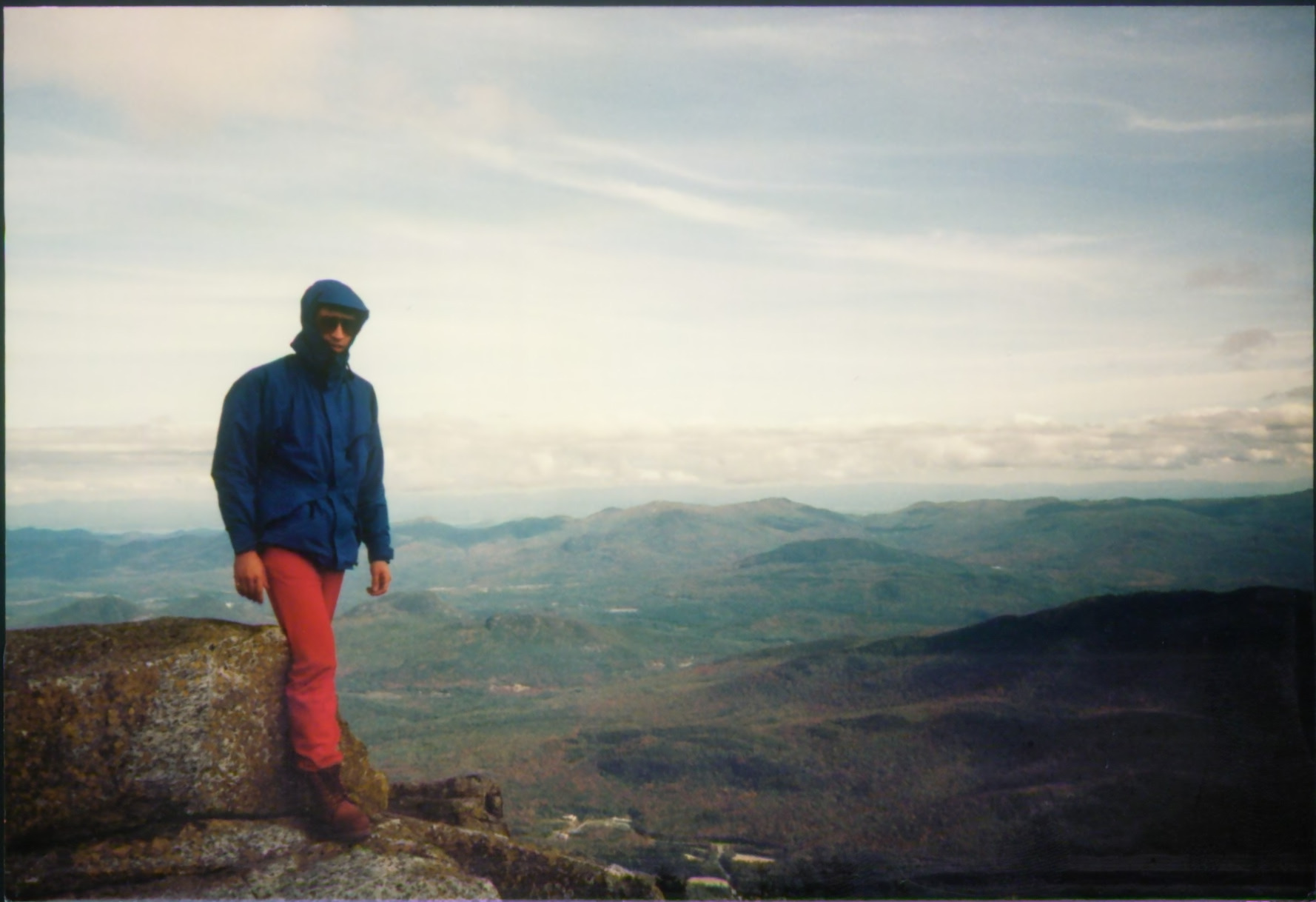} };
      \begin{scope}[x={(image.north east)},y={(image.south west)}]
        \draw[yellow1, very thick] (0.2166, 0.8114) rectangle (0.5097, 0.9716);
        \draw[coralred, very thick] (0.1487, 0.2888) rectangle (0.4418, 0.4491);
      \end{scope}
     \end{tikzpicture}
     \end{minipage}
     \begin{minipage}[c]{.65\linewidth}
     \begin{minipage}[c]{.494\linewidth}
     \begin{center}
          \includegraphics[width=\linewidth]{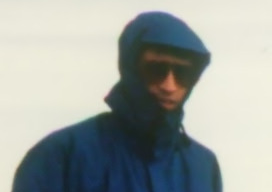}\vspace{.15em}
          \includegraphics[width=\linewidth]{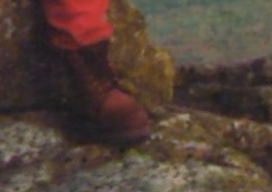} 
    \end{center}
     \end{minipage}\hspace{-.2em}
     \begin{minipage}[c]{.494\linewidth}
      \begin{center}
          \includegraphics[width=\linewidth]{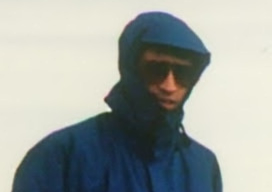}\vspace{.15em}
          \includegraphics[width=\linewidth]{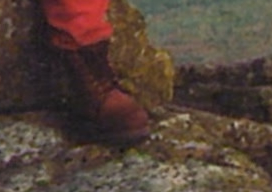}
    \end{center}
     \end{minipage}
     \end{minipage}     
     \end{minipage}
     \vspace{.2em}
     
     \begin{minipage}[c]{\linewidth}
     \begin{minipage}[c]{.35\linewidth}
     
     \begin{tikzpicture}
      \node[anchor=north west,inner sep=0] (image) at (0,0) {\includegraphics[clip,trim=150 0 680 60, width=\linewidth]{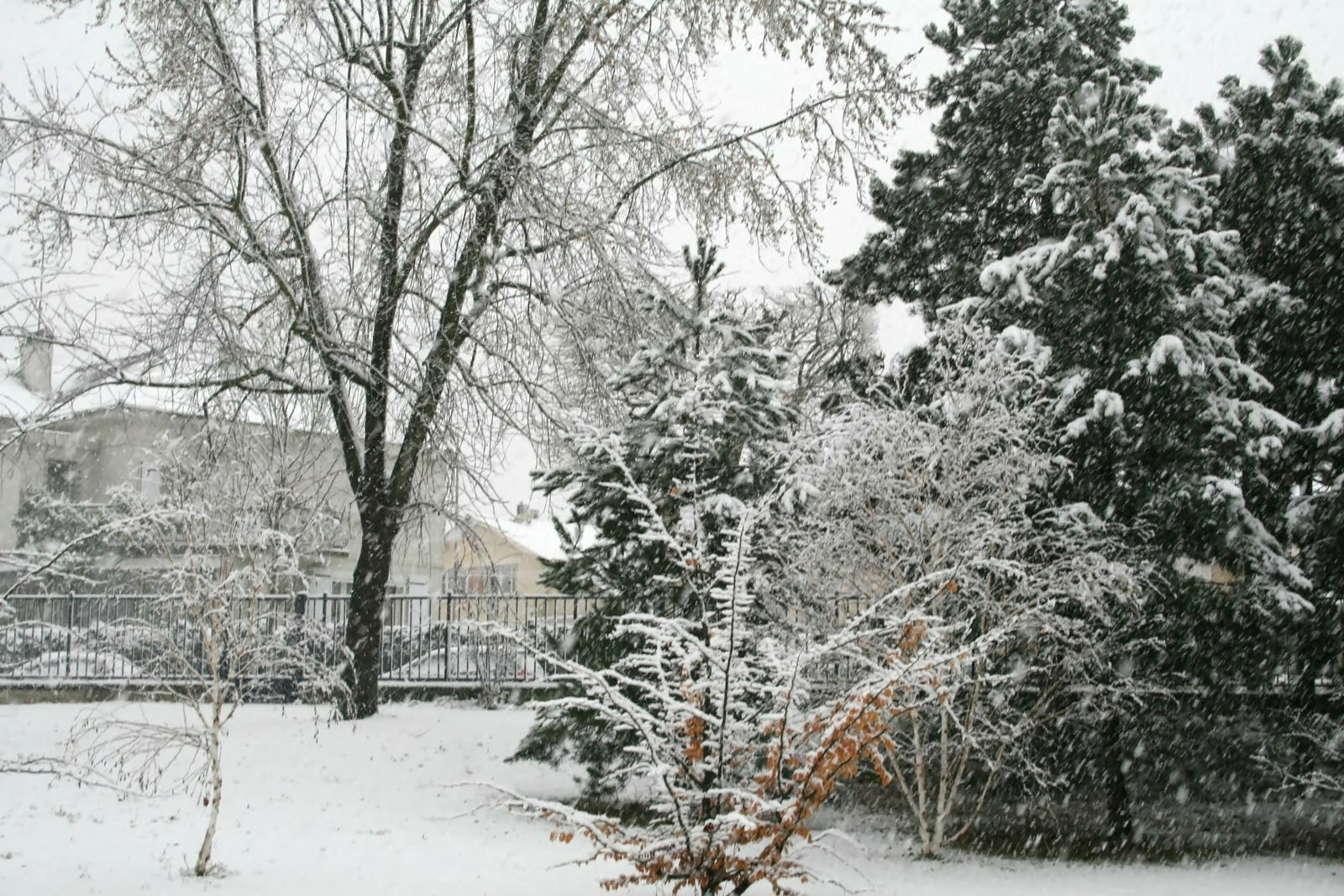} };
      \begin{scope}[x={(image.north east)},y={(image.south west)}]
        \draw[orange1, very thick] (0.6714, 0.4449) rectangle (0.8481, 0.5402);
        \draw[coralred, very thick] (0.1519, 0.1470) rectangle (0.3286, 0.2423);
     \end{scope}
     \end{tikzpicture}
     \end{minipage}
     \begin{minipage}[c]{.65\linewidth}
     \begin{minipage}[c]{.494\linewidth}
     \begin{center}
          \includegraphics[width=\linewidth]{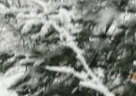}\vspace{.15em}
          \includegraphics[width=\linewidth]{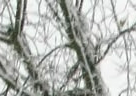} 
         \end{center}
     \end{minipage}\hspace{-.2em}
          \begin{minipage}[c]{.494\linewidth}%
          \begin{center}
          \includegraphics[width=\linewidth]{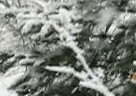}\vspace{.15em} 
          \includegraphics[width=\linewidth]{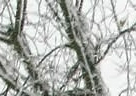} 
         \end{center}
     \end{minipage}
     \end{minipage}     
     \end{minipage}

     \vspace{.2em}

     \begin{minipage}[c]{\linewidth}
     \begin{minipage}[c]{.35\linewidth}
     \centering
      \begin{tikzpicture}
      \node[anchor=north west,inner sep=0] (image) at (0,0) {\includegraphics[clip,trim=2200 350 1000 700, width=\linewidth]{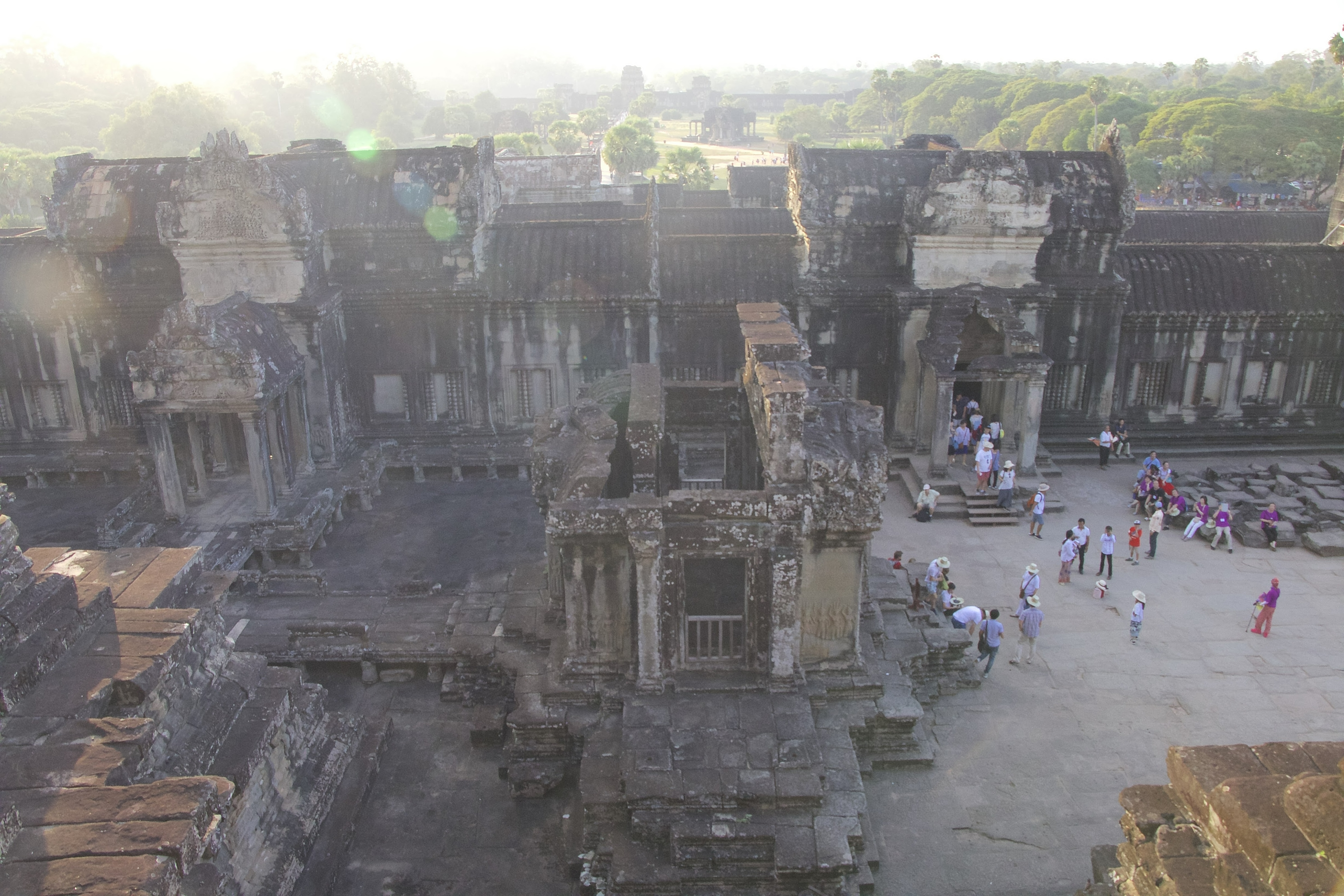}};
      \begin{scope}[x={(image.north east)},y={(image.south west)}]
      \draw[coralred, very thick] (0.7138, 0.6658) rectangle (0.8740, 0.7506);
      \draw[yellow1, very thick] (0.7845, 0.3470) rectangle (0.9446, 0.4318);
      \end{scope}
     \end{tikzpicture}
     
     blurry input
     \end{minipage}
     \begin{minipage}[c]{.65\linewidth}
     \begin{minipage}[c]{.494\linewidth}
       \centering
       \includegraphics[width=\linewidth]{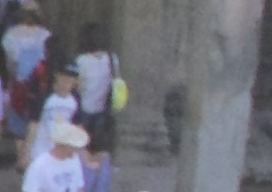}\vspace{.15em} 
       \includegraphics[width=\linewidth]{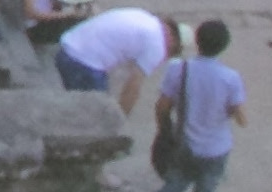} 
          
       blurry crop
      \end{minipage}\hspace{-.2em}
       \begin{minipage}[c]{.494\linewidth}
       \centering
        \includegraphics[width=\linewidth]{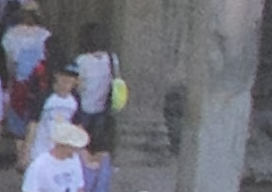}\vspace{.15em} 
        \includegraphics[width=\linewidth]{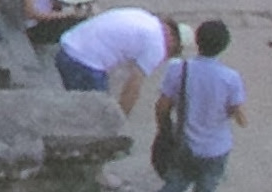} 
          
        polyblur
     \end{minipage}
     \end{minipage}     
     \end{minipage}

   \caption{Mild blur, as shown in these examples, can be efficiently (42ms/MP on a mid-range smartphone CPU) removed by combining multiple applications of the estimated blur. Readers are encouraged to zoom-in for better visualization.}
\label{fig:teaser}
\end{figure}
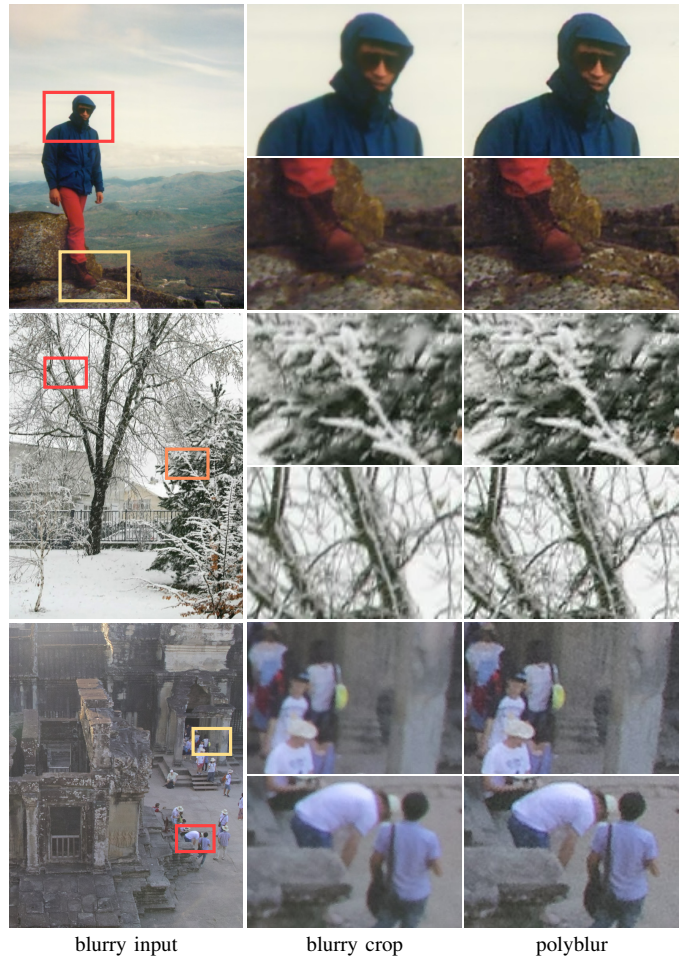

\IEEEpeerreviewmaketitle

\section{Introduction}
\label{sec:introduction}

\begin{figure*}
    \centering
    \scriptsize
    \begin{minipage}[c]{.196\linewidth}
    \centering
    \includegraphics[width=\linewidth]{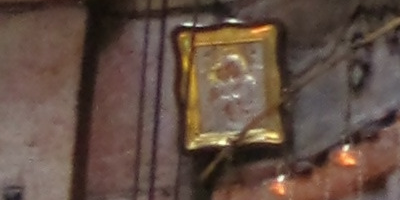}
    $u_0$ - input image
    \end{minipage}
    \begin{minipage}[c]{.196\linewidth}
    \centering
    \includegraphics[width=\linewidth]{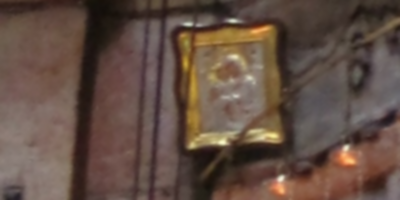}
    $u_1 = u_0 \ast k$ 
    \end{minipage}
    \begin{minipage}[c]{.196\linewidth}
    \centering
    \includegraphics[width=\linewidth]{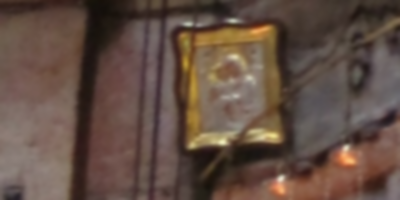}
    $u_2 = u_1 \ast k$ 
    \end{minipage}   
    \begin{minipage}[c]{.196\linewidth}
    \centering
    \includegraphics[width=\linewidth]{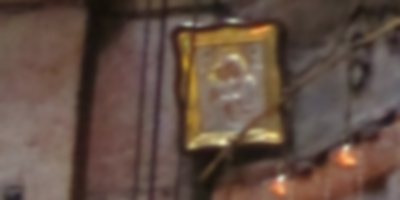}
    $u_3 = u_2\ast k $ 
    \end{minipage}   
    \begin{minipage}[c]{.196\linewidth}
    \centering
    \includegraphics[width=\linewidth]{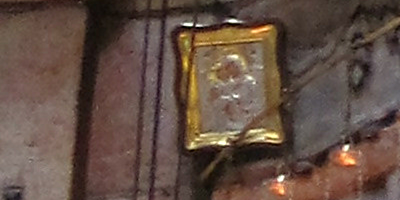}
    $u_F = \sum_i a_i u_i$
    \end{minipage}  
    \vspace{-0.5em}
    
      \caption{The deblurred image is generated by combining multiple re-applications of the estimated blur (polyblur).}
    \label{fig:polyblur-example}
\end{figure*}

\IEEEPARstart{I}{mage} sharpness is undoubtedly one of the most relevant attributes defining the visual quality of a photograph. Blur is caused by numerous factors such as the camera’s focus not being correctly adjusted, objects appearing at different depths, or when relative motion between the camera and the scene occurs during exposure. Even in perfect conditions, there are unavoidable physical limitations that introduce blur. Light diffraction due to the finite lens aperture, integration of the light in the sensor and other possible lens aberrations introduce blur leading to a loss of details. Additionally, other components of the image processing pipeline itself, particularly demosaicing and denoising, can introduce blur. 

Removing blur from images is a longstanding problem in image processing and computational photography spanning more than 50 years~\cite{kovasznay1955image,kundur1996blind,lai2016comparative}. Progress has been clear and sustained. From image enhancement algorithms~\cite{he2010guided,talebi2016fast}, blind and non-blind deconvolution methods~\cite{fergus2006removing,levin2009understanding,perrone2014total,shan2008high} where sophisticated priors are combined with optimization schemes, to very recent years with the incipient application of deep neural models
~\cite{gao2019dynamic,kupyn2018deblurgan,kupyn2019deblurgan,su2017deep, tao2018scale,zhang2020deblurring,wieschollek2017learning}.

Many of these methods are capable of processing significantly degraded images revealing previously unseen image details. The gain in quality is in many cases extraordinary, but impractical.  These methods make extensive use of prior information (learned or modeled) producing images that are often unrealistic. This is mainly because the inverse problem they tackle is seriously ill-posed.

In this work, we detach ourselves from the current trend and focus on the particular case where the blur in the image is small. As we show in our experimental results, for this case, the vast majority of existing methods generate notorious image artifacts in addition to requiring great computational power. This is mainly because most of the algorithms were not specifically designed for this very common and practical use case.
We introduce a highly efficient blind image restoration method that removes mild blur in natural images. The proposed algorithm first estimates image blur and then compensates for it by combining multiple applications of the estimated blur in a principled way (Figure~\ref{fig:polyblur-example}). The method is inspired by the fact that the inverse of an operator that is close to the identity (e.g., mild blur), can be well approximated by means of a low-degree polynomial on the operator itself. We design polynomial filters that use the estimated blur as a base and approximate the inverse without neglecting that image noise can get amplified. 

The removal of blur commonly leads to the introduction of halos (oversharpening) mostly noticeable near image edges. To address this, we present a mathematical characterization of halos and propose a blending mechanism to render an artifact-free final image. This step, which is also highly efficient, is important to achieve consistent high quality.

Experiments with both real and synthetic data show that, in the context of mild blur in natural images, our proposed method outperforms traditional and modern blind deblurring methods and runs in a fraction of the time.  
The simplicity of the polynomial filter, together with the choice of the Gaussian function as blur the model, enables a highly efficient implementation. The method runs at interactive rates so it can be used on mobile devices. Polyblur can estimate and remove mild blur on a 12MP image on a modern mobile phone in a fraction of a second. 
Additionally, we show that Polyblur can be used to blindly correct blur on an image before applying an off-the-shelf deep super-resolution method. Learning-based single image super-resolution algorithms (SISR) are  usually trained on images where the degradation operator has been modeled in an over-simplistic way (typically by bilinear or bicubic blur). Polyblur in combination with a standard deep SISR network leads to superior results than other highly complex and computationally demanding methods. \vspace{.3em}

\noindent \textbf{Contributions.} We introduce a novel method to estimate and remove mild blur (very common in mobile photography) that is (i) highly efficient and simple, (ii) theoretically sound, and (iii) produces competitive or better results while being orders of magnitude faster.  Additionally, our method can blindly correct blur before applying off-the-shelf deep super-resolution methods leading to superior results vs. highly complex and computationally demanding techniques. Our algorithm's components are built in a principled way: from explicit, empirically verified assumptions in the blur estimation, to mathematical models in the polynomial deblurring and halo removal. Each component is designed to maximize efficiency (12MP image processed in milliseconds on a mobile CPU).

The remainder of the paper is organized as follows. In Section~\ref{sec:related_work} we discuss the closely related work, while in
Section~\ref{sec:polynomial_filters} we present the polynomial approximation of the blur inverse and the adopted polynomial deblurring family. Section~\ref{sec:model} introduces the anisotropic Gaussian blur model and the necessary tools that allow an efficient estimation of the parameters. Section~\ref{sec:implementation_details} summarizes the main components of the proposed blind estimation and removal algorithm, and experimental results on real and simulated data are shown in Section~\ref{sec:experiments}. Conclusions are finally summarized in Section~\ref{sec:conclusion}.

\section{Related Work}
\label{sec:related_work}
Image blur is generally modeled as a linear operator acting on a sharp latent image. If the blur is shift-invariant then the blurring operation amounts to a convolution with a blur kernel.  This implies,
\begin{equation}
v = k \ast u + n,
\end{equation}
where $v$ is the captured image, $u$ the underlying sharp image, $k$ the unknown blur kernel and $n$ is additive noise. In what follows we review the different families of methods that remove image blur. \vspace{.5em}

\noindent \textbf{Blind deconvolution and variational optimization.}
The typical approach to image deblurring is by formulating the problem as one of blind deconvolution where the goal is to recover the sharp image $u$ without knowing the blur kernel $k$~\cite{chan1998total,kundur1996blind,lai2016comparative}. Most blind deconvolution methods run on two steps: a blur kernel $k$ is first estimated and then a non-blind deconvolution technique is applied~\cite{cho2009fast,fergus2006removing,pan2014deblurring,pan2017deblurring,xu2010two,xu2013unnatural}. 
Fergus et al.~\cite{fergus2006removing} is one of the best examples of the blind deconvolution variational family that seeks to combine image priors, assumptions on the blurring operator, and optimization frameworks, to estimate both the blurring kernel and the sharp image~\cite{cai2009blind,chen2019blind,jin2018normalized,krishnan2011blind,michaeli2014blind,perrone2014total,shan2008high,xu2013unnatural}.
Estimating the blur kernel is easier than jointly estimating the kernel and the sharp image together. Levin et al.~\cite{levin2009understanding,levin2011efficient} show that it is better to first solve a maximum a posteriori estimation of the kernel than the latent image and the kernel simultaneously. Notwithstanding, even in non-blind deblurring, the significant attenuation of the image spectrum by the blur and model imperfections, lead to an ill-posed inverse problem~\cite{anger2018modeling,krishnan2009fast}. \vspace{.5em}

\noindent \textbf{Sharpening methods.}
Sharpening methods aim to reduce mild blur and increase overall image contrast by locally modifying the image. Unsharp masking, arguably the most popular sharpening algorithm is very sensitive to noise and generally leads to oversharpening artifacts~\cite{kim2005optimal,polesel2000image}. 
Zhu and Milanfar~\cite{zhu2011restoration} propose an adaptive sharpening method that uses the local structure and a local sharpness measure to address noise reduction and sharpening simultaneously. Bilateral filtering~\cite{tomasi1998bilateral}, and its many variants/adaptations~\cite{buades2005non,he2010guided,milanfar2012tour,talebi2016fast} can be used to enhance the local contrast and high frequency details of an image. The overall idea is to proceed similarly to that of unsharp masking: the input image is filtered and then a proportion of the residual image is added back to the original input. This procedure boosts high-frequencies that are removed by the adaptive filter.
Since sharpening methods do not explicitly estimate image blur, their deblurring performance is limited. In fact, sharpening algorithms are strictly local operators and do not have access to additional information from outside the local neighborhood. Whereas deblurring algorithms like ours use the estimation of blur, which in general is global or done on a larger neighborhood, to remove the blur. \vspace{.5em}

\noindent \textbf{Deblurring meets deep-learning.}
In the past five years, with the popularization of deep convolutional networks several image deblurring methods have been introduced~\cite{chakrabarti2016neural,gao2019dynamic,kupyn2018deblurgan,kupyn2019deblurgan,su2017deep,sun2015learning,tao2018scale,wieschollek2017learning,wieschollek2016end,zhang2020deblurring}.
Most of these methods target strong motion blur and are in general trained with large datasets with realistically synthesized image blur. Efficient deblurring using deep models is a very challenging task~\cite{nah2020ntire}.

Blind deconvolution methods do a remarkable job when the image is seriously damaged and manage to enhance very low quality images. However, their performance in the particular case where blur is mild is limited since they introduce, in general, noticeable artifacts and have a significant computational cost. At the other extreme, adaptive sharpening methods do a fine job boosting image contrast when the input image has very little blur. Since they do not incorporate any explicit measure of the blur, their deblurring capabilities are restricted. Deep learning based methods require high computational cost, making them impractical in many contexts.  This work focuses on the problem of estimating and removing slight blur. As we show in the experimental part, in this setting, the proposed method produces superior results to blind deconvolution techniques and other popular adaptive sharpening algorithms, and similar results vs. advanced deep networks while being significantly more efficient.

\section{Deblurring with polynomial filters}
\label{sec:polynomial_filters}

\subsection{Approximating the blur inverse}
Let us first assume that we have an estimation of the blur kernel $k$. To recover the image $u$ we need to solve an inverse (deconvolution) problem. If the blur kernel is very large, the inversion is significantly ill-posed and some form of image prior will be required. In this work, we assume that the image blur is mild (see examples on Figure~\ref{fig:teaser}) so, as we will show in the experimental section, there is no need to incorporate any sophisticated image prior. Instead we proceed by building a linear restoration filter constructed from the estimated blur. 
An interesting observation~\cite{kaiser1977sharpening,milanfar2018rendition,tao2017zero} is that by carefully combining different iterations of the blur operator on the blurry image we can approximate the inverse of the blur. 

\noindent One way to illustrate this is as follows. %
\begin{lemma}
\label{lm:neumann}
Let $K$ be the convolution operator with the blur kernel $k$ and $I$ the identity operator,  then if $\|I-K\| < 1$ under some matrix norm, $K^{-1}$ the inverse of $K$ exists and,
\begin{equation}
K^{-1} = \sum_{i=0}^\infty (I-K)^i.
\label{eq:inv_power_series}
\end{equation}
\end{lemma}
The proof is a direct consequence of the convergence of the geometric series. If we assume circular boundary conditions for the convolution, then the eigenvectors of the matrix $K$ are the Fourier modes of $k$. This implies that $(I-K)^i = F^H (I-D)^i F$, where $F$ is the Fourier basis, $F^H$ is the Hermitian transpose of $F$, and $D$ is a diagonal matrix with the eigenvalues of $K$.

In particular, the series converges if the kernel Fourier coefficients $\hat{k}(\zeta)$ satisfy $\left|\hat{k}(\zeta)-1\right| < 1$.
In the case of blur filters that conserve the overall luminosity, a reasonable hypothesis is that $k(\bx) \ge 0$, and $\int k(\bx) d\bx = 1$. This implies that  $|\hat{k}(\zeta)| \le 1$ which is not enough to guarantee convergence.\vspace{.5em}

\noindent \textbf{Polynomial filters.}
According to Lemma~\ref{lm:neumann} we can approximate the inverse with a polynomial filter using the blur kernel as a base. For instance, if we truncate the power series and keep up to order 3, the polynomial approximate inverse of $K$ is,
\begin{equation}
K^{\text{inv}}_3 = 4I -6K + 4K^2 -K^3.
\end{equation}
This motivates the use of more general polynomials
\begin{equation}
p(K) = \sum_{i=0}^d a_i K^i,
\end{equation}
where the order $d$ and coefficients $(a_0,\ldots,a_d)$ can be designed to amplify or attenuate differently depending on how the blur is affecting a particular component. \vspace{.5em}

\begin{figure}
    \centering
    \includegraphics[width=\linewidth]{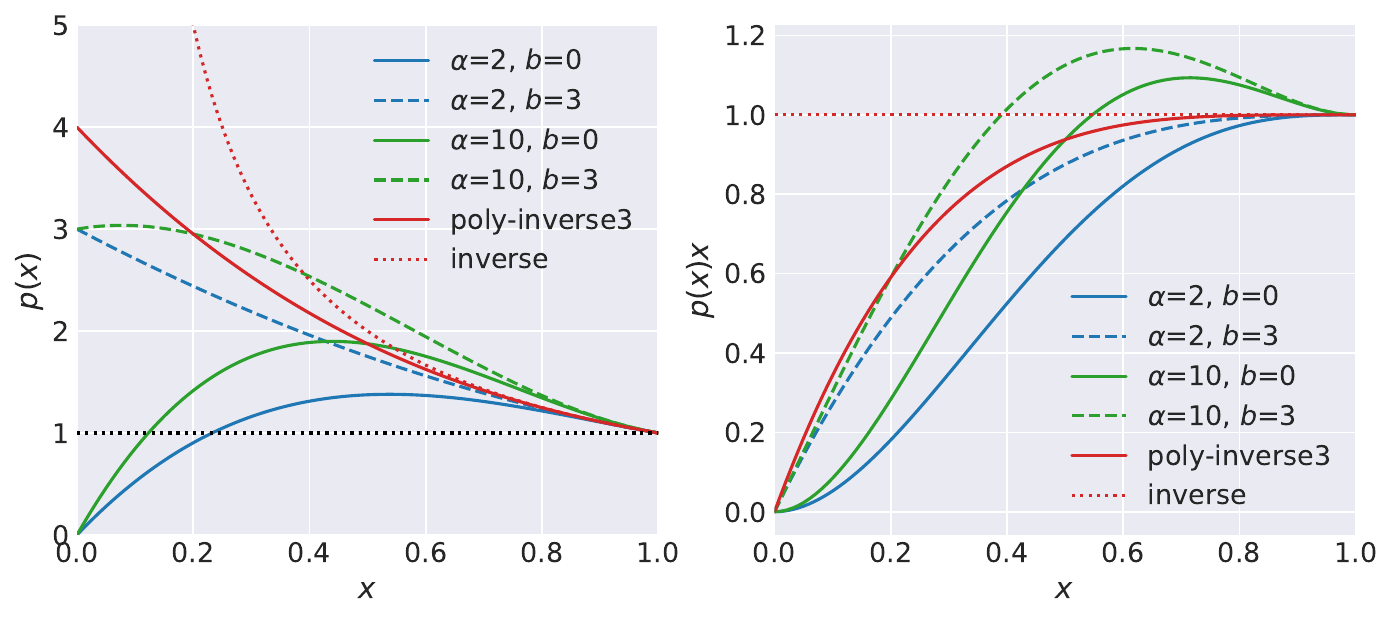}
    \caption{Polynomial filters can be analyzed directly on the \textit{Real} line. We present an example of the proposed $p_{3,\alpha,b}$ family.}
    \label{fig:polynomial-vis}
\end{figure}

\noindent \textbf{Symmetric filters with non-negative Fourier coefficients.}
Let us first assume that the blur filter $k(\bx)$ is symmetric $k(\bx)=k(-\bx)$, and has non-negative Fourier coefficients. We will later show how we can generalize the approach to any filter. In this setting, $\hat{k}(\zeta)$ the Fourier coefficients of the filter $k$ are in $[0,1]$.
The Fourier coefficients of the polynomial filter are related with the ones from the base filter through the same polynomial, that is,
\begin{equation}
\widehat{p(k)}(\zeta) = \sum_{i=0}^d a_i \hat{k}(\zeta).
\end{equation}
A polynomial filter can be analyzed in the Fourier domain by looking into how the interval $I=[0,1]$ is mapped by the polynomial. This reflects how the differently attenuated Fourier components get amplified or attenuated by the polynomial filter. In Figure~\ref{fig:polynomial-vis} we show a plot of different example polynomials $p(x)$ when applied to a signal $x\in I$. 

If we apply the polynomial filter to an image $v$ that has been affected by the same base blur $k$ as the one used to build the polynomial filter, we get
\begin{equation}
p(K)v = p(K)Ku + p(K)n,
\end{equation}
where it becomes evident that there is a trade-off between inverting the blur (i.e., $p(K)K \approx Id.$) and avoiding noise amplification, i.e., $p(K)n \approx 0$. Our goal is to design polynomials that (i) try to invert the effect of blur in the frequencies that have not been significantly attenuated, and (ii) avoid over-amplifying frequencies affected by the blur.  The right plot in Figure~\ref{fig:polynomial-vis} shows $p(x) x$, which can be interpreted as the combined effect of applying the polynomial on a signal that has been attenuated by the same blur.

\subsection{Designing deblurring polynomial filters} 
We want to build a polynomial filter $p(K)K \approx Id$, when $K$ is close to the identity, or in terms of the polynomial, $p(x) \approx 1/x$ if $x\approx 1$. 
Note that a Taylor expansion of $1/x$ at $x=1$ of degree $d$, leads to the polynomial filter from truncating~\eqref{eq:inv_power_series} to degree $d$. This polynomial does a good job approximating the inverse, but significantly amplifies noise, particularly in the most attenuated components. Instead, we propose to approximate the inverse but have better control of how noise is amplified. Let us assume we design a polynomial of degree $d$, i.e., we need to define the $d+1$ coefficients. We will approximate the inverse by forcing the polynomial to have equal derivatives at the ones of the inverse function at $x=1$. This is done up to order $d-2$,
\begin{equation}
p^{(i)}(x=1) = (-1)^i \, i!,
\label{eq:inv_derivate_value}
\end{equation}
for $i=1,\ldots,d-2$. We also have the additional constraint that $p(x=1)=1$ (no change in luminosity). The remaining two degrees of freedom are left as \emph{design parameters}. We can control how the mid-frequencies get amplified by controlling $p^{(d-1)}(x=1)= \alpha$, and also, how noise is amplified at the frequencies that are completely attenuated by the blur, by controlling $p(x=0)=b$. This system of $d+1$ linear equations leads to a (closed-form) family of polynomials, $p_{d,\alpha,b}$. The value of $\alpha$ and $b$ should vary in a range close to the one given by the truncated power series, i.e., $\alpha = (-1)^d d!$ and  $b = d+1$.

\begin{figure}
\scriptsize
    \begin{center}
      \begin{overpic}[clip, trim=0 20 0 80, width=.245\linewidth]{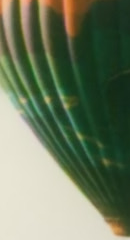}
        \put(3,3){\begin{color}{arsenic}Input\end{color}}
      \end{overpic}
      \begin{overpic}[clip, trim=0 20 0 80, width=.245\linewidth]{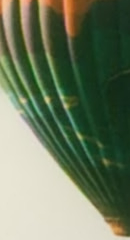}
        \put(2,2){\begin{color}{arsenic}$\alpha\!=\!4$\end{color}}
        \put(2,12){\begin{color}{arsenic}$b=\!1$\end{color}}
      \end{overpic}
      \begin{overpic}[clip, trim=0 20 0 80, width=.245\linewidth]{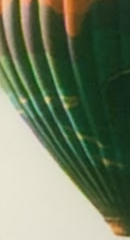}
        \put(2,2){\begin{color}{arsenic}$\alpha\!=\!8$\end{color}}
        \put(2,12){\begin{color}{arsenic}$b=\!1$\end{color}}
      \end{overpic}
      \begin{overpic}[clip, trim=0 20 0 80, width=.245\linewidth]{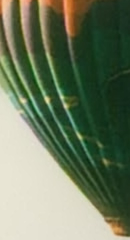}
        \put(2,2){\begin{color}{arsenic}$\alpha\!=\!16$\end{color}}
        \put(2,12){\begin{color}{arsenic}$b=\!1$\end{color}}
      \end{overpic} 
      
      \vspace{.2em}
 
      \begin{overpic}[clip, trim=0 20 0 80, width=.245\linewidth]{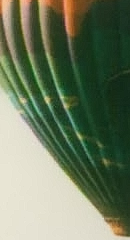}
        \put(2,2){\begin{color}{arsenic}$\alpha\!=\!2$\end{color}}
        \put(2,12){\begin{color}{arsenic}$b=\!4$\end{color}}
      \end{overpic}
      \begin{overpic}[clip, trim=0 20 0 80, width=.245\linewidth]{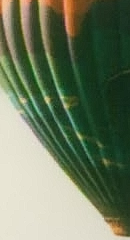}
        \put(2,2){\begin{color}{arsenic}$\alpha\!=\!4$\end{color}}
        \put(2,12){\begin{color}{arsenic}$b=\!4$\end{color}}
      \end{overpic}
      \begin{overpic}[clip, trim=0 20 0 80, width=.245\linewidth]{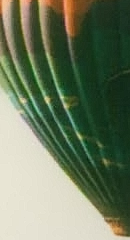}
        \put(2,2){\begin{color}{arsenic}$\alpha\!=\!8$\end{color}}
        \put(2,12){\begin{color}{arsenic}$b=\!4$\end{color}}
      \end{overpic}
      \begin{overpic}[clip, trim=0 20 0 80, width=.245\linewidth]{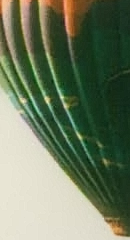}
         \put(2,2){\begin{color}{arsenic}$\alpha\!=\!16$\end{color}}
        \put(2,12){\begin{color}{arsenic}$b=\!4$\end{color}}
      \end{overpic}
    
    \end{center}
    
    \caption{Effect of using different Polynomials from the $p_{3,\alpha,b}$ family.}
    \label{fig:pol3_results}
\end{figure}
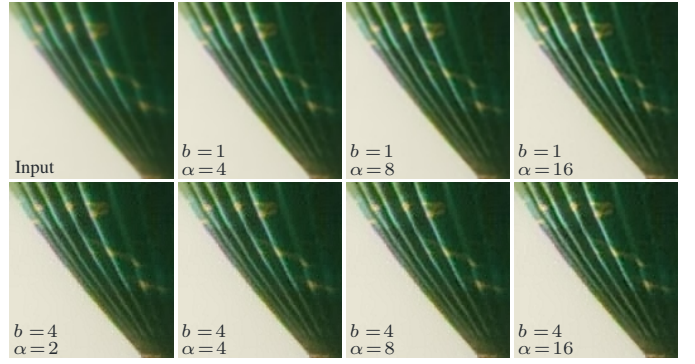

The polynomial computational cost is proportional to the order, so the lower, the better. We choose order three since it is the lowest order to control both mid-frequencies boosting ($\alpha$), and noise amplification ($b$).  The impact of using a low-order polynomial is that in a single application of the filter blur may still remain. However, the remaining blur may be removed by repeated applications of the filter as we show in the experimental results. The polynomial filter family of degree $d=3$ is
\begin{multline}
p_{3,\alpha,b}(x) = (\alpha/2 - b + 2)x^3 + (3b -\alpha -6)x^2 \\ 
  + (5 -3b + \alpha/2)x + b.
  \label{eq:pol3alphad}
\end{multline}
In Figure~\ref{fig:polynomial-vis} we show different polynomials and how their shape is affected by the design parameters $\alpha$ and $b$. All the results presented in this article are with polynomials of this third order family. As an illustration, we show in Figure~\ref{fig:pol3_results} the behavior using different coefficients in the polynomial deblurring.

\vspace{.5em}

\subsection{Generalization to any blur filter} 
\begin{figure}
     \setlength{\fboxrule}{1pt}
    \setlength{\fboxsep}{0pt}
\scriptsize
    \centering
    \begin{minipage}[c]{.4\linewidth}
    \centering
    
    \vspace{1.1em}
    \end{minipage}
    
    \begin{minipage}[c]{\linewidth}
    \begin{minipage}[c]{.245\linewidth}
    \centering
    {\color{blue}\fbox{\includegraphics[width=.45\linewidth]{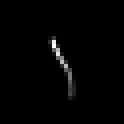}}}\vspace{.2em}

    \includegraphics[clip,trim=90 0 100 0, width=\linewidth]{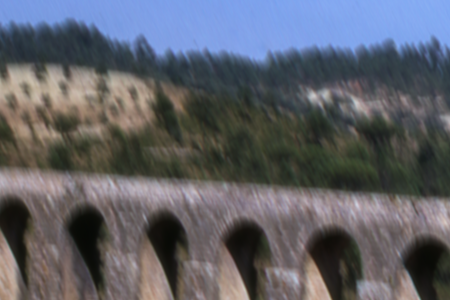}
    
    (a) blurry input
    \end{minipage}
    \begin{minipage}[c]{.245\linewidth}
    \centering
    
     {\color{orange}\fbox{\includegraphics[width=.45\linewidth]{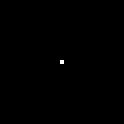}}}
     {\color{red}\fbox{\includegraphics[width=.45\linewidth]{./figs/correction/kernel_x4.png}}}\vspace{.2em}

    \includegraphics[clip,trim=90 0 100 0, width=\linewidth]{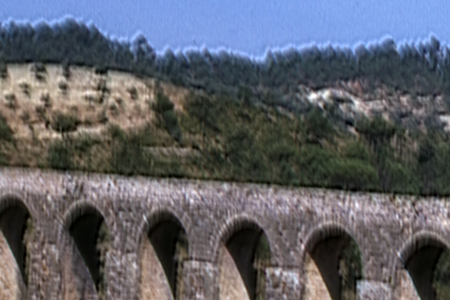}    
    
    (b) no-correction
    \end{minipage}   
    \begin{minipage}[c]{.245\linewidth}
    \centering

     {\color{orange}\fbox{\includegraphics[width=.45\linewidth]{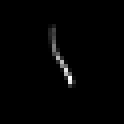}}}
     {\color{red}\fbox{\includegraphics[width=.45\linewidth]{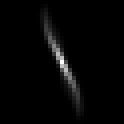}}}\vspace{.2em}
     
    \includegraphics[clip,trim=90 0 100 0, width=\linewidth]{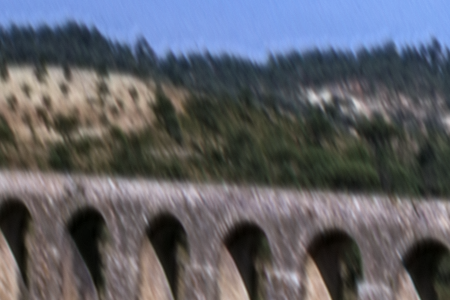}  
    
    (c) flip-correction
    \end{minipage}   
    \begin{minipage}[c]{.245\linewidth}
    \centering

     {\color{orange}\fbox{\includegraphics[width=.45\linewidth]{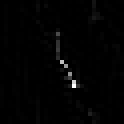}}}
     {\color{red}\fbox{\includegraphics[width=.45\linewidth]{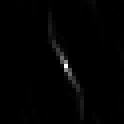}}}\vspace{.2em}

    \includegraphics[clip,trim=90 0 100 0, width=\linewidth]{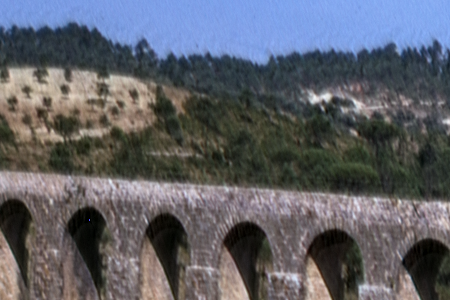}    
    (d) phase-correction
    \end{minipage}   
    \end{minipage}
    \vspace{.0em}

    \caption{Kernels with negative or complex Fourier coefficients (as the one shown in blue) need to be compensated before applying our polynomial deblurring.  
    In blue the input blur kernel $k$, in orange each respective correction filter $c_k$, while in red corrected kernels $h = c_k \ast k$ used in the Polynomial deblurring (b-d).}
    \label{fig:correcton-filter}
\end{figure}

Kernels  with  negative  or  complex  Fourier  coefficients (as the one shown in Fig.~\ref{fig:correcton-filter}a) cannot be directly deblurred using our Polynomial filtering (Fig.~\ref{fig:correcton-filter}b).  A way around this restriction is to apply a correction filter $c_k(\bx)$ to the input image so that total blur kernel $h=c_k \ast k$ has non-negative Fourier coefficients and the solution to the deblurring problem remains the same,
\begin{align}
v = u \ast k  \Longrightarrow  v_\text{eq} = v \ast c_k = u \ast k \ast c_k = u \ast k_\text{eq}.
\end{align}
Filtering the image with the flipped kernel, i.e., $c_k(\bx) = k(-\bx)$ will lead to $h(\bx)$ having real non-negative Fourier coefficients $\hat{h}(\zeta) = |\hat{k}(\zeta)|^2$. However, this correction filter introduces additional blur to the image, making the deblurring problem more ill-posed than the original one (Fig.~\ref{fig:correcton-filter}c).
An alternative correction filter is to compensate for the \emph{phase} but without introducing any additional attenuation of the spectra. This can be done by the pure phase filter,
$\hat{c}_k(\zeta) = \bar{\hat{k}}(\zeta) / |\hat{k}(\zeta)|$,
where $\bar{\hat{k}}(\zeta)$ denotes the complex conjugate of $\hat{k}(\zeta)$.
$c_k$ is a pure phase filter, i.e., it has a constant Fourier magnitude of one at all frequencies. 

In practice, given $v$ we can estimate $k$, and compute $c_k$. Then, we compute $v_\text{eq}$ and $k_\text{eq}$. The solution $u$, to $v_\text{eq} = u \ast k_\text{eq}$ is the same as in the original problem but $k_\text{eq}$ has non-negative Fourier coefficients allowing us to apply our polynomial deblurring. Note that in the Gaussian blur case no correction is needed.
As can be shown in the example in Figure~\ref{fig:correcton-filter}d, phase correction leads to a deblurred artifact-free image.
 
\section{Parametric mild blur model and estimation}
\label{sec:model}
Since we target mild-blur removal, there is not much gain in having a very fine model of the blur. We thus propose to model the blur kernel with an anisotropic Gaussian function, specified with three parameters: $\sigma_0$, the standard deviation of the main axis, $\rho = \sigma_1/\sigma_0$, the ratio between the the principal axis and the orthogonal one standard deviations, and $\theta$, the angle between the major axis and the horizontal. Thus, the Gaussian blur kernel at pixel $(x,y)$ is:
\begin{equation}
k(x,y) = Z \exp \left(- (a_0 x^2 + 2a_1xy + a_2y^2) \right), 
\end{equation}
where 
$a_0 = \frac{\cos(\theta)^2}{2\sigma_0^2} + \frac{\sin(\theta)^2}{2\rho^2\sigma_0^2}$, $a_1=\frac{\sin(2\theta)}{4\sigma_0^2}(\frac{1}{\rho^2}-1)$, 
$a_2= \frac{\sin(\theta)^2}{2\sigma_0^2} + \frac{\cos(\theta)^2}{2\rho^2\sigma_0^2}$ 
and $Z$ is a normalization constant so the area of the kernel is one. Although this is a rough model for arbitrary blur in images, if the blur is small, either directional or isotropic, the anisotropic Gaussian parameterization is reasonable. Figure~\ref{fig:calibration_model}(left) shows examples of Gaussian blur kernels.

\subsection{From natural image model to blur estimation}\label{subsec:natural}
Estimating a blur kernel given only an image is a challenging ill-posed problem. It is necessary to make some assumptions, for instance by using an image prior (learned from data or from a statistical or any kind of mathematical model).  In this work, we focus on efficiency and require that the estimation of the blur is very fast. Thus, variational models, for instance those typically involved in blind-deconvolution approaches are prohibitively expensive. Instead, we will rely on the following rough observations about the image gradient distribution. 

\begin{assumption}In a sharp image, the maximum value of the image gradient in any direction is mostly constant and roughly independent of the image.
\end{assumption}

\begin{assumption}If a sharp image is affected by Gaussian blur, the blur level in the direction of the principal axis will be linearly related to the inverse of the maximum image gradient in the principal directions. 
\end{assumption}
The validity of these assumptions is discussed in Appendix~\ref{ap:model_calibration_assumptions}. Based on these assumptions we proceed as follows. We first estimate the maximum magnitude of the image derivative at all possible orientations and then take the minimum value among them. This leads to
\begin{equation}
f_\theta = \min_{\psi \in [0,\pi)} f_\psi =  \min_{\psi \in [0,\pi)} \max_{\bx} | \nabla_\psi v(\bx)|,
\label{eq:grad_feature}
\end{equation}
where $v(\textbf{x})$ is the input image, and $\nabla_\psi f(\bx)  = \nabla v(\bx) \cdot (\cos \psi, \sin \psi)$, is the directional derivative of $v$ at direction $\psi$.
Then, the blur kernel parameters are
\begin{equation}
\theta = \argmin_{\psi \in [0,\pi)} f_\psi, \quad \sigma_0 = \frac{c}{f_\theta}\,\, \text{and} \quad \sigma_1 = \frac{c}{f_{\theta_\perp}},
\label{eq:model_angle}
\end{equation}
where $c$ is a coefficient controlling the assumed linear relation between the gradient feature and the level of blur.

\vspace{.3em}
\noindent \textbf{Observation.} The estimation of the image gradient at any given image pixel introduces additional blur (since some sort of interpolation is needed). This will be in addition to the image blur that the image already has. Let us assume  the blur estimation introduces an isotropic Gaussian blur of strength $\sigma_b$, then the total blur of the image will be approximately ${\sigma^2_{0_T}}= \sigma_b^2+ \sigma_0^2$ (due to the semigroup property of Gaussian function). This leads to
\begin{align}
\sigma_0 = \sqrt{\frac{c^2}{f_\theta^2} - \sigma_b^2}, \quad \sigma_1 = \sqrt{\frac{c^2}{f_{\theta_\perp}^2} - \sigma_b^2},
\label{eq:model_final}
\end{align}
where $c$ and $\sigma_b$ are two parameters to be calibrated.

\subsection{Model calibration.}
To calibrate the parameters $c$ and $\sigma_b$ we proceed as follows. Given a set of sharp high quality images, we simulate $K=1000$ random Gaussian blurry images, by randomly sampling the blur space and the image set. The Gaussian blur kernels are generated by sampling random values for $\sigma_0  \in [0.3, 4]$ and $\rho \in [0.15,1]$. Additive Gaussian white noise of standard deviation $1\%$ is added to each simulated blurry image.

For each of the blurry images we compute the gradient features and the maximum and minimum values according to~\eqref{eq:grad_feature}. The parameters $c$ and $\sigma_b$ are estimated by minimizing the mean absolute error. Figure~\ref{fig:calibration_model} presents the calibration results. Additional details are included in Appendix~\ref{ap:model_calibration_assumptions}.

\begin{figure}
    \centering
    \begin{minipage}[c]{.22\linewidth}
    \centering
    \includegraphics[width=\linewidth]{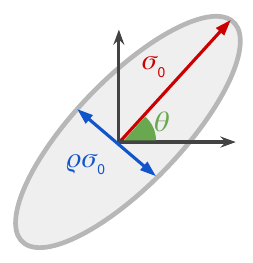}
    
    \includegraphics[width=\linewidth]{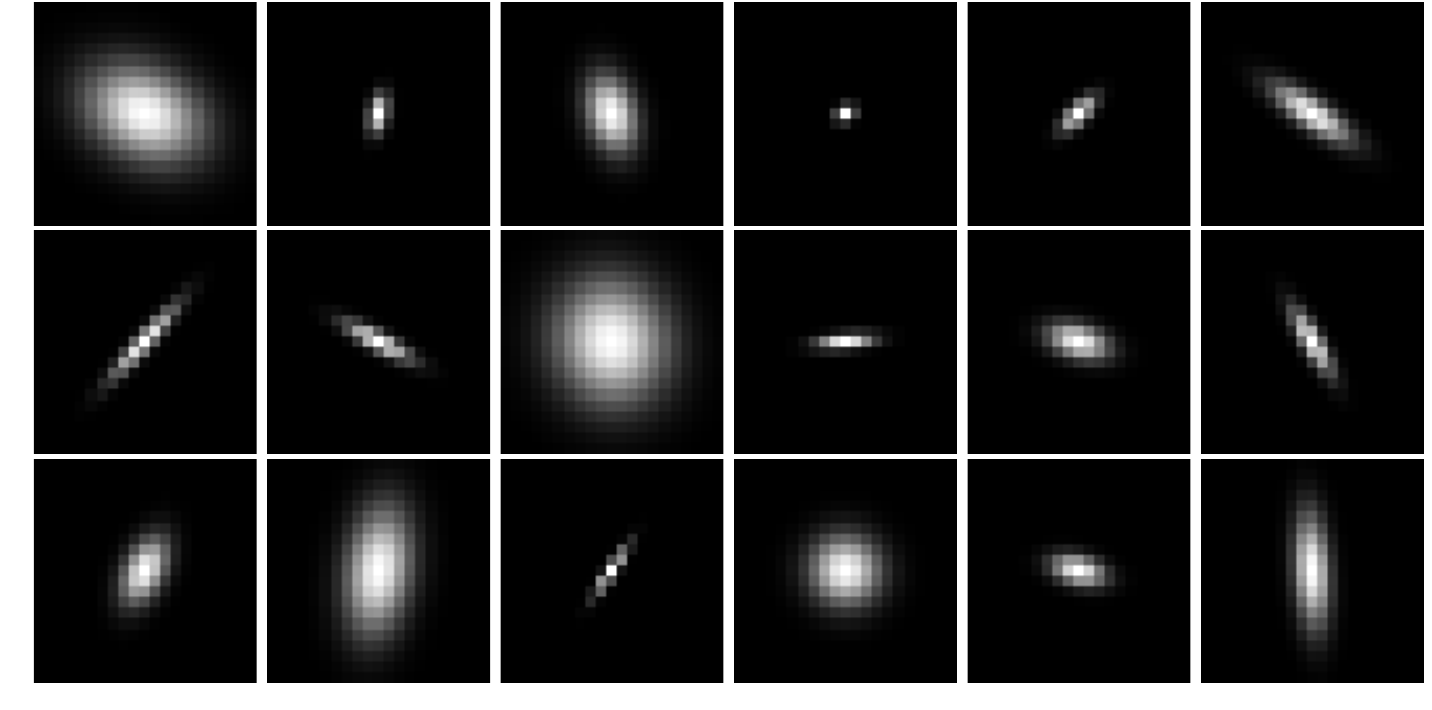}
    
    \end{minipage}
    \begin{minipage}[c]{.76\linewidth}
    \centering
    \includegraphics[width=\columnwidth]{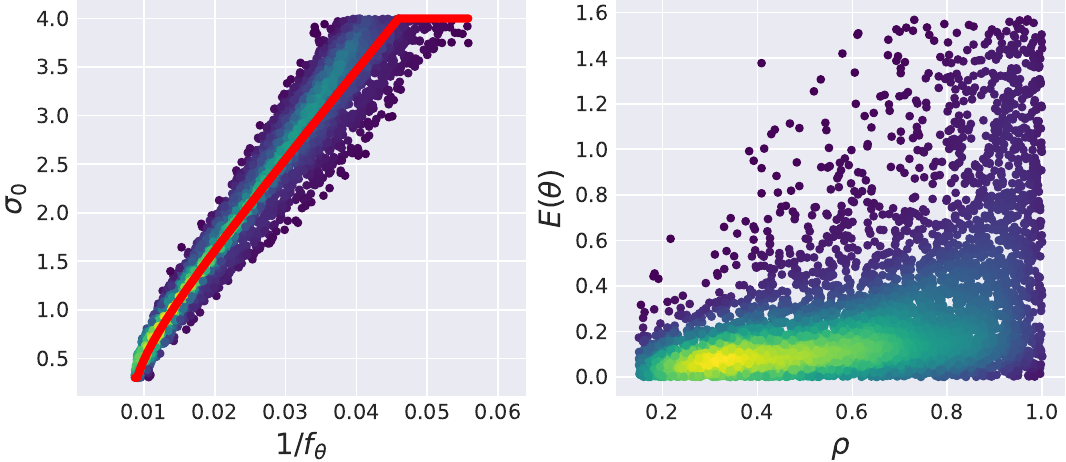}
    
    \end{minipage}
    
    \caption{Gaussian blur model and examples (left). The middle plot shows the relationship between the computed gradient feature $f_\theta$ and $\sigma_0$ on simulated images, and the calibrated model (Eq.~\eqref{eq:model_final}, red curve). The right plot shows the error on the estimation of the blur angle $\theta$, as a function of $\rho$. Angle estimation error is very low for anisotropic (directional) kernels (low $\rho$ value).}
    \label{fig:calibration_model}
\end{figure}

\section{Polyblur implementation details}
\label{sec:implementation_details}
In this section we describe a Polyblur implementation to allow us to achieve the performance to deblur a 12MP image on a modern mobile platform in a fraction of a second. The core parts of the algorithm are implemented using the language~\textit{Halide}~\cite{ragan2012decoupling}.  
Halide separates the algorithm implementation from its execution schedule (i.e., loop nesting, parallelization, etc) enabling us us to more easily fuse pipeline stages for locality and to make use of SIMD and thread parallelism. Note that despite of this choice, most part of the algorithm can run in parallel and a parallel implementation could take full advantage of the parallel processing power of modern GPUs.\vspace{.3em}

\noindent \textbf{Blur estimation.}
We follow Sec.~\ref{subsec:natural} to estimate the blur. As the first step we normalize the image using quantiles ($q=0.0001$ and $1-q$) to be robust to outliers. From the image gradient we compute the directional derivative at $m=6$ angles uniformly covering $[0, \pi)$. The maximum direction of the magnitude at each angle is found and among the $m$ maximum values, we find the minimum value and angle ($f_0, \theta_0)$ through bicubic interpolation. Using~(\ref{eq:model_final}) we compute $\sigma_0$ and $\sigma_1$. More details about the blur estimation, as long as a pseudo-code for the blur estimation step, are included in Appendix~\ref{ap:blur_estimation_details}.  \vspace{.3em}

\noindent \textbf{Deblurring step.}
The deblurring filter has a closed form given by the estimated blur and the polynomial in~(\ref{eq:pol3alphad}). In the case of using a third order polynomial, the restoration filter support is roughly three times the one of the estimated base blur. 
Large blur kernel convolutions can be efficiently computed in Fourier domain. 
If the Gaussian blur is separable, the convolution can be efficiently computed in-place. In this case, we do not compute the polynomial filter and directly apply and accumulate repeated applications of the Gaussian blur as follows:
\begin{align}
v_0 = a_d v, \,\,  v_i &= k \ast v_{i-1} + a_{d-i} v, \,\, \text{for} \,\, i\!=\!1,\ldots,d,
\end{align}
where $v$ is the input blurry image, $k$ the estimated blur kernel, and $a_0,\ldots,a_d$ the deblurring polynomial coefficients, and $v_d$ the deblurred output image.
Non-separable Gaussian filtering can also be efficiently computed by two 1D Gaussian filters in non-orthogonal axis. This can further optimize our procedure. \vspace{.3em}

\noindent \textbf{Halo detection and removal.}
Halos can be generated due to misestimation of blur or more generally due to model mismatch. Halos appear at pixels where the blurry image and restored image have opposite gradients (gradient reversal). Let $v(\bx)$ be the blurry image and $\bar{v}(\bx)$ the deblurred one. Pixels with gradient reversal are those where
$
M(\bx) = - \nabla v(\bx) \cdot \nabla \bar{v}(\bx),
$
is positive.

\begin{figure}
\scriptsize
\begin{minipage}[c]{\linewidth}
    \centering

     \begin{tikzpicture}
      \node[anchor=north west,inner sep=0] (image) at (0,0) {\includegraphics[clip, trim=100 320 30 270, width=.225\linewidth]{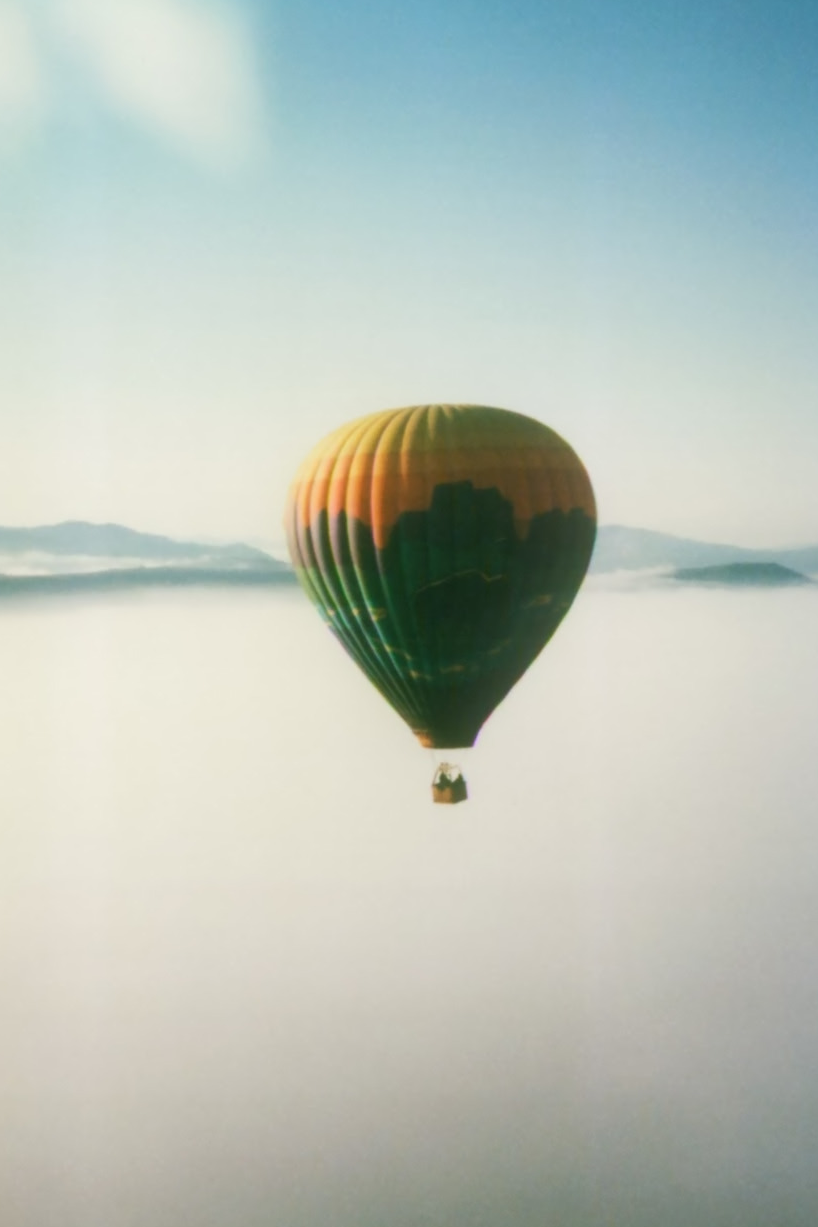}};
      \begin{scope}[x={(image.north east)},y={(image.south west)}]
        \draw[red1, very thick] (0.55, 0.55) rectangle (0.7, 0.7);
        \draw[orange1, ultra thick] (0.28,0.77) -- (0.7,0.15);
      \end{scope}
     \end{tikzpicture}
    \begin{overpic}[clip, trim=110 70 80 220, width=.248\linewidth]{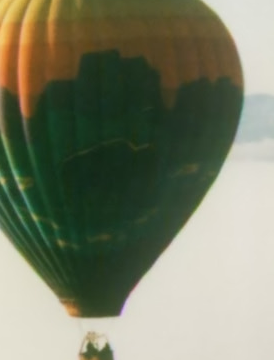}
    \put(60,8){\begin{color}{arsenic}input\end{color}}
    \end{overpic}
    \begin{overpic}[clip, trim=110 70 80 220, width=.248\linewidth]{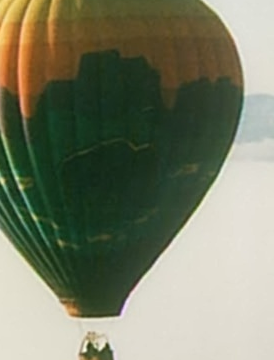}
    \put(60,8){\begin{color}{arsenic}no-\textsc{hr}\end{color}}
    \end{overpic}
    \begin{overpic}[clip, trim=110 70 80 220, width=.248\linewidth]{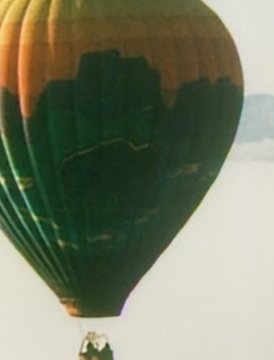}
    \put(60,8){\begin{color}{arsenic}w/\textsc{hr}\end{color}}
    \end{overpic}

\end{minipage}  
\begin{minipage}[c]{\linewidth}
    \includegraphics[width=\linewidth]{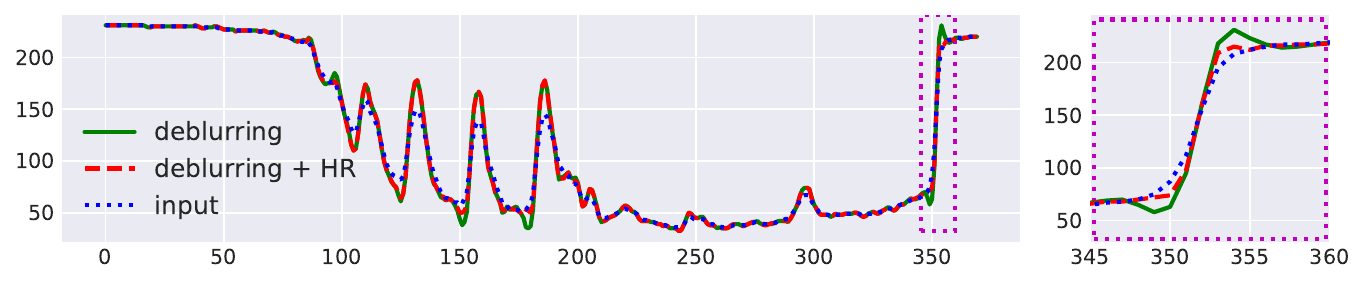}
\end{minipage}
    
    \caption{Halo Removal (\textsc{hr}). Top row shows a crop for the input image, the deblurred one, and the final merge with the halo removal step. In the bottom we show a profile (and a zoom-in on the right) of the insensitive values orthogonal to the balloon boundary (orange segment). The gradient reversal has been eliminated.}
    \label{fig:halo_removal}
\end{figure}

Let us compute a new image $\bar{v}_z(\bx)$ formed as a per pixel convex combination (blending) of the input image $v(\bx)$ and the deblurred one $\bar{v}(\bx)$ using a weights $z(\bx) \in [0,1]$,
\begin{align}
\label{eq:convex_combination}
v_z(\bx) = z(\bx) v(\bx) + (1-z(\bx)) \bar{v}(\bx).
\end{align}
We want to avoid halos in the final image $v_z(\bx)$, but keep as much as possible of the deblurred one $\bar{v}(\bx)$. If $z(\bx)$ does not change too fast, then
\begin{equation}
\nabla v_z (\bx) =  z (\bx) \nabla v(\bx) + (1-z(\bx)) \nabla \bar{v}(\bx).
\end{equation}
To avoid halos on the final image we require $\nabla v(\bx) \cdot \nabla v_z(\bx) >0$. Then, we get that for pixels where $M(\bx)>0$, having
$$
z(\bx) \le \frac{M(\bx)}{\|\nabla v(\bx)\|^2 + M(\bx)},
$$
leads to a new image that does not have  any gradient reversal introduced in the deblurring step. To keep as much as possible of the deblurred image, $z(\bx)$ should be as small as possible. Then the final image is generated by adopting
\begin{align}
z(\bx) = \max \left(\frac{M(\bx)}{\|\nabla v(\bx)\|^2 + M(\bx)}, 0 \right),
\end{align}
in the convex combination given by~\eqref{eq:convex_combination}.
Figure~\ref{fig:halo_removal} shows an example of a filtered image and the halo correction. \vspace{.3em}

\noindent \textbf{Polyblur iterated.} 
To remove some remaining image blur, we can re-apply Polyblur. This implies re-estimating the image blur on the previous deblurred image, and applying a new polynomial deblurring. In the next section we present some results when iterating polyblur multiple times.

\section{Experiments}
\label{sec:experiments}
We carried out a series of experiments to evaluate Polyblur and compare against other blind deblurring~\cite{goldstein2012blur,hosseini2019convolutional,kupyn2019deblurgan,pan2016l0,tao2018scale,zhang2013multi} and sharpening methods~\cite{he2012guided,zhu2011restoration}. The comparison is threefold: we compare traditional metrics such as PSNR, SSIM and perceptual ones such as LPIPS~\cite{zhang2018unreasonable}; we do a visual inspection of artifacts (qualitative); and we report processing times. We also present results on images \emph{in the wild} having mild blur. Finally, we show how Polyblur can be used to remove blur before applying an off-the-shelf image superresolution deep method. Table~\ref{tab:performance} shows average execution times for each step in the Polyblur algorithm. Our method can process a 12MP image on a modern mobile platform in 600ms.
\vspace{.3em}

\begin{figure}[t]
 \footnotesize
 \centering
 \begin{minipage}[c]{\linewidth}
  \centering
  \begin{overpic}[width=.33\linewidth]{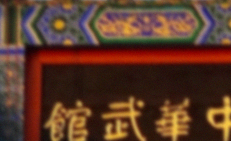}
    \put(3,3){\begin{color}{black}Input\end{color}}
    \put(2,2){\begin{color}{white}Input\end{color}}
  \end{overpic}\hspace{-.15em}
  \includegraphics[width=.33\linewidth]{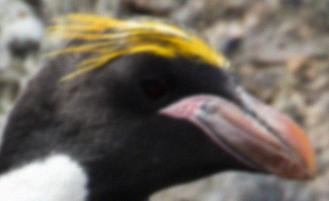}\hspace{-.17em}
  \includegraphics[width=.33\linewidth]{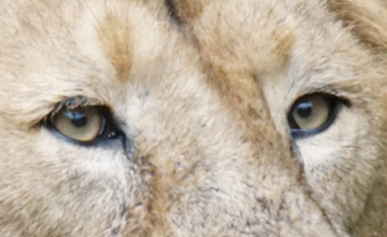}
 \end{minipage}\vspace{.15em}
 \begin{minipage}[c]{\linewidth}
  \centering
  \begin{overpic}[width=.33\linewidth]{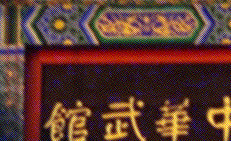}
    \put(3,3){\begin{color}{black}ConvDeblurring\end{color}}
    \put(2,2){\begin{color}{white}ConvDeblurring\end{color}}
  \end{overpic}\hspace{-.15em}
  \includegraphics[width=.33\linewidth]{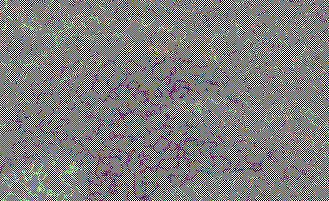}\hspace{-.17em}
  \includegraphics[width=.33\linewidth]{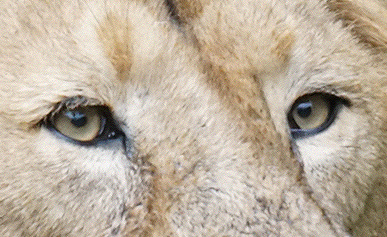}
 \end{minipage}\vspace{.15em}
 \begin{minipage}[c]{\linewidth}
  \centering
  \begin{overpic}[width=.33\linewidth]{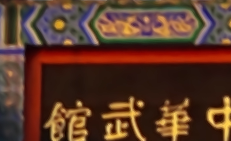}
    \put(3,3){\begin{color}{black}GLAS\end{color}}
    \put(2,2){\begin{color}{white}GLAS\end{color}}
  \end{overpic}\hspace{-.15em}
  \includegraphics[width=.33\linewidth]{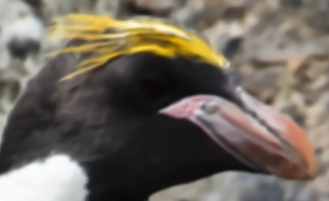}\hspace{-.17em}
  \includegraphics[width=.33\linewidth]{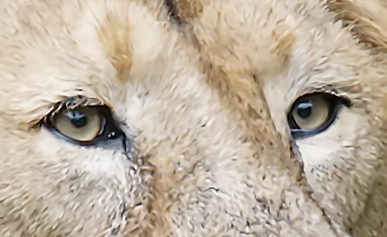}
 \end{minipage}\vspace{.15em}
 \begin{minipage}[c]{\linewidth}
  \centering
  \begin{overpic}[width=.33\linewidth]{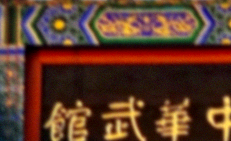}
    \put(3,3){\begin{color}{black}GuidedFilter\end{color}}
    \put(2,2){\begin{color}{white}GuidedFilter\end{color}}
  \end{overpic}\hspace{-.15em}
  \includegraphics[width=.33\linewidth]{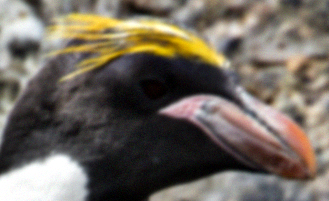}\hspace{-.17em}
  \includegraphics[width=.33\linewidth]{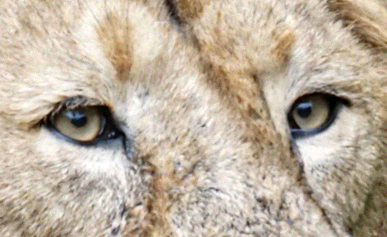}
 \end{minipage}\vspace{.15em}
 \begin{minipage}[c]{\linewidth}
  \centering
  \begin{overpic}[width=.33\linewidth]{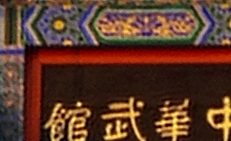}
    \put(3,3){\begin{color}{black}Spectral-Irr.\end{color}}
    \put(2,2){\begin{color}{white}Spectral-Irr.\end{color}}
  \end{overpic}\hspace{-.15em}
  \includegraphics[width=.33\linewidth]{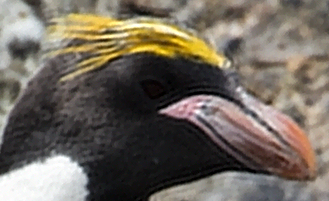}\hspace{-.17em}
  \includegraphics[width=.33\linewidth]{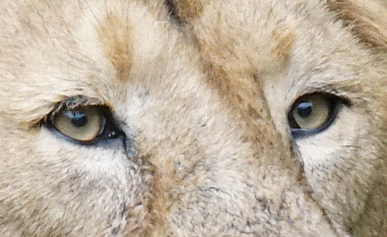}
 \end{minipage}\vspace{.15em}
 \begin{minipage}[c]{\linewidth}
  \centering
  \begin{overpic}[width=.33\linewidth]{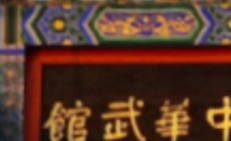}
    \put(3,3){\begin{color}{black}SRN-Deblur\end{color}}
    \put(2,2){\begin{color}{white}SRN-Deblur\end{color}}
  \end{overpic}\hspace{-.15em}
  \includegraphics[width=.33\linewidth]{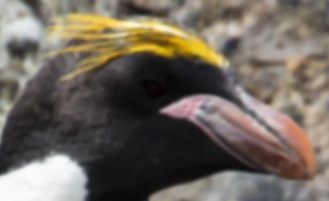}\hspace{-.17em}
  \includegraphics[width=.33\linewidth]{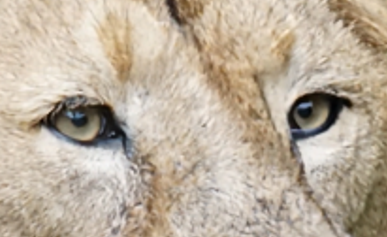}
 \end{minipage}\vspace{.15em}
 \begin{minipage}[c]{\linewidth}
  \centering
  \begin{overpic}[width=.33\linewidth]{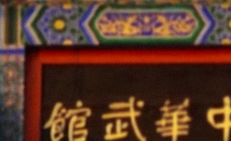}
    \put(3,3){\begin{color}{black}DeblurGANv2\end{color}}
    \put(2,2){\begin{color}{white}DeblurGANv2\end{color}}
  \end{overpic}\hspace{-.15em}
  \includegraphics[width=.33\linewidth]{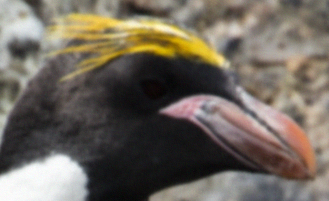}\hspace{-.17em}
  \includegraphics[width=.33\linewidth]{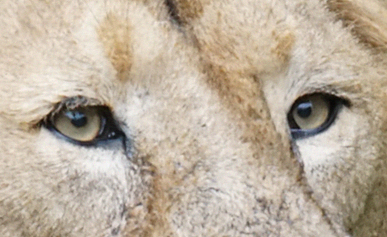}
 \end{minipage}\vspace{.15em}
 \begin{minipage}[c]{\linewidth}
  \centering
  \begin{overpic}[width=.33\linewidth]{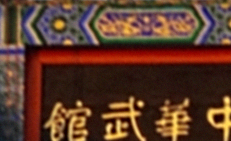}
    \put(3,3){\begin{color}{black}Polyblur\end{color}}
    \put(2,2){\begin{color}{white}Polyblur\end{color}}
  \end{overpic}\hspace{-.15em}
  \includegraphics[width=.33\linewidth]{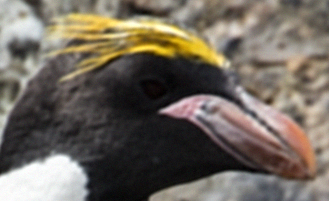}\hspace{-.17em}
  \includegraphics[width=.33\linewidth]{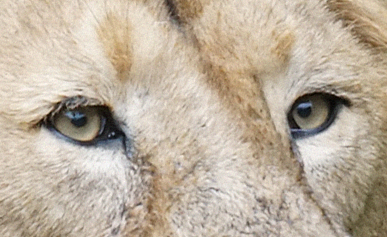}
 \end{minipage}\vspace{.15em}
 \begin{minipage}[c]{\linewidth}
  \centering
  \begin{overpic}[width=.33\linewidth]{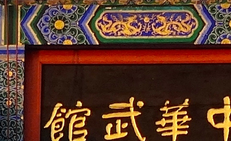}
    \put(3,3){\begin{color}{black}Groundtruth\end{color}}
    \put(2,2){\begin{color}{white}Groundtruth\end{color}}
  \end{overpic}\hspace{-.15em}
  \includegraphics[width=.33\linewidth]{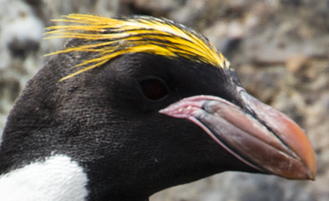}\hspace{-.17em}
  \includegraphics[width=.33\linewidth]{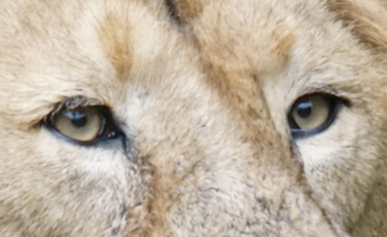}
 \end{minipage}
 \caption{Comparison of Polyblur with other deblurring and adaptive sharpening methods on synthetically blurred DIV2K dataset.}
 \label{fig:synthetic_dataset}
\end{figure}

\noindent \textbf{Comparison on simulated blur.}
We generated a mild-blur dataset by artificially blurring sharp images from the DIV2K validation dataset. Mild blur is simulated by applying a random Gaussian kernel of different sizes, shapes, and orientations ($\sigma_0  \sim \mathcal{U}[0.3, 4]$). Additive white Gaussian noise of standard deviation $1\%$ is added on top. More details are included in the supplementary material.
Polyblur produces the best results in terms of PSNR and MS-SSIM while being significantly faster than most of the other deblurring methods (Table~\ref{tab:div2k-polyblur}). As shown in Figure~\ref{fig:synthetic_dataset}, Polyblur leads to naturally pleasant images, while most of the compared methods introduce artifacts. The visual quality of the output results is similar to highly complex methods such as DeblurGAN-v2~\cite{kupyn2019deblurgan}. Our CPU implementation of Polyblur is still $50\%$ faster than the highly optimized mobile version of DeblurGANv2 that runs on GPU.

Re-applying Polyblur produces slightly worse quantitative results on the simulated Gaussian blur dataset (Table~\ref{tab:div2k-polyblur}) but in general leads to better qualitative results on real images where blur might not be perfectly Gaussian (e.g., Figures 23--27 in the supplemental material).

\begin{table}
\centering
\small
\begin{tabular}{lccc}
\toprule
Algorithm phase                         & 12MP   & 3MP   & 1MP          \\ \midrule
1. Blur estimation          & 302 (357) & 88 (95) & 23 (25) \\
2a. Polyblur (separable)    & 66 (194)  & 16 (82) &  9 (35)   \\  
2b. Polyblur (Fourier) & 266 (472) & 35 (212) & 16(134)\\
3. Halo removal            & 14 (70)  & 61 (20)  & 3 (7)
\end{tabular}
\vspace{-.5em}
\caption{Times in ms run on desktop(mobile). Desktop: Intel Xeon Haswell 2.3GHz, mobile: Snapdragon 855.}
\label{tab:performance}
\end{table}

\begin{table}[h]
\small
    \centering
    \begin{tabular}{lcccc} \toprule
         Method & PSNR & MS-SSIM & LPIPS & Time\\\midrule
         SRN~\cite{tao2018scale} & 27.58 & 0.948 & 0.304 & 75-152$^\dagger$\\ %
         SparseDeb~\cite{zhang2013multi} & 27.48 & 0.940 & 0.296 & $>5000$ \\
         DebGANv2~\cite{kupyn2019deblurgan} & 29.79 & 0.963 & \blue{0.213} & 350$^\dagger$\\
         DebGANv2m~\cite{kupyn2019deblurgan} & 29.53 & 0.958 & \red{0.209} & \blue{60}$^\dagger$\\
         Spectral~\cite{goldstein2012blur}  & 27.14 & 0.950 & 0.230 & $>5000$ \\
         Deblur-l0~\cite{pan2016l0}  & 27.31 & 0.945 & 0.263 & $>5000$ \\
         GLAS~\cite{zhu2011restoration}  & 28.15 & 0.957 & 0.284 & $>5000$\\
         GuidFilt~\cite{he2012guided}  & 26.98 & 0.934 & 0.282 & 80\\
         ConvDeb~\cite{hosseini2019convolutional}$^\ast$ &27.08 & 0.938 & 0.238 & 90\\
         Polyblur-1it & \red{30.38} & \red{0.968} & 0.236 & \red{42}\\
         Polyblur-2it & \blue{30.06} & \blue{0.964} & 0.255 & 84\\
         Polyblur-3it & 29.44 & 0.955 & 0.269 & 126\\
         Blurry-noisy & 29.31 & 0.956 & 0.245 & -
    \end{tabular}
    \vspace{-.3em}
    \caption{Mild deblurring on the synthetic dataset generated from DIV2K. Lower LPIPS values imply better quality.  Polyblur-$N$it indicates applying Polyblur $N$ times. ConvDeb~\cite{hosseini2019convolutional}$^\ast$ did not produce reasonable results in 32/100 images (average values are computed in the rest). Times are in ms (for a 1MP image), and we present them here as an indicative reference. $^\dagger$ indicates GPU.}
    \label{tab:div2k-polyblur}
\end{table}

\begin{figure}
\scriptsize
\centering
\begin{minipage}[c]{1.0\linewidth}
\centering
 \begin{minipage}[c]{0.245\linewidth}
  \centering
  \includegraphics[width=\linewidth]{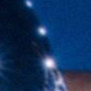}\vspace{.15em}
  
  \begin{overpic}[width=\linewidth]{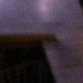}
    \put(3,3){\begin{color}{white}input\end{color}}
  \end{overpic}
  
 \end{minipage}\hspace{-.1em}
 \begin{minipage}[c]{0.245\linewidth}
  \centering
  \includegraphics[width=\linewidth]{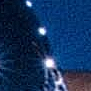}\vspace{.15em}
  
  \begin{overpic}[width=\linewidth]{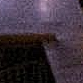}
    \put(3,3){\begin{color}{white}no-prefilter\end{color}}
  \end{overpic}
  
 \end{minipage}\hspace{-.1em}
  \begin{minipage}[c]{0.245\linewidth}
  \centering
  \includegraphics[width=\linewidth]{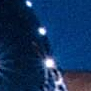}\vspace{.15em}
  
  \begin{overpic}[width=\linewidth]{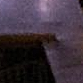}
    \put(3,3){\begin{color}{white}w/pull-push\end{color}}
  \end{overpic}\
  
 \end{minipage}\hspace{-.1em}
  \begin{minipage}[c]{0.245\linewidth}
  \centering
  \includegraphics[width=\linewidth]{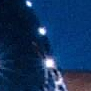}\vspace{.15em}
  
  \begin{overpic}[width=\linewidth]{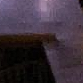}
    \put(3,3){\begin{color}{white}w/dom. transform\end{color}}
  \end{overpic}
  
 \end{minipage}
\end{minipage}
 
\vspace{.1em}
\caption{Pre-filtering. When the input image presents noise and other artifacts, Polyblur can amplify artifacts. As a pre-step, we may apply any filter that separates high-frequency texture and other details~\cite{gastal2011domain,isidoro2018pull}, apply Polyblur ($\alpha\!=\!6$, $b\!=\!1$), and finally add back the residual.}
\label{fig:denoise}
\end{figure}

\vspace{.3em}
\noindent \textbf{Dealing with noise and compression artifacts.}
If computational resources are available, a prefiltering step separating noise-like structure from the rest can be applied. If the input image is very noisy or has compression artifacts, this prefiltering step will keep Polyblur from amplifying artifacts present in the image. Since the residual image can be added back at the end, this step does not need to be carried out by a state-of-the-art denoiser. %
Fig.~\ref{fig:denoise} shows an example of Polyblur on a noisy and compressed image. We evaluated two alternatives, the Pull-Push denoiser~\cite{isidoro2018pull} and the Domain Transform~\cite{gastal2011domain} edge-preserving filter.  In both cases we achieve the desired deblurring, and a pleasant result.

\vspace{.5em}

\noindent \textbf{Results on images \emph{in the wild}.} In Figure~\ref{fig:teaser} we present a selection of results of Polyblur applied to some images in the wild. As shown, Polyblur manages to remove mild blur, as the one present in most images, without introducing any new artifacts. Please refer to the supplemental material for comparison with other state-of-the-art deblurring and sharpening methods.

\begin{figure}[t]
\scriptsize
    \centering
    \begin{minipage}[c]{.245\linewidth}
   \centering
   \begin{overpic}[width=1.\linewidth]{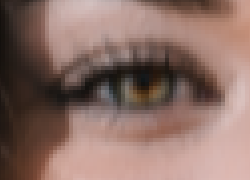}
     \put(3,4){\begin{color}{white}input (low-res)\end{color}}
   \end{overpic}\vspace{.15em}
   
    \begin{overpic}[width=1.\linewidth]{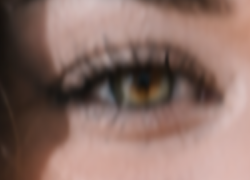}
     \put(2,2){\begin{color}{white}\textsc{edsr}\end{color}}
   \end{overpic}\vspace{.15em}
   
    \begin{overpic}[width=1.\linewidth]{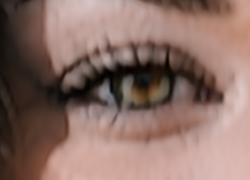}
     \put(2,2){\begin{color}{white}KernelgGAN + \textsc{zssr}\end{color}}
   \end{overpic}\vspace{.15em}
   
    \begin{overpic}[width=1.\linewidth]{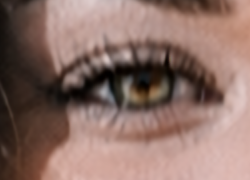}
     \put(2,2){\begin{color}{white}polyblur-1it + \textsc{edsr}\end{color}}
   \end{overpic}\vspace{.15em}
   
    \begin{overpic}[width=1.\linewidth]{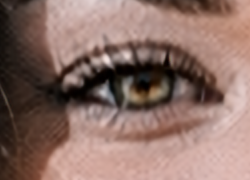}
     \put(2,2){\begin{color}{white}polyblur-2it + \textsc{edsr}\end{color}}
   \end{overpic}\vspace{.15em}
   
    \begin{overpic}[width=1.\linewidth]{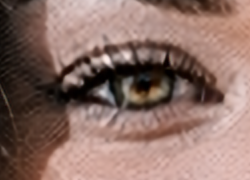}
     \put(2,2){\begin{color}{white}polyblur-3it + \textsc{edsr}\end{color}}
   \end{overpic}\vspace{.15em}
   
   \begin{overpic}[width=1.\linewidth]{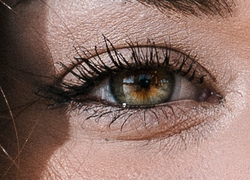}
     \put(2,2){\begin{color}{white}groundtruth\end{color}}
   \end{overpic}

    \end{minipage}\hspace{-.15em}
    \begin{minipage}[c]{.245\linewidth}
    \centering
    \includegraphics[width=\linewidth]{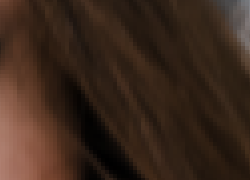}\vspace{.15em}
    
    \includegraphics[width=\linewidth]{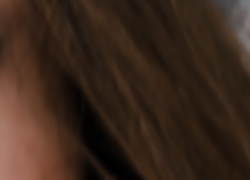}\vspace{.15em}
    
    \includegraphics[width=\linewidth]{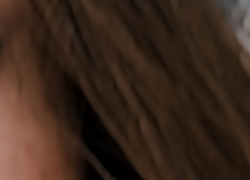}\vspace{.15em}
    
    \includegraphics[width=\linewidth]{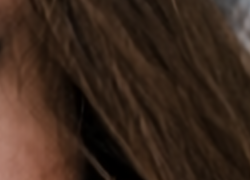}\vspace{.15em}
    
    \includegraphics[width=\linewidth]{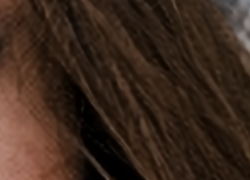}\vspace{.15em}
    
    \includegraphics[width=\linewidth]{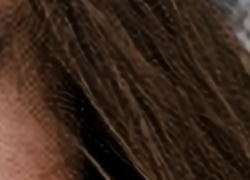}\vspace{.15em}
    
    \includegraphics[width=\linewidth]{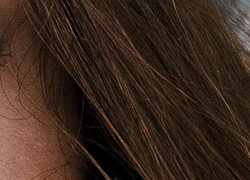}
     
    \end{minipage}\hspace{-.15em}
    \begin{minipage}[c]{.245\linewidth}
    \centering
    
    \includegraphics[width=\linewidth]{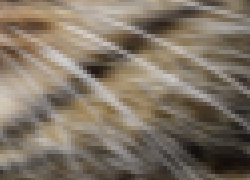}\vspace{.15em}
    
    \includegraphics[width=\linewidth]{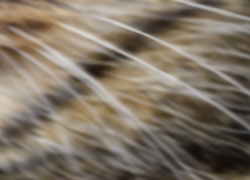}\vspace{.15em}
    
    \includegraphics[width=\linewidth]{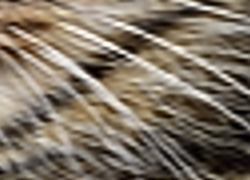}\vspace{.15em}
    
    \includegraphics[width=\linewidth]{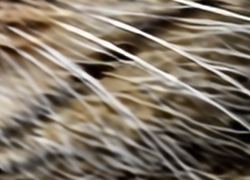}\vspace{.15em}
    
    \includegraphics[width=\linewidth]{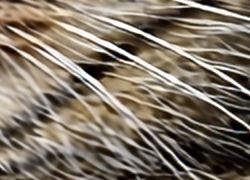}\vspace{.15em}
    
    \includegraphics[width=\linewidth]{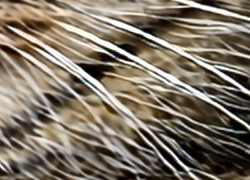}\vspace{.15em}
    
    \includegraphics[width=\linewidth]{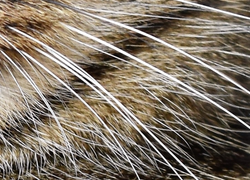}
    
    \end{minipage}\hspace{-.15em}
    \begin{minipage}[c]{.245\linewidth}
    \centering
    
    \includegraphics[width=\linewidth]{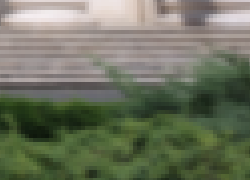}\vspace{.15em}
    
    \includegraphics[width=\linewidth]{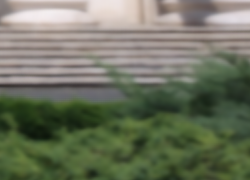}\vspace{.15em}
    
    \includegraphics[width=\linewidth]{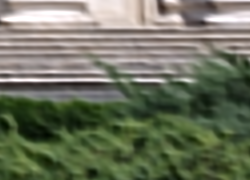}\vspace{.15em}
    
    \includegraphics[width=\linewidth]{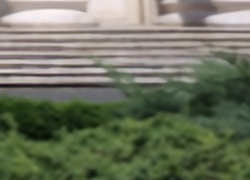}\vspace{.15em}
    
    \includegraphics[width=\linewidth]{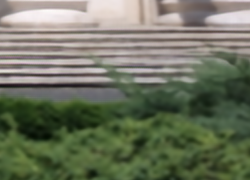}\vspace{.15em}
    
    \includegraphics[width=\linewidth]{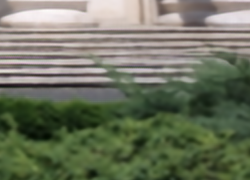}\vspace{.15em}
    
    \includegraphics[width=\linewidth]{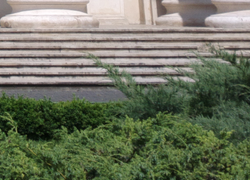}

    \end{minipage}
    \vspace{0em}
    \caption{Image $4\times$ super-resolution using an off-the-shelf deep SISR model with \emph{unknown} PSF (EDSR~\cite{lim2017enhanced}). More results are given in the supplementary material. }
    \label{fig:edsr-polyblur}
\end{figure}

\vspace{.5em}
\noindent \textbf{Deblurring before super-resolution.}
Single image super-resolution has seen remarkable progress in the last few years mostly due the use of deep models trained on image datasets. 
The common practice~\cite{yang2019deep} is to simulate low-resolution training images by simply applying a bicubic downsampling operator.  Unfortunately,  inference performance suffers significantly if images do not tightly follow the training distribution. 
We evaluate the performance of applying Polyblur as a pre-step before using an off-the-shelf deep network for doing $4\times$ image upscaling.  We trained from scratch an EDSR~\cite{lim2017enhanced} network with 32 layers and 64 filters using DIV2K training dataset. We compare against KernelGAN~\cite{bell2019blind} and the Correction filter introduced in~\cite{hussein2020correction}. 
Figure~\ref{fig:edsr-polyblur} shows examples of applying Polyblur before upscaling the image. Table~\ref{tab:edsr-polyblur} shows a quantitative comparison of our approach and the competitors. Polyblur produces the best quantitative and qualitative results.

\begin{table}[h]
\small
    \centering
    \begin{tabular}{lcc} \toprule
         Method & PSNR & SSIM  \\\midrule
         Bicubic & 25.354 & 0.6775 \\
         EDSR~\cite{lim2017enhanced} & 25.628 & 0.6971 \\
         KernelGAN~\cite{bell2019blind} & 26.810 & 0.7316 \\
         CorrectionFilter\cite{hussein2020correction} + EDSR & 26.204 & 0.736 \\
         Polyblur-1it + EDSR &\blue{26.966} & \blue{0.7396} \\
         Polyblur-2it + EDSR & \red{27.115} & \red{0.7452} \\
         Polyblur-3it + EDSR & 26.955 & 0.7368 \\
    \end{tabular}
    \vspace{-.3em}
    \caption{4x Super-resolution on DIV2KRK dataset~\cite{bell2019blind}. Polyblur-$N$it indicates applying Polyblur $N$ times. PSNR and SSIM values are directly taken from~\cite{bell2019blind,hussein2020correction}. In red font the best method, in blue the second best.}
    \label{tab:edsr-polyblur}
\end{table}

\section{Conclusion}
\label{sec:conclusion}
We introduced a highly efficient algorithm for estimating and removing mild blur that is ubiquitously present in many captured images (especially on handheld devices). A parametric blur kernel is first estimated using simple gradient features that encode the direction and intensity of the blur. The estimated blur is then removed by combining multiple applications of the blur kernel in a well founded way that allows us to approximate the inverse while controlling the noise amplification. Our method successfully handles blurs that are reasonably well captured by an anisotropic Gaussian of standard deviations 0.3--3.0. This includes small unidirectional motion blur (e.g., shown in Fig.~1 middle panel), lens blur, and slight defocus blur that are very common in mobile photography. Since we approximate the inverse with a low-order polynomial, the algorithm fails gracefully without introducing jarring artifacts. The most severe failure case is due to kernel mismatch, in which case the algorithm either fails to remove all blur in the image, or over-sharpens the image. The halo removal is designed to handle the latter case (real example in Figure 6). Our experiments show that, in the context of mild blur, the algorithm produces similar or better results than other state-of-the-art image deblurring methods while being significantly faster. The whole deblurring process runs in a fraction of a second on a 12MP image on a modern mobile platform. Our method can be used to blindly correct blur before applying an off-the-shelf deep super-resolution model leading to superior results than other computationally demanding techniques.

\appendices

\section{Blur model validation and calibration}
\label{ap:model_calibration_assumptions}
The Gaussian blur estimation presented in Section~\ref{sec:model} is based on computing gradient features. 
To calibrate $c$ and $\sigma_b$, we proceed as follow. Given a set of 50 sharp high quality images, we simulate $K=1000$ random Gaussian blurry images, by randomly sampling the blur space and the image set. The Gaussian blur kernels are generated by sampling random values for $\sigma_0  \in [0.3, 4]$ and $\rho \in [0.15, 1]$. Additive Gaussian white noise of standard deviation 1\% is added to each simulated blurry image.

\begin{figure}
\footnotesize
    \centering
    \begin{minipage}[c]{.8\linewidth}
    \centering
    \includegraphics[clip,trim=0 0 0 110, width=.9\linewidth]{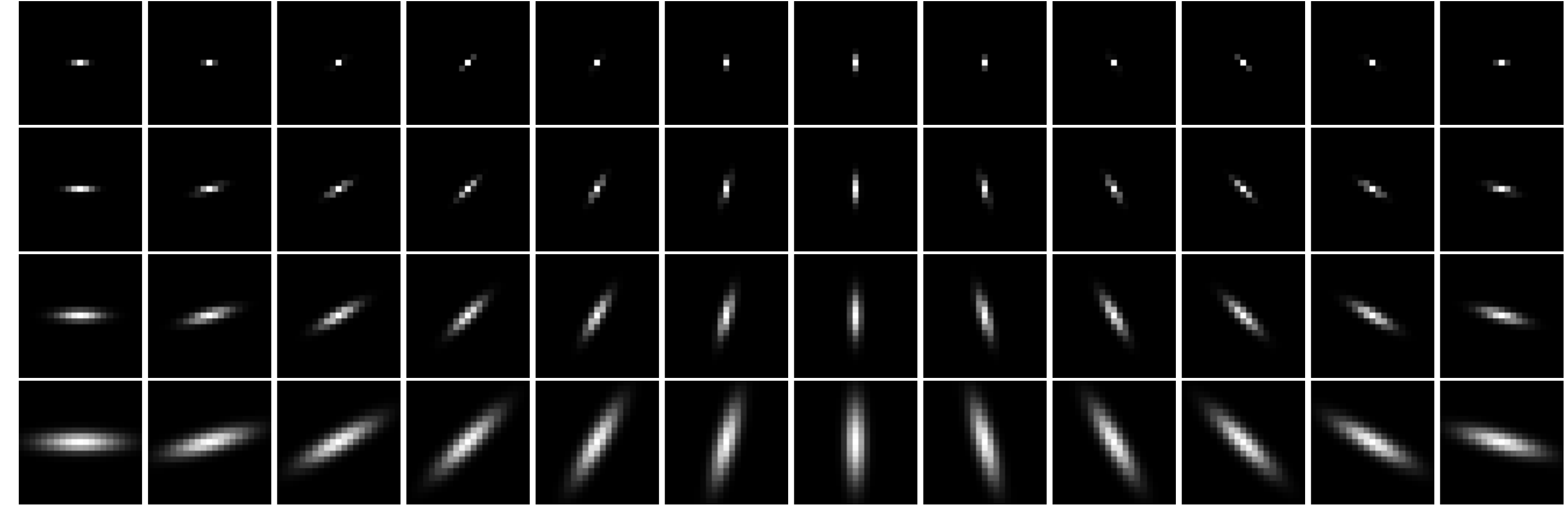}
    $\rho=0.25$
    \end{minipage} \vspace{.4em} %
    
    \begin{minipage}[c]{.8\linewidth}
    \centering
    \includegraphics[clip,trim=0 0 0 110,width=.9\linewidth]{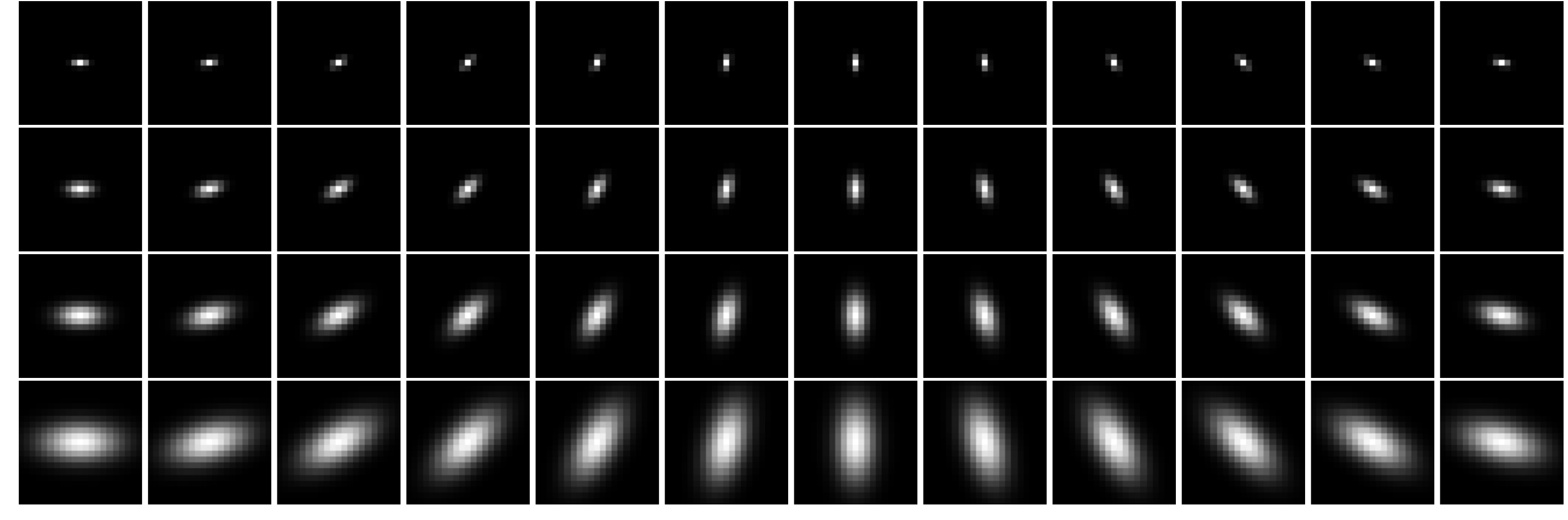}
    
    $\rho=0.50$
    \end{minipage} \vspace{.4em}
    
    \begin{minipage}[c]{.8\linewidth}
    \centering
    \includegraphics[clip,trim=0 0 0 110,width=.9\linewidth]{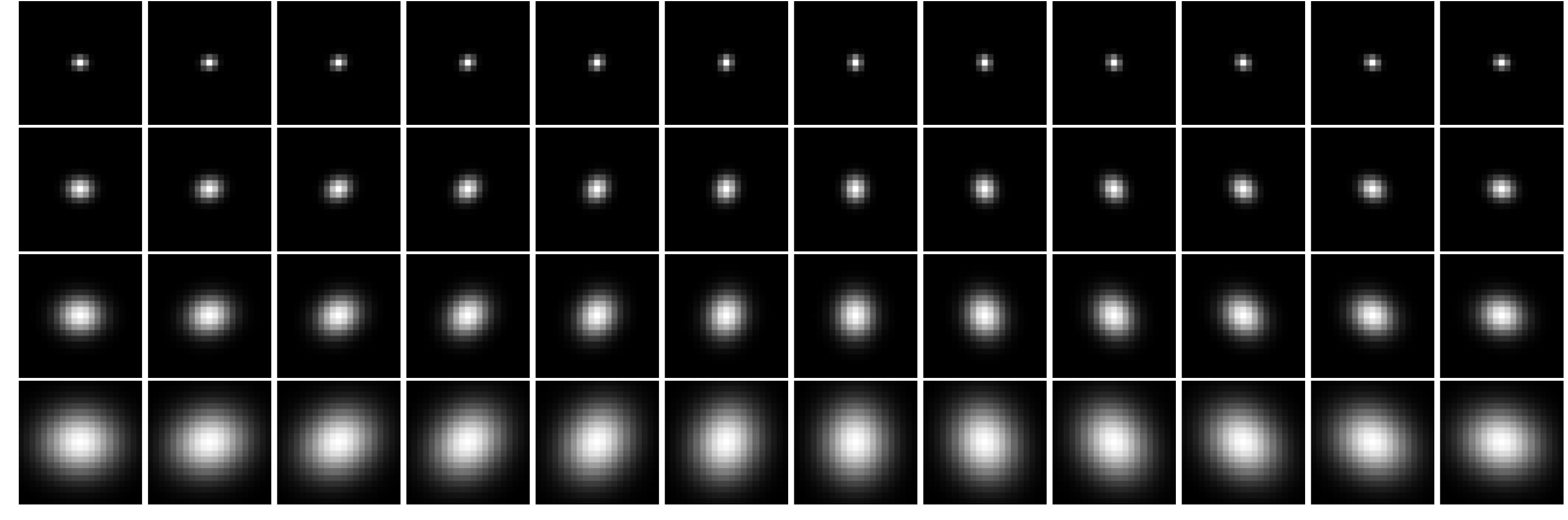}
    
    $\rho=0.85$
    \end{minipage} \vspace{.4em}
    
    \includegraphics[width=.8\linewidth, clip, trim=0 0 0 10]{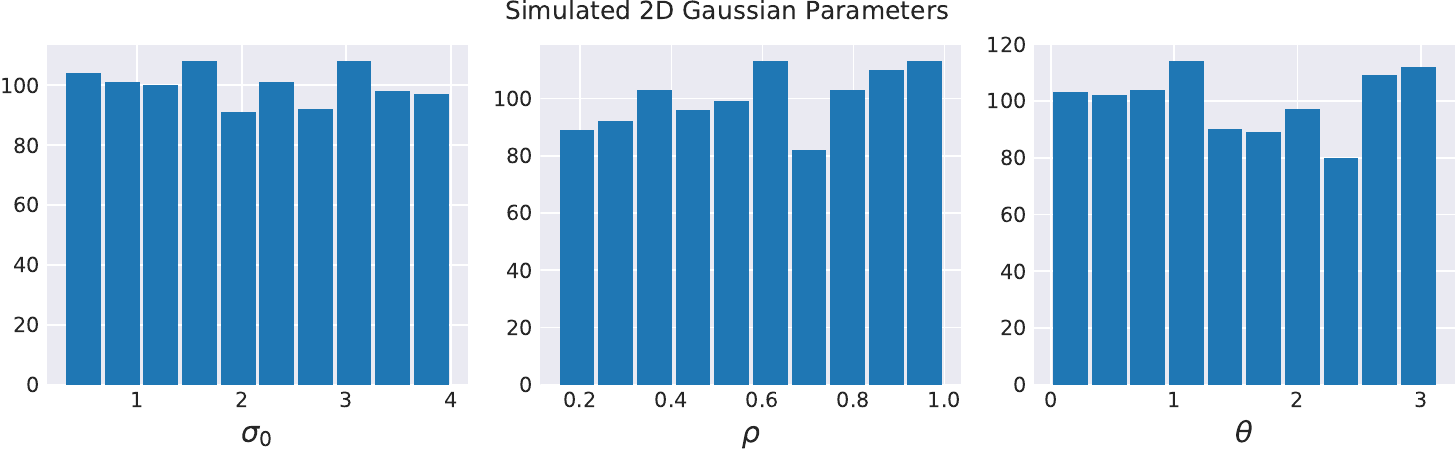}

    \caption{Gaussian Blur kernel examples and distribution of simulated parameters.}
    \label{fig:simulated_parameters}
\end{figure}

Examples of simulated Gaussian blur kernels are shown in Figure~\ref{fig:simulated_parameters}. 
For each of the blurry images we compute the gradient features according to~\eqref{eq:grad_feature}. The parameters $c$ and $\sigma_b$ are estimated by minimizing the mean absolute error (MAE).  The calibrated parameters are $c=89.8$ and $\sigma_b=0.764$. Note that the values of $c$ and $\sigma_b$ are implementation dependent (e.g., the finite difference scheme used to compute image gradient).

\begin{figure}[t]
    \centering

    \includegraphics[width=\linewidth]{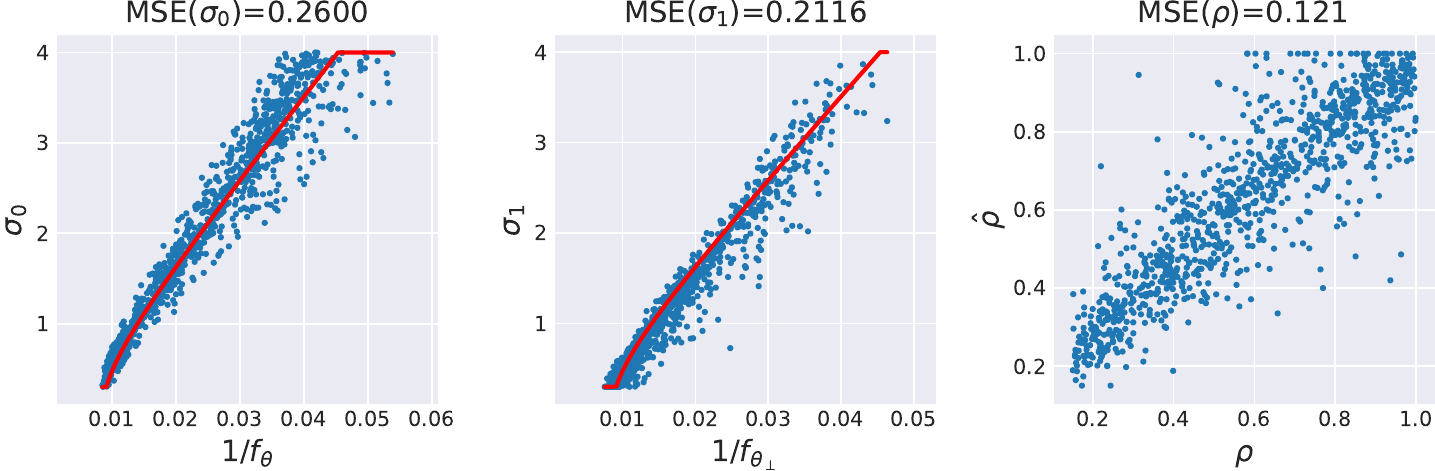}

    \caption{Gaussian blur model calibration. Error in the estimation of $\sigma_0$ and $\sigma_1$ (and $\rho$).}

    \label{fig:sigma_rho}

\end{figure}

Figure~\ref{fig:sigma_rho}-left shows the relation between the inverse of the estimated gradient feature (i.e., $1/f_\theta$) and the simulated blur kernel $\sigma_0$ value. Each of the blur points represents one simulated image. As we can see, sharp images that have very low blur values ($\sigma_0 \ll 1$) have very similar feature values. The same analysis holds for the gradient feature at the orthogonal direction, and its relation to $\sigma_1 = \rho \sigma_0$ (Figure~\ref{fig:sigma_rho}-middle). 
This validates our \emph{Assumption 1}.

Additionally, Figure~\ref{fig:sigma_rho} shows the almost linear relation (with some spread) between the inverse of the gradient feature $1/f_\theta$ and the blur level $\sigma_0$. This is exactly \emph{Assumption~2}.

\noindent Our model contemplates for slight blur that the gradient operator may have introduced when computing the gradient features ($\sigma_b$). This is further analyzed at the end of this appendix for a synthetic image.

In Figure~\ref{fig:sigma_rho}-right we show a plot of the real $\rho$ value of the simulated blur kernel and the estimated one. Although there are some outliers, the estimation is in general close to the real value (MSE($\rho$) = 0.121).

\vspace{.5em}

\begin{figure}[t]
    \centering

    \includegraphics[width=\linewidth]{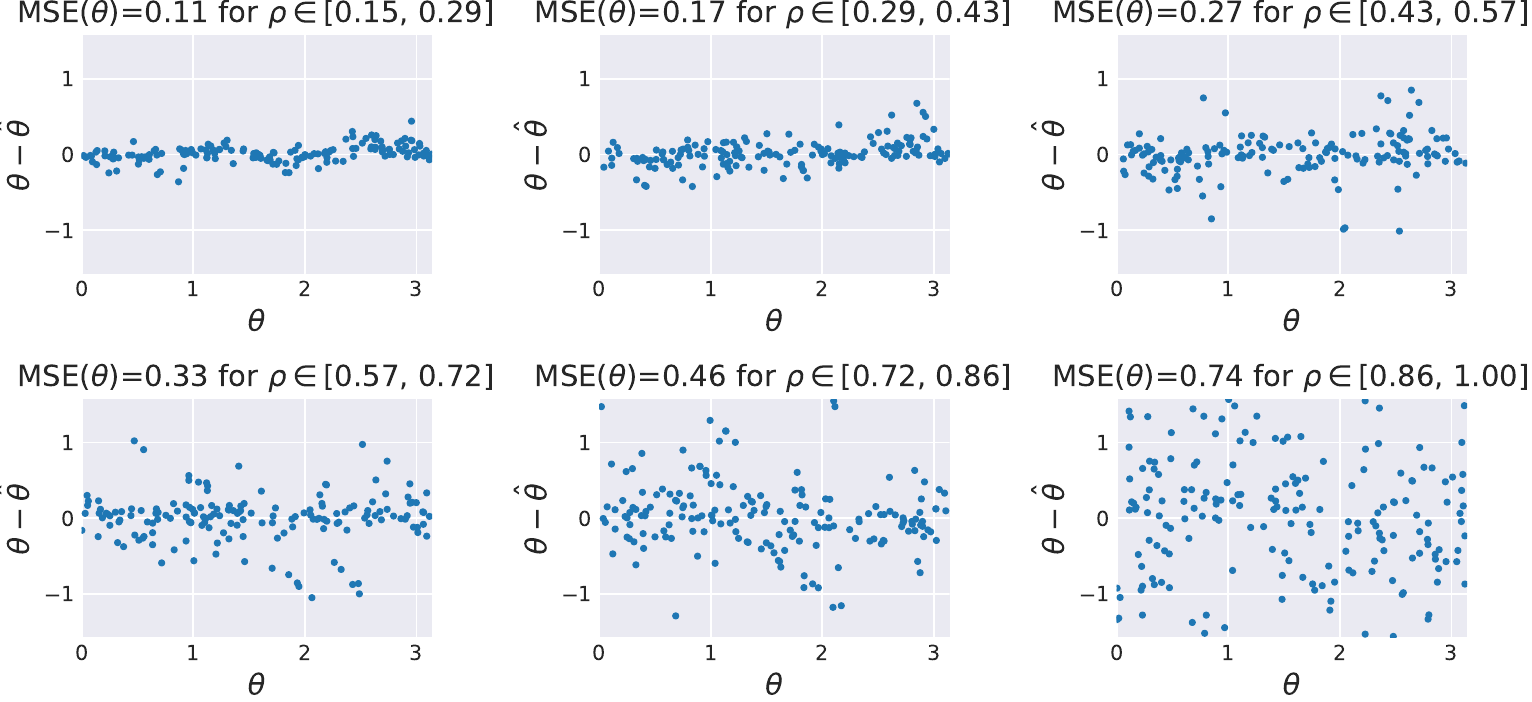}

    \caption{Blur model calibration. Error in estimated parameters. Calibrated parameters are $c=89.8, \sigma_b=0.764$.}
    \label{fig:theta_error}
\end{figure}

\noindent\textbf{Estimation of blur direction $\theta$}. The estimation of the blur direction is done by computing the angle $\theta$ with minimum gradient feature value in Eq.~\eqref{eq:grad_feature}. Figure~\ref{fig:theta_error} shows the error on the estimation of the angle for each simulated blur. As we can see, the error in the angle is quite low for blur kernels highly directional ($\rho < 0.5$). For large values of $\rho$ (e.g., $\rho \in [ 0.75, 1.0]$, the kernels are almost isotropic, and the estimation is inaccurate. Nevertheless, being almost isotropic, the kernel shape is not affected by the angle value in this case.

\begin{figure}[t]
\centering
\includegraphics[width=.8\linewidth]{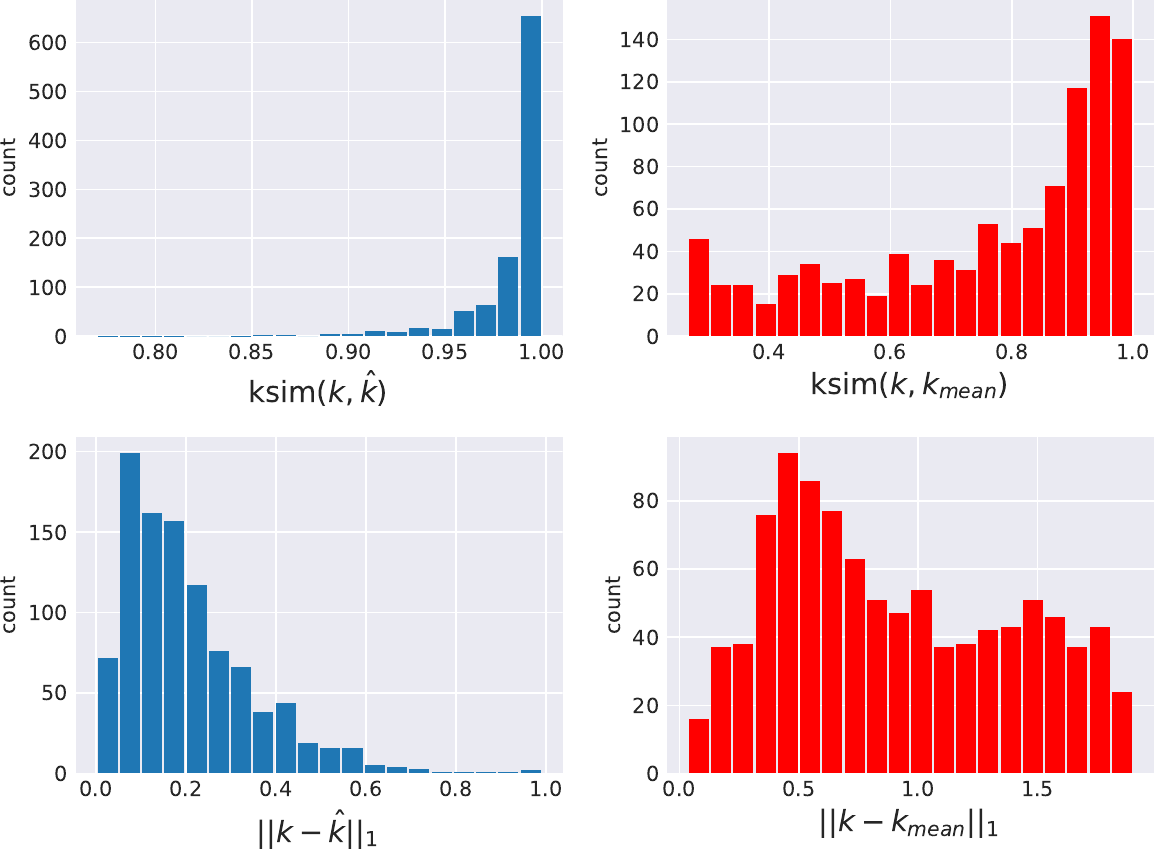}
\caption{Kernel error metrics on the simulated data. Top-row shows the histogram of the kernel similarity between estimated kernels and the respective ground-truth one (left), and between the ground-truth kernels and a fixed isotropic Gaussian kernel (denoted by $k_\text{mean}$) having standard deviation the mid-value on the simulated range (right). Bottom-row shows the histogram of the $\ell_1$ difference between the estimated kernels and the respective ground-truth one (left), and  between the ground-truth kernels and $k_\text{mean}$.}
\label{fig:kernel_error}
\end{figure}

\noindent \textbf{Error metrics on estimated kernel values.} The ultimate goal in blur estimation is to estimate a blur that is close to the real one. In Figure~\ref{fig:kernel_error} we present the distribution of two different error metrics that evaluate the distance between the estimated kernel and the simulated one directly on the kernel space. The first metric is the kernel similarity, which is the  normalized cross correlation:
\begin{equation}
\textrm{ksim}(k, \hat{k}) = \frac{1}{\|k\| \| \hat{k} \| }\sum_i k_i \hat{k}_i.
\end{equation}
When both kernels are equal the kernel similarity is 1. The second metric we compute is the $\ell_1$ norm between estimated kernel values.  To give an idea of the range of both metrics, we also present the error distribution of the real kernel and an average kernel\footnote{The average kernel is defined as the isotropic kernel having standard deviation the central value of the simulated range [0.35, 4.0].}. This shows the estimations are accurate.

\vspace{.5em}

\begin{figure}
\footnotesize
    \centering
    \includegraphics[width=.75\linewidth]{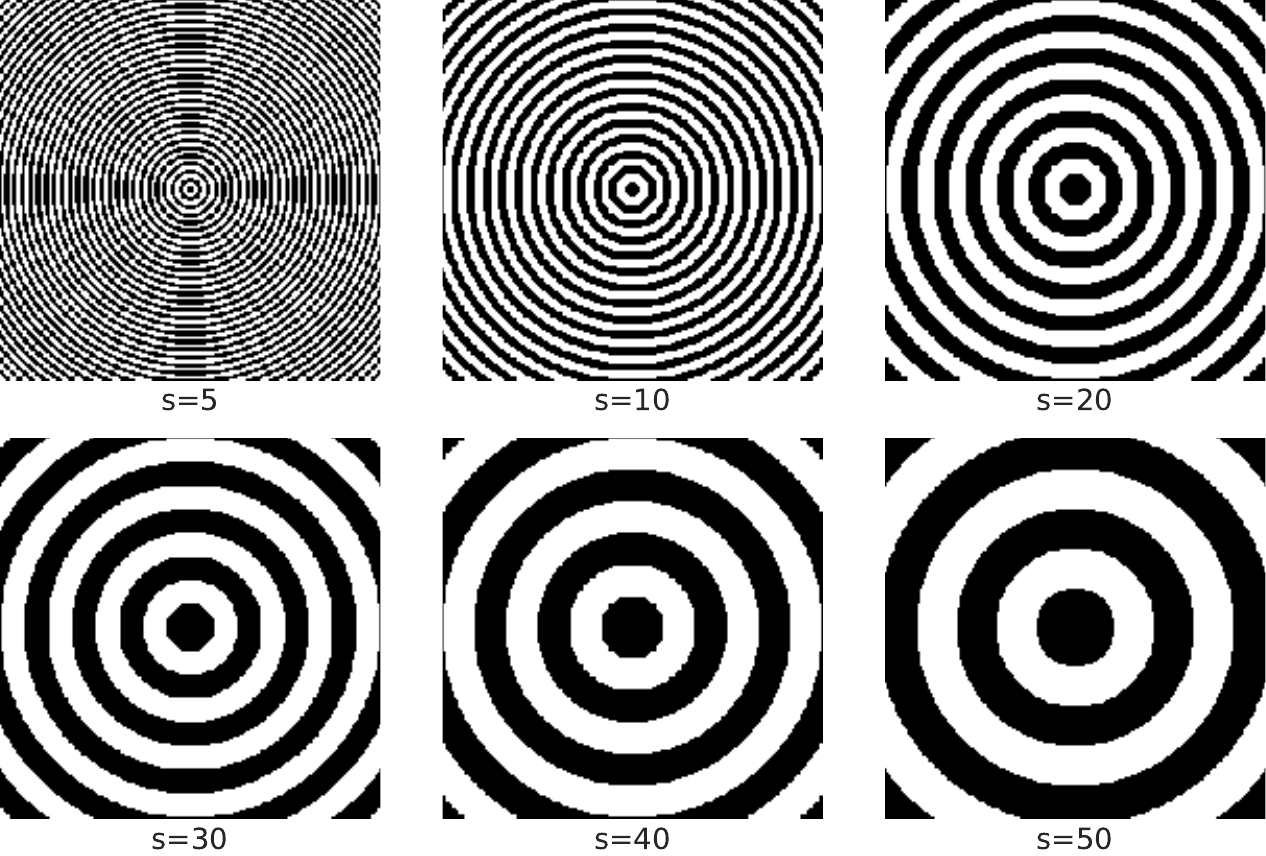}
    
    (a) Concentric circles pattern.
    \vspace{.5em}
    
   \includegraphics[width=.6\linewidth]{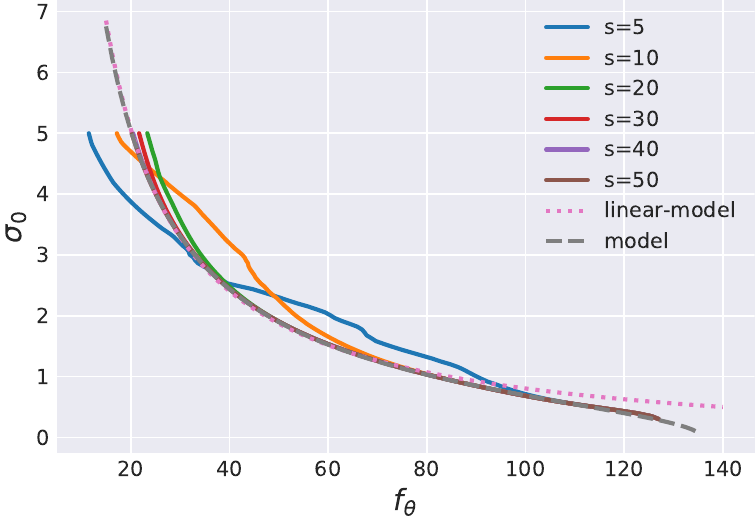}
    
    (b) Gradient feature $f_\theta$ as a function of simulated blur level $\sigma_0$.
    
    \caption{Blur model on concentric circles (synthetic image). }
    \label{fig:periodic_circles}
\end{figure}

\noindent \textbf{Concentric circles: A synthetic example.} We generated a family of synthetic images with concentric circles at different distances (controlled by a parameter $s \in [5,50]$, see Figure~\ref{fig:periodic_circles}a and blurred them with isotropic Gaussian blur of intensity ($\sigma_0 \in [0.3, 5]$). In Figure~\ref{fig:periodic_circles}b we show the computed gradient feature ($f_\theta$) for the different blurry images (different $\sigma_0$) on each simulated pattern ($s=5$ to $s=50$).
We calibrated the blur model (i.e., estimated $c$ and $\sigma_b$ values) using the $s=50$ synthetic pattern image.  The model is highly accurate when the circular rings are separated enough so the image can be locally considered a step-edge ($s\ge 20$). However, when the concentric circles are very close (e.g., $s=5$), the model is not accurate and the blur estimation is biased.  

For comparison purposes we also fit a purely linear model that directly maps the inverse of the gradient feature to the sigma value. The linear model (which is the proposed model with $\sigma_b=0$) is very close to the proposed model
except in low sigma values. This is because the proposed model takes into account the (very little) blur $\sigma_b$ that the gradient operator introduces, leading to a more precise estimation.

\section{Blur estimation implementation details}
\label{ap:blur_estimation_details}
In Algorithm~\ref{alg:blur_estimation} we present the pseudo-code for our estimation of Gaussian blur.  We follow Section~4 to estimate the blur. As the first step the input image is normalized using quantiles ($q=0.0001$ and $1-q$) to be robust to outliers. From the image gradient ($u_x, u_y$) we compute the directional derivative at $n_\text{angles}$ (usually $n_\text{angles}=6$) uniformly covering $[0, \pi)$. The maximum direction of the magnitude for each sample angle is found. Among the $n_\text{angles}$ maximum values, we find the minimum value and angle ($f_0, \theta_0)$ through bicubic interpolation. Using Eq.~\ref{eq:model_final} we compute $\sigma_0$ and $\sigma_1$ (and compute $\rho = \sigma_1/\sigma_0$).

The gradient features can be efficiently computed in parallel not just between different angles but also between each different pixel. Computing the maximum can be represented as a gather operation that can be optimized using shared memory and tiling. Within our pipeline this represents around $40-53\%$ of the computation, this comes from the fact that we compute a maximum magnitude $n_\text{angles}$ times.

\begin{algorithm}[t]
\SetAlgoLined
\SetKwData{Left}{left}\SetKwData{This}{this}\SetKwData{Up}{up}
\SetKwFunction{Union}{Union}\SetKwFunction{FindCompress}{FindCompress}
\SetKwInOut{Input}{input}\SetKwInOut{Output}{output}

 \Input{image $u$, $q$, $n_\text{angles}$, $\sigma_\text{max}, \sigma_\text{min}$, $\rho_\text{max}, \rho_\text{min}$  }
 \Output{Blur parameters: $\sigma_0, \rho, \theta$.}\vspace{.75em}
\text{\textbf{// 1. Compute gradient features}}\;
$n$ = Normalize($u,q$)\;
$u_x$,$u_y$ = ComputeGradient($n$, $\sigma_\text{max}$)\;
\For{$i \in [0, n_\text{angles}]$}{
 $\Psi_i$ = $i \cdot \pi / n_\text{angles}$\;
 $u_{\Psi_i} = u_x \cdot cos(\Psi_i) - u_y \cdot sin(\Psi_i)$\;
 $f_{\Psi_i}$ = Max($|u_{\Psi_i}|$)\;
}
$I_{\Psi}$ = Interpolate$(\{f_{\Psi}\})$\;
$f_{\theta},\theta_{0}$ = Min$(I_{\Psi})$\;
$f_{\theta_\perp},\theta_{0_\perp}$ = Ortho($f_{\theta},\theta_{0}$)\;\vspace{.75em}
\text{\textbf{// 2. Compute and clamp Gaussian}}\;
$\sigma_{0}, \sigma_{1}$ = Regression($f_{\theta}$ , $f_{\theta_\perp}$) // From Eq.~\eqref{eq:model_final}\;
Clamp($\sigma_{0}, \sigma_{1}, \sigma_\text{max}, \sigma_\text{min}$)\;
$\rho = \sigma_0 / \sigma_1$\;
Clamp($\rho, \rho_\text{min}, \rho_\text{max}$)\;
 \caption{Gaussian blur estimation. }
 \label{alg:blur_estimation}
\end{algorithm}

\section{Additional results}

\noindent \textbf{Comparison on DIV2K dataset.} In Figures \ref{fig:synt_results1}--\ref{fig:synt_results5} we present additional results to the one presented on Section~6. We generated a mild-blur dataset by artificially blurring sharp images from the DIV2K dataset~\cite{ntire2017}. Each of the 100 images in the validations dataset were blurred by a random Gaussian blur kernel of different sizes, shapes, and orientations (i.e., $\sigma_0  \sim \mathcal{U}[0.3, 4]$, $\rho \sim \mathcal{U}[0.15, 1.0]$, $\theta \sim \mathcal{U}[0, \pi]$). Additive white Gaussian noise of standard deviation $1\%$ was added on top.  We compared Polyblur (one, two, and three iterations) to the following adaptive sharpening or deblurring methods: SRN-Deblur~\cite{tao2018scale}, Sparse Deblurring~\cite{zhang2013multi}, DeblurGANv2 (inception and mobilenet architectures)~\cite{kupyn2019deblurgan}, Spectral Irregularities~\cite{goldstein2012blur}, L0-Deblur~\cite{pan2016l0}, GLAS~\cite{zhu2011restoration}, Guided Filter~\cite{he2012guided}, Convolutional Deblurring~\cite{hosseini2019convolutional}. 

\definecolor{red1}{rgb}{0.8431    0.1882    0.1529}
\definecolor{orange1}{rgb}{0.9882    0.5529    0.3490}
\definecolor{yellow1}{rgb}{0.9961    0.8784    0.5451}
\definecolor{blue1}{rgb}{0.3000    0.3000    0.8490}
\definecolor{green2}{rgb}{0.2    0.8    0.3}
\definecolor{green1}{rgb}{0.1020    0.5961    0.3137}
\definecolor{red2}{rgb}{0.8431    0.1882    0.1529}

\begin{figure*}
\footnotesize
    \centering
     \begin{tikzpicture}
      \node[anchor=north west,inner sep=0] (image) at (0,0) {\includegraphics[clip, trim=0 70 0 70, width=.495\linewidth]{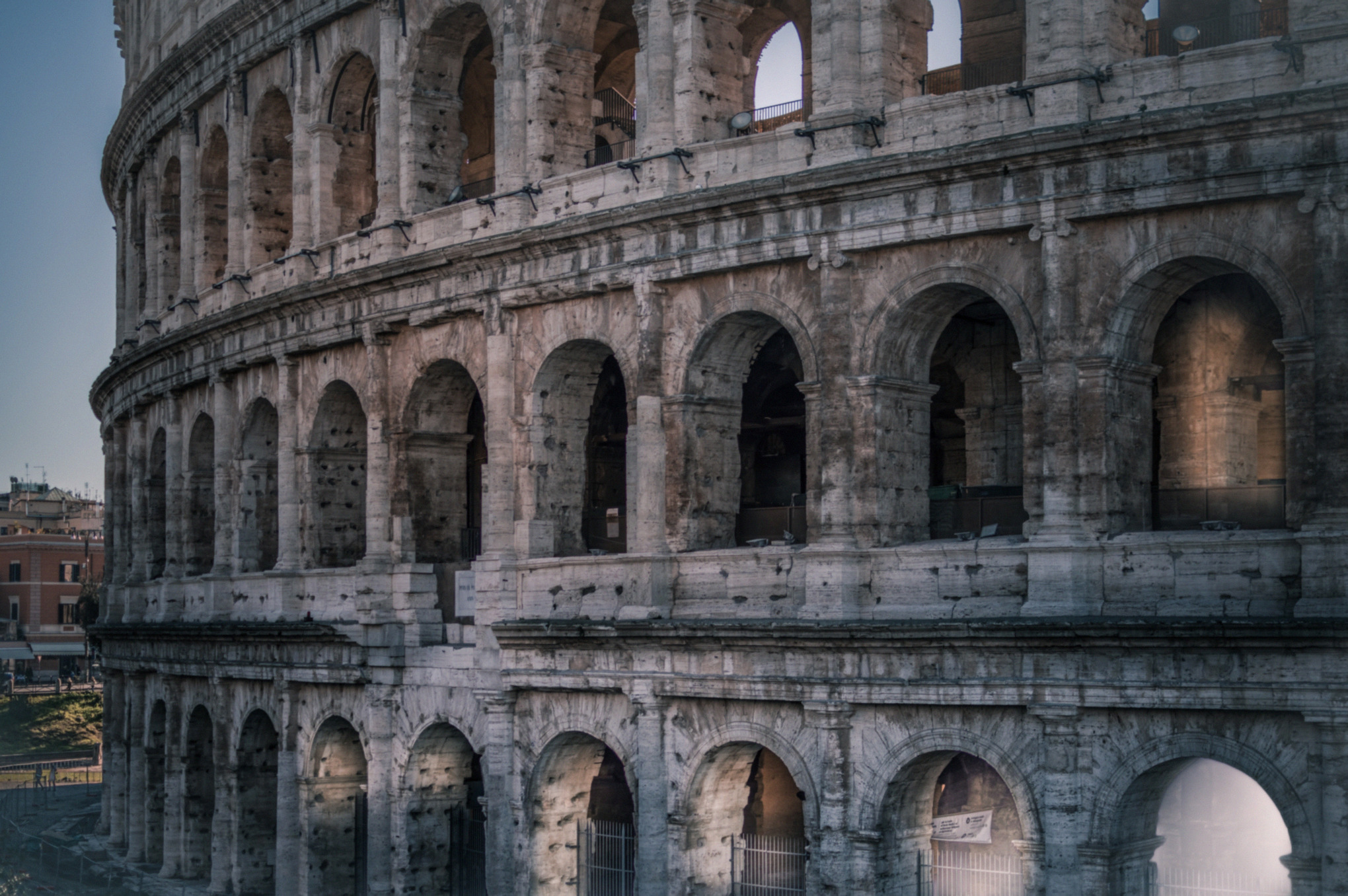}};
      \begin{scope}[x={(image.north east)},y={(image.south west)}]
        \draw[red1, very thick] (0.4387, 0.3766) rectangle (0.5613, 0.6234);
       \node[] at (0.0525,0.9625) {\begin{color}{white}Input\end{color}};
      \end{scope}
     \end{tikzpicture}
      \begin{overpic}[width=.245\linewidth]{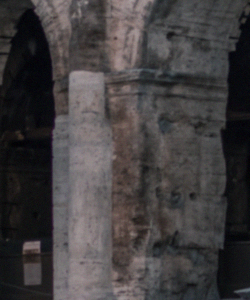}
        \put(3,3){\begin{color}{white}Input\end{color}}
        
      \end{overpic}
     \begin{overpic}[width=.245\linewidth]{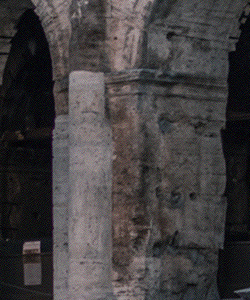}
        \put(3,3){\begin{color}{white}Conv. Deblurring\end{color}}
      \end{overpic}\vspace{.15em}
      
     \begin{overpic}[width=.245\linewidth]{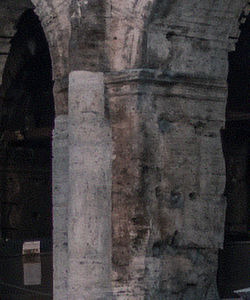}
        \put(3,3){\begin{color}{white}L0-Deblur\end{color}}
      \end{overpic} 
     \begin{overpic}[width=.245\linewidth]{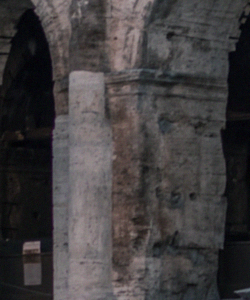}
        \put(3,3){\begin{color}{white}DeblurGANv2-inception\end{color}}
      \end{overpic}      
     \begin{overpic}[width=.245\linewidth]{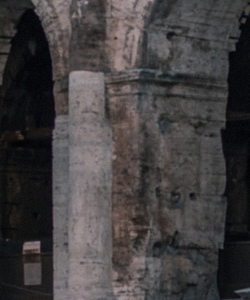}
        \put(3,3){\begin{color}{white}DeblurGANv2-mobilenet\end{color}}
      \end{overpic}     
     \begin{overpic}[width=.245\linewidth]{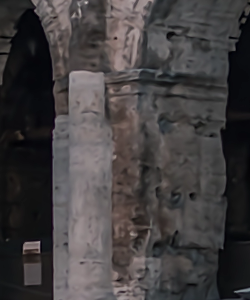}
        \put(3,3){\begin{color}{white}GLAS\end{color}}
      \end{overpic}\vspace{.15em}
      
     \begin{overpic}[width=.245\linewidth]{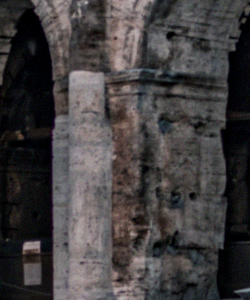}
        \put(3,3){\begin{color}{white}Guided Filter\end{color}}
      \end{overpic}                 
     \begin{overpic}[width=.245\linewidth]{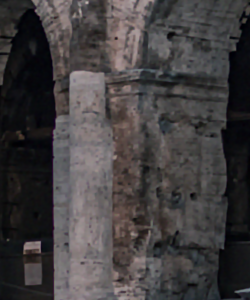}
        \put(3,3){\begin{color}{white}SparseDeblur\end{color}}
      \end{overpic}         
     \begin{overpic}[width=.245\linewidth]{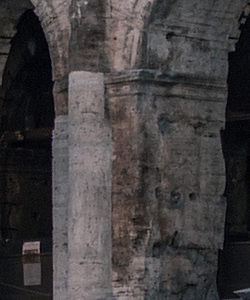}
        \put(3,3){\begin{color}{white}Spectral Irreg\end{color}}
      \end{overpic}   
     \begin{overpic}[width=.245\linewidth]{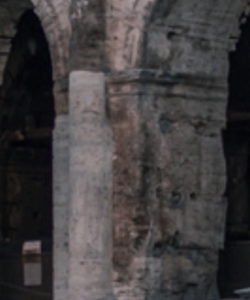}
        \put(3,3){\begin{color}{white}SRN-Deblur\end{color}}
      \end{overpic}\vspace{.15em}
      
     \begin{overpic}[width=.245\linewidth]{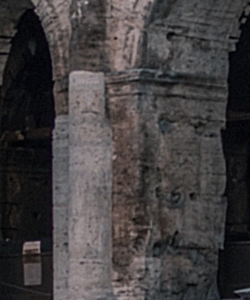}
        \put(3,3){\begin{color}{white}Polyblur-it1\end{color}}
      \end{overpic}         
     \begin{overpic}[width=.245\linewidth]{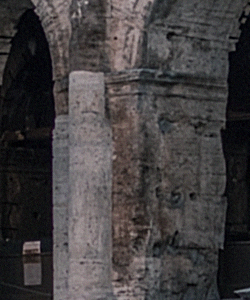}
        \put(3,3){\begin{color}{white}Polyblur-it2\end{color}}
      \end{overpic}   
     \begin{overpic}[width=.245\linewidth]{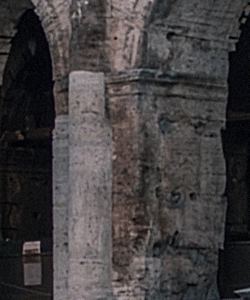}
        \put(3,3){\begin{color}{white}Polyblur-it3\end{color}}
      \end{overpic}   
     \begin{overpic}[width=.245\linewidth]{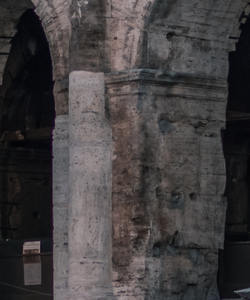}
        \put(3,3){\begin{color}{white}Reference\end{color}}
      \end{overpic}         
    \caption{Example of comparison on one image from the DIV2K validation dataset with synthetic blur.}
    \label{fig:synt_results1}
\end{figure*}

\begin{figure*}
\footnotesize
    \centering
   \begin{tikzpicture}
      \node[anchor=north west,inner sep=0] (image) at (0,0) {\includegraphics[clip, trim=0 50 0 80, width=.495\linewidth]{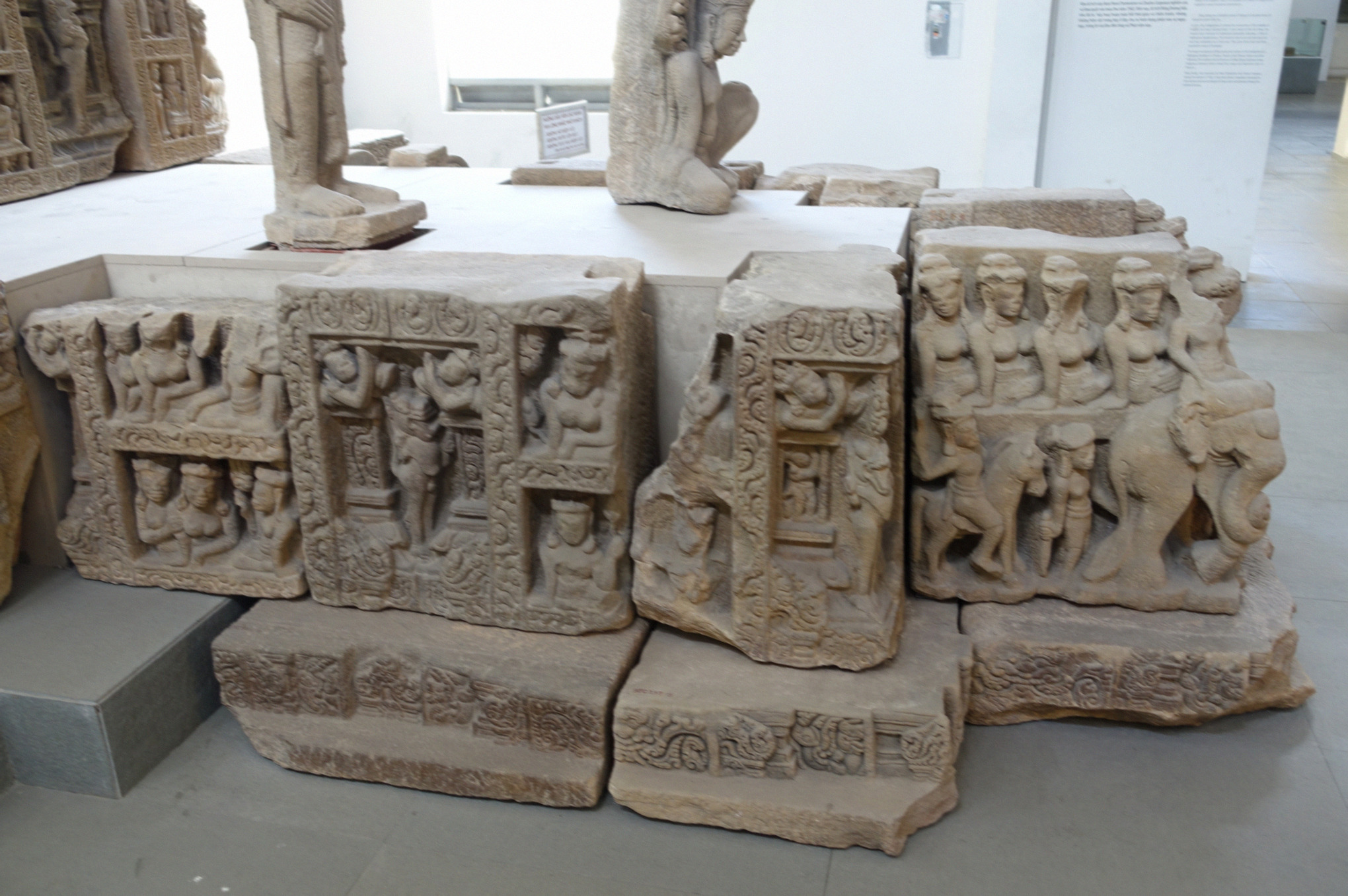}};
      \begin{scope}[x={(image.north east)},y={(image.south west)}]
       \draw[red1, very thick] (0.5368, 0.5285) rectangle (0.6593, 0.7732);
       \node[] at (0.0525,0.9625) {\begin{color}{white}Input\end{color}};
      \end{scope}
     \end{tikzpicture}
      \begin{overpic}[width=.245\linewidth]{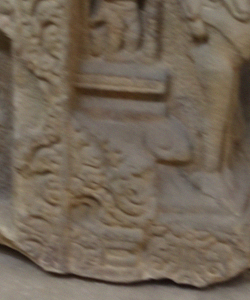}
        \put(3,3){\begin{color}{white}Input\end{color}}
      \end{overpic}
     \begin{overpic}[width=.245\linewidth]{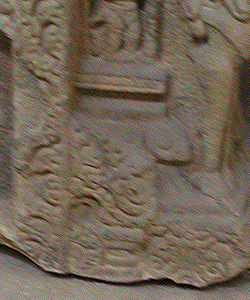}
        \put(3,3){\begin{color}{white}Conv. Deblurring\end{color}}
      \end{overpic}\vspace{.15em}
      
     \begin{overpic}[width=.245\linewidth]{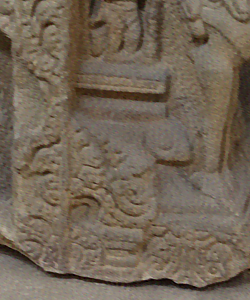}
        \put(3,3){\begin{color}{white}L0-Deblur\end{color}}
      \end{overpic} 
     \begin{overpic}[width=.245\linewidth]{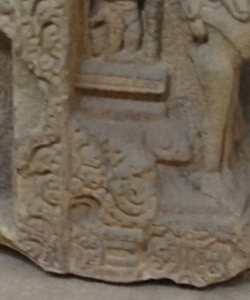}
        \put(3,3){\begin{color}{white}DeblurGANv2-inception\end{color}}
      \end{overpic}      
     \begin{overpic}[width=.245\linewidth]{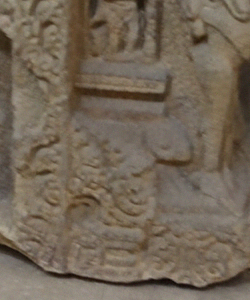}
        \put(3,3){\begin{color}{white}DeblurGANv2-mobilenet\end{color}}
      \end{overpic}     
     \begin{overpic}[width=.245\linewidth]{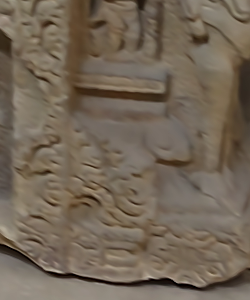}
        \put(3,3){\begin{color}{white}GLAS\end{color}}
      \end{overpic}\vspace{.15em}
      
     \begin{overpic}[width=.245\linewidth]{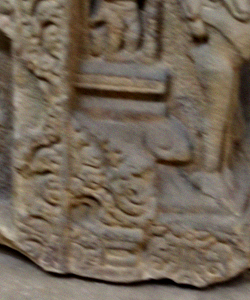}
        \put(3,3){\begin{color}{white}Guided Filter\end{color}}
      \end{overpic}                 
     \begin{overpic}[width=.245\linewidth]{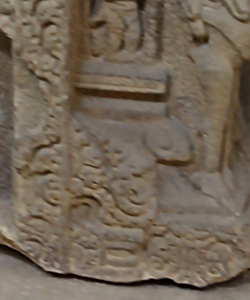}
        \put(3,3){\begin{color}{white}SparseDeblur\end{color}}
      \end{overpic}         
     \begin{overpic}[width=.245\linewidth]{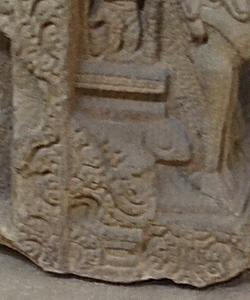}
        \put(3,3){\begin{color}{white}Spectral Irreg\end{color}}
      \end{overpic}   
     \begin{overpic}[width=.245\linewidth]{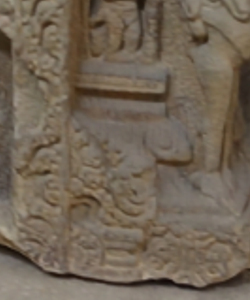}
        \put(3,3){\begin{color}{white}SRN-Deblur\end{color}}
      \end{overpic}\vspace{.15em}
      
     \begin{overpic}[width=.245\linewidth]{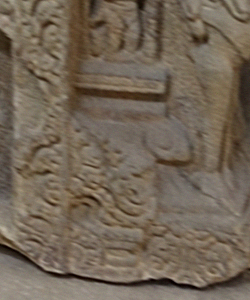}
        \put(3,3){\begin{color}{white}Polyblur-it1\end{color}}
      \end{overpic}         
     \begin{overpic}[width=.245\linewidth]{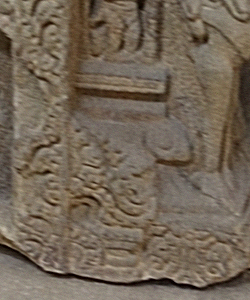}
        \put(3,3){\begin{color}{white}Polyblur-it2\end{color}}
      \end{overpic}   
     \begin{overpic}[width=.245\linewidth]{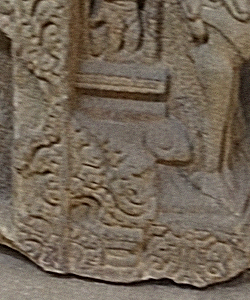}
        \put(3,3){\begin{color}{white}Polyblur-it3\end{color}}
      \end{overpic}   
     \begin{overpic}[width=.245\linewidth]{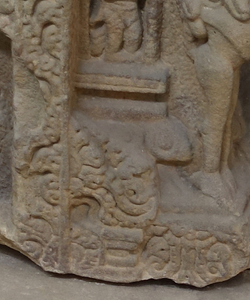}
        \put(3,3){\begin{color}{white}Reference\end{color}}
      \end{overpic}         
    \caption{Example of comparison on one image from the DIV2K validation dataset with synthetic blur.}
    \label{fig:synt_results2}
\end{figure*}

\begin{figure*}
\footnotesize
    \centering
     \begin{tikzpicture}
      \node[anchor=north west,inner sep=0] (image) at (0,0) {\includegraphics[clip, trim=0 520 0 720, width=.495\linewidth]{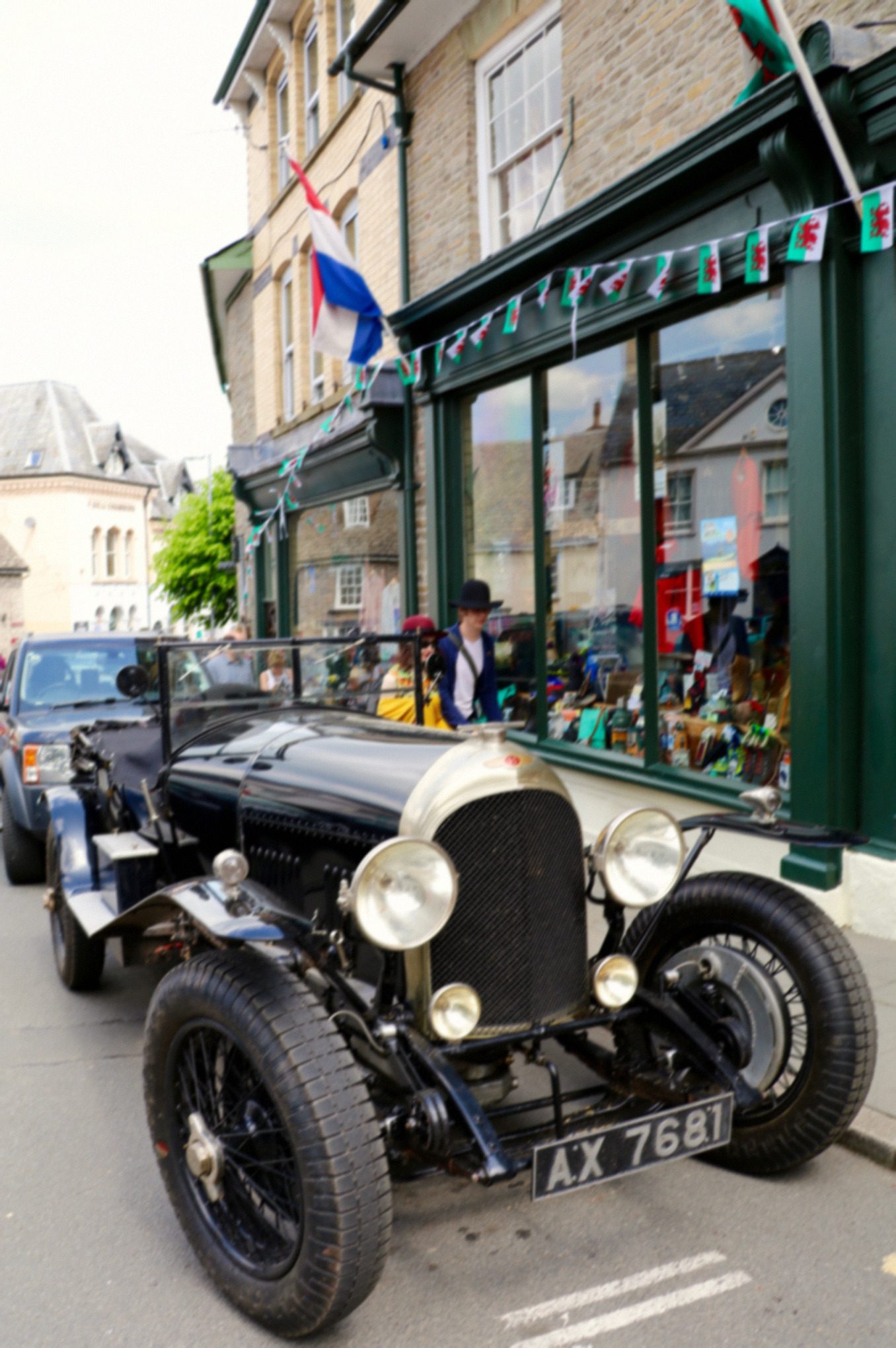}};
      \begin{scope}[x={(image.north east)},y={(image.south west)}]
       \draw[red, very thick] (0.3982, 0.1450) rectangle (0.5826, 0.5200);
       \node[] at (0.05,0.96) {\begin{color}{arsenic}Input\end{color}};
       \node[] at (0.0525,0.9625) {\begin{color}{white}Input\end{color}};
      \end{scope}
     \end{tikzpicture}
      \begin{overpic}[width=.245\linewidth]{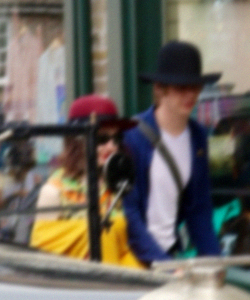}
        \put(3,3){\begin{color}{arsenic}Input\end{color}}
        \put(3.5,2.5){\begin{color}{white}Input\end{color}}
      \end{overpic}
     \begin{overpic}[width=.245\linewidth]{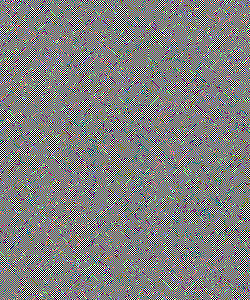}
        \put(3,3){\begin{color}{arsenic}Conv. Deblurring\end{color}}
        \put(3.5,2.5){\begin{color}{white}Conv. Deblurring\end{color}}
      \end{overpic}\vspace{.15em}
      
     \begin{overpic}[width=.245\linewidth]{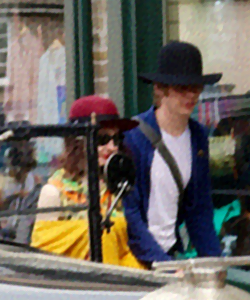}
        \put(3,3){\begin{color}{arsenic}L0-Deblur\end{color}}
        \put(3.5,2.5){\begin{color}{white}L0-Deblur\end{color}}
      \end{overpic} 
     \begin{overpic}[width=.245\linewidth]{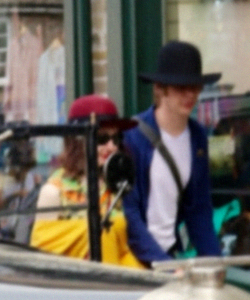}
        \put(3,3){\begin{color}{arsenic}DeblurGANv2-inception\end{color}}
        \put(3.5,2.5){\begin{color}{white}DeblurGANv2-inception\end{color}}
      \end{overpic}      
     \begin{overpic}[width=.245\linewidth]{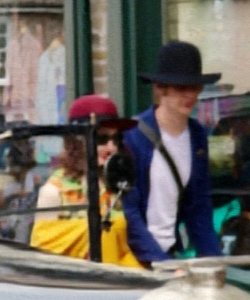}
        \put(3,3){\begin{color}{arsenic}DeblurGANv2-mobilenet\end{color}}
        \put(3.5,2.5){\begin{color}{white}DeblurGANv2-mobilenet\end{color}}
      \end{overpic}     
     \begin{overpic}[width=.245\linewidth]{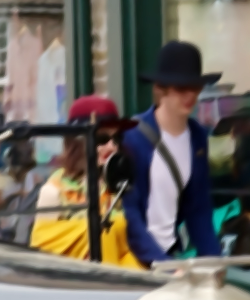}
        \put(3,3){\begin{color}{arsenic}GLAS\end{color}}
        \put(3.5,2.5){\begin{color}{white}GLAS\end{color}}
      \end{overpic}\vspace{.15em}
      
     \begin{overpic}[width=.245\linewidth]{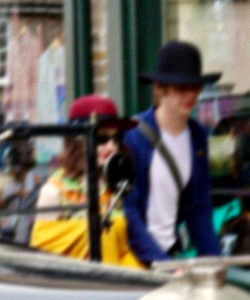}
        \put(3,3){\begin{color}{arsenic}Guided Filter\end{color}}
        \put(3.5,2.5){\begin{color}{white}Guided Filter\end{color}}
      \end{overpic}                 
     \begin{overpic}[width=.245\linewidth]{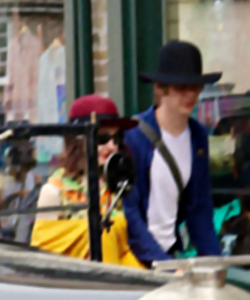}
        \put(3,3){\begin{color}{arsenic}SparseDeblur\end{color}}
        \put(3.5,2.5){\begin{color}{white}SparseDeblur\end{color}}
      \end{overpic}         
     \begin{overpic}[width=.245\linewidth]{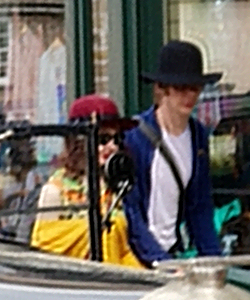}
        \put(3,3){\begin{color}{arsenic}Spectral Irreg\end{color}}
        \put(3.5,2.5){\begin{color}{white}Spectral Irreg\end{color}}
      \end{overpic}   
     \begin{overpic}[width=.245\linewidth]{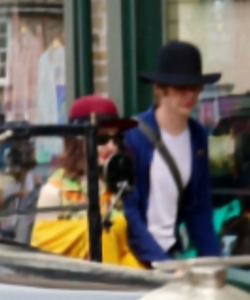}
        \put(3,3){\begin{color}{arsenic}SRN-Deblur\end{color}}
        \put(3.5,2.5){\begin{color}{white}SRN-Deblur\end{color}}
      \end{overpic}\vspace{.15em}
      
     \begin{overpic}[width=.245\linewidth]{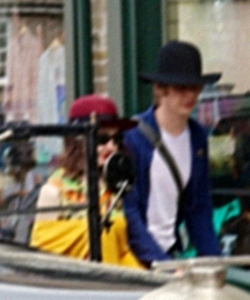}
        \put(3,3){\begin{color}{arsenic}Polyblur-it1\end{color}}
        \put(3.5,2.5){\begin{color}{white}Polyblur-it1\end{color}}
      \end{overpic}         
     \begin{overpic}[width=.245\linewidth]{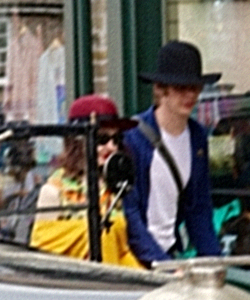}
        \put(3,3){\begin{color}{arsenic}Polyblur-it2\end{color}}
        \put(3.5,2.5){\begin{color}{white}Polyblur-it2\end{color}}
      \end{overpic}   
     \begin{overpic}[width=.245\linewidth]{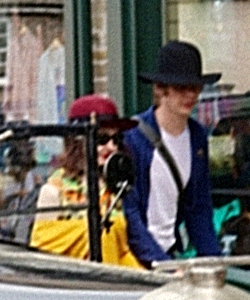}
        \put(3,3){\begin{color}{arsenic}Polyblur-it3\end{color}}
        \put(3.5,2.5){\begin{color}{white}Polyblur-it3\end{color}}
      \end{overpic}   
     \begin{overpic}[width=.245\linewidth]{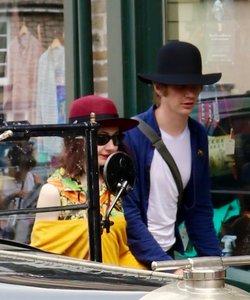}
        \put(3,3){\begin{color}{arsenic}Reference\end{color}}
        \put(3.5,2.5){\begin{color}{white}Reference\end{color}}
      \end{overpic}         
    \caption{Example of comparison on one image from the DIV2K validation dataset with synthetic blur.}
    \label{fig:synt_results3}
\end{figure*}

\begin{figure*}
\footnotesize
    \centering
     \begin{tikzpicture}
      \node[anchor=north west,inner sep=0] (image) at (0,0) {\includegraphics[clip, trim=0 140 0 0, width=.495\linewidth]{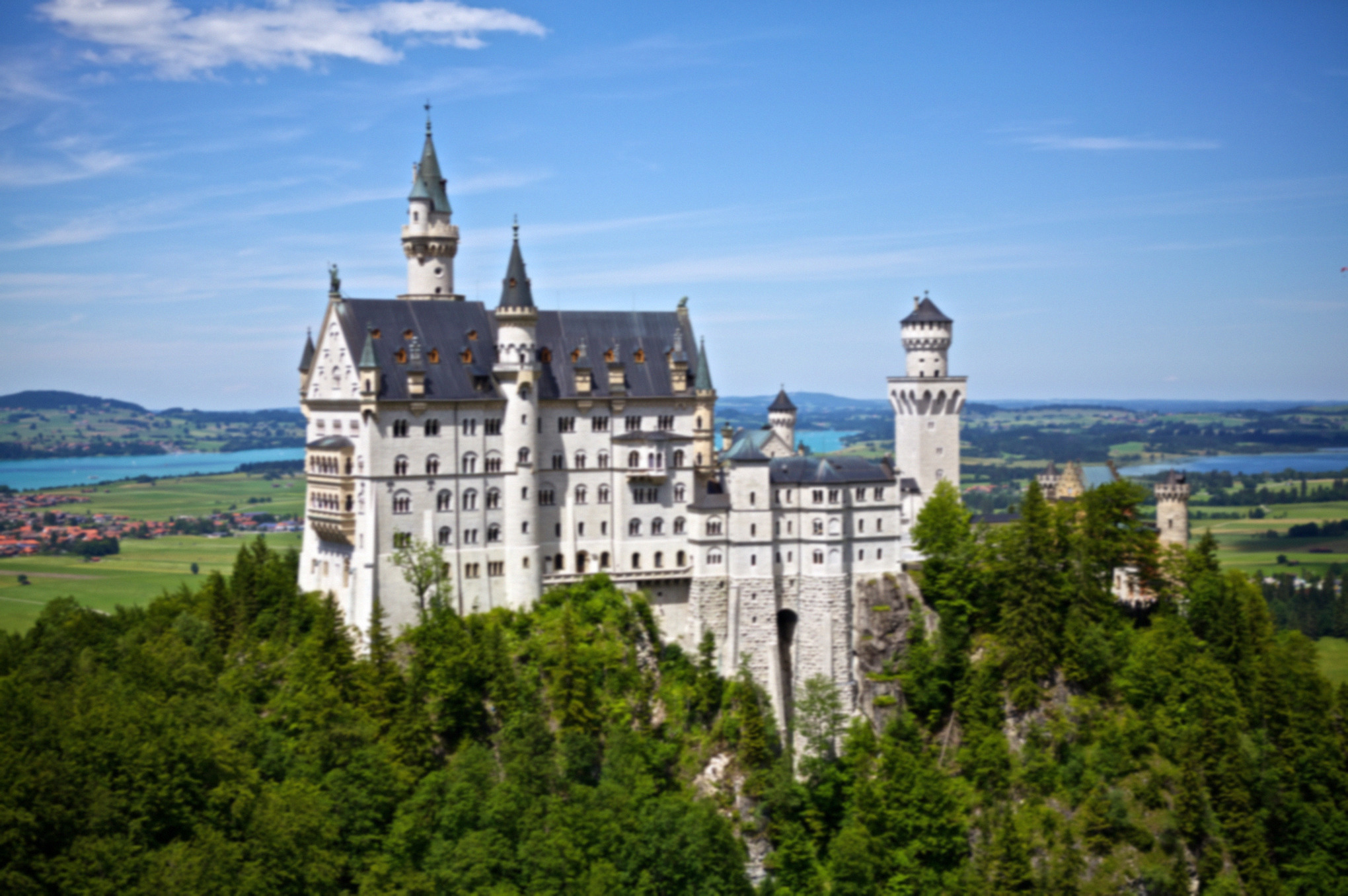}};
      \begin{scope}[x={(image.north east)},y={(image.south west)}]
       \draw[red1, very thick] (0.1422, 0.4441) rectangle (0.2647, 0.6908);  
        \node[] at (0.05,0.96) {\begin{color}{white}Input\end{color}};
       \end{scope}
     \end{tikzpicture}    
      \begin{overpic}[width=.245\linewidth]{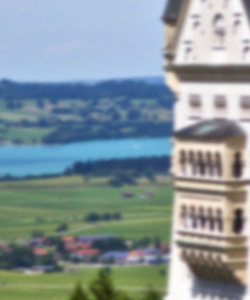}
        \put(3,3){\begin{color}{white}Input\end{color}}
      \end{overpic}
     \begin{overpic}[width=.245\linewidth]{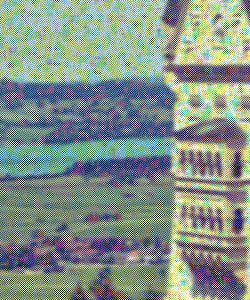}
        \put(3,3){\begin{color}{white}Conv. Deblurring\end{color}}
      \end{overpic}\vspace{.15em}
      
     \begin{overpic}[width=.245\linewidth]{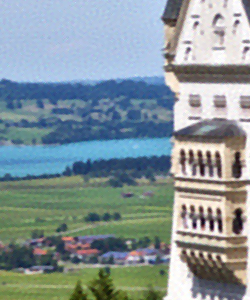}
        \put(3,3){\begin{color}{white}L0-Deblur\end{color}}
      \end{overpic} 
     \begin{overpic}[width=.245\linewidth]{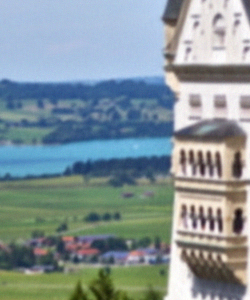}
        \put(3,3){\begin{color}{white}DeblurGANv2-inception\end{color}}
      \end{overpic}      
     \begin{overpic}[width=.245\linewidth]{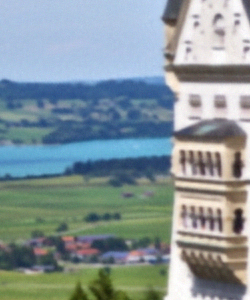}
        \put(3,3){\begin{color}{white}DeblurGANv2-mobilenet\end{color}}
      \end{overpic}     
     \begin{overpic}[width=.245\linewidth]{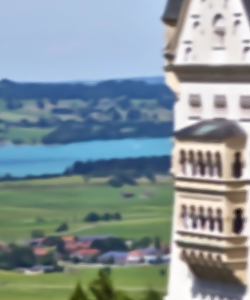}
        \put(3,3){\begin{color}{white}GLAS\end{color}}
      \end{overpic}\vspace{.15em}
      
     \begin{overpic}[width=.245\linewidth]{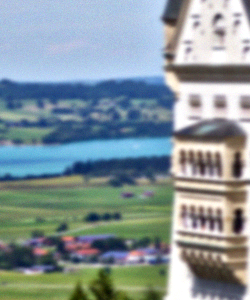}
        \put(3,3){\begin{color}{white}Guided Filter\end{color}}
      \end{overpic}                 
     \begin{overpic}[width=.245\linewidth]{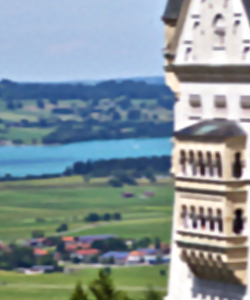}
        \put(3,3){\begin{color}{white}SparseDeblur\end{color}}
      \end{overpic}         
     \begin{overpic}[width=.245\linewidth]{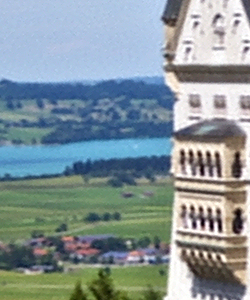}
        \put(3,3){\begin{color}{white}Spectral Irreg\end{color}}
      \end{overpic}   
     \begin{overpic}[width=.245\linewidth]{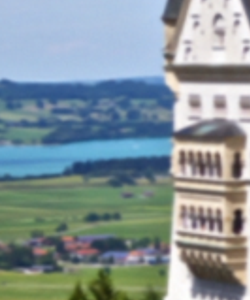}
        \put(3,3){\begin{color}{white}SRN-Deblur\end{color}}
      \end{overpic}\vspace{.15em}
      
     \begin{overpic}[width=.245\linewidth]{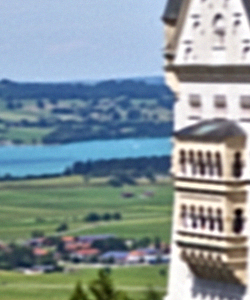}
        \put(3,3){\begin{color}{white}Polyblur-it1\end{color}}
      \end{overpic}         
     \begin{overpic}[width=.245\linewidth]{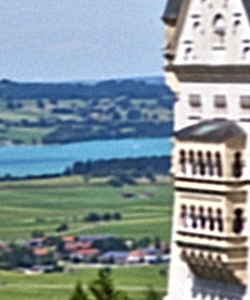}
        \put(3,3){\begin{color}{white}Polyblur-it2\end{color}}
      \end{overpic}   
     \begin{overpic}[width=.245\linewidth]{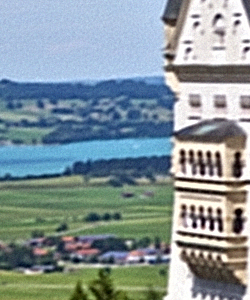}
        \put(3,3){\begin{color}{white}Polyblur-it3\end{color}}
      \end{overpic}   
     \begin{overpic}[width=.245\linewidth]{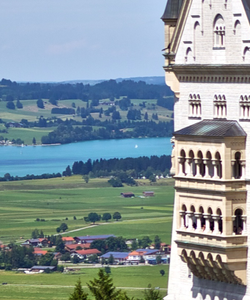}
        \put(3,3){\begin{color}{white}Reference\end{color}}
      \end{overpic}         
    \caption{Example of comparison on one image from the DIV2K validation dataset with synthetic blur.}
    \label{fig:synt_results4}
\end{figure*}

\begin{figure*}
\footnotesize
    \centering
    \begin{tikzpicture}
      \node[anchor=north west,inner sep=0] (image) at (0,0) {\includegraphics[clip, trim=0 1050 0 100, width=.495\linewidth]{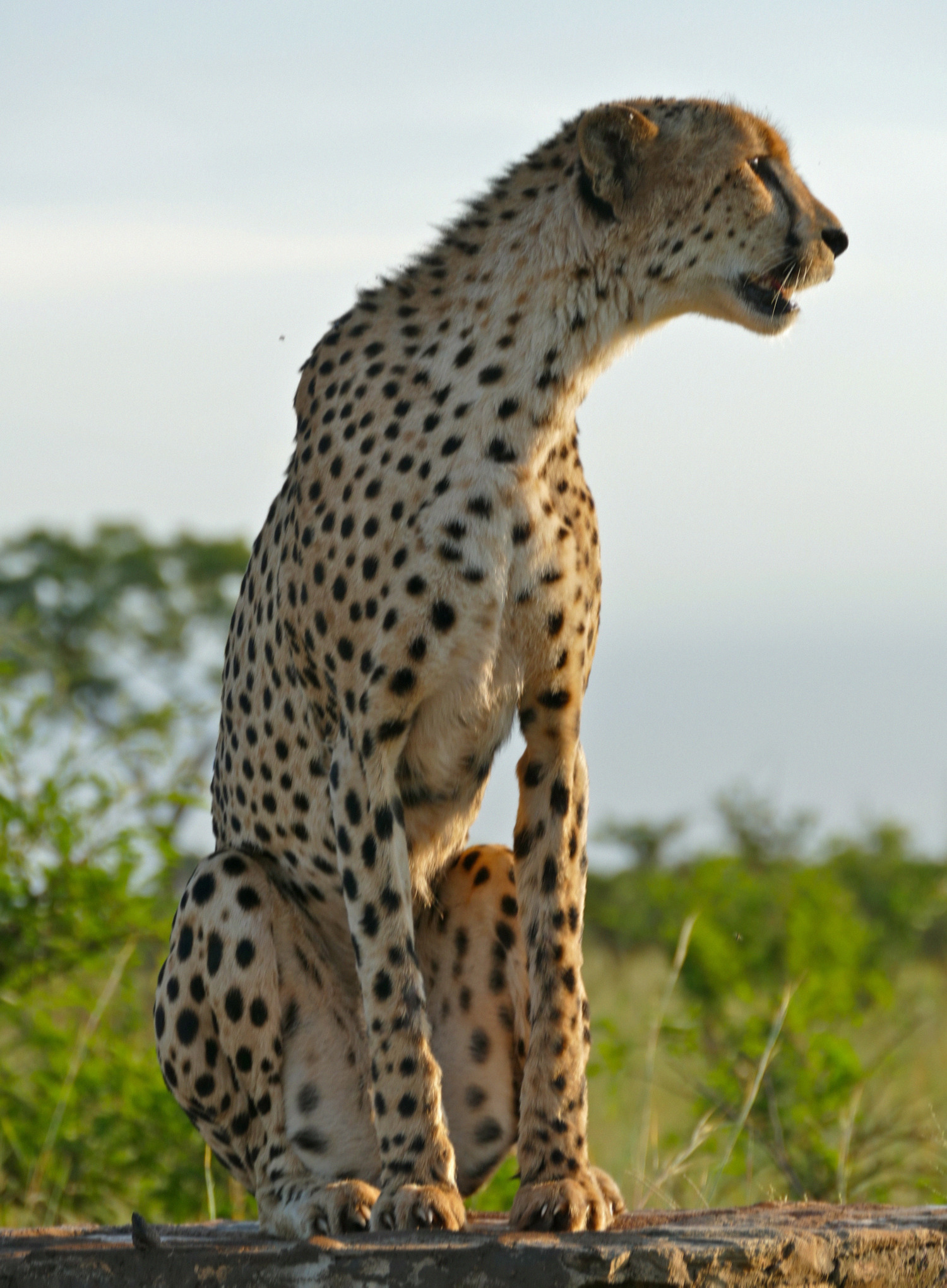}};
      \begin{scope}[x={(image.north east)},y={(image.south west)}]
         \draw[red1, very thick] (0.6900, 0.0607) rectangle (0.8567, 0.3978);
         \node[] at (0.05,0.96) {\begin{color}{white}Input\end{color}};
       \end{scope}
     \end{tikzpicture}    
      \begin{overpic}[width=.245\linewidth]{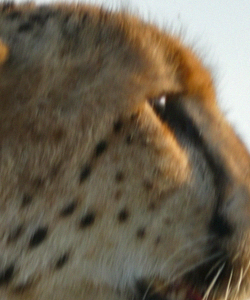}
        \put(3,3){\begin{color}{white}Input\end{color}}
      \end{overpic}
     \begin{overpic}[width=.245\linewidth]{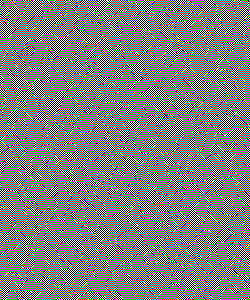}
        \put(3,3){\begin{color}{white}Conv. Deblurring\end{color}}
      \end{overpic}\vspace{.15em}
      
     \begin{overpic}[width=.245\linewidth]{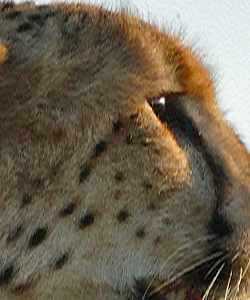}
        \put(3,3){\begin{color}{white}L0-Deblur\end{color}}
      \end{overpic} 
     \begin{overpic}[width=.245\linewidth]{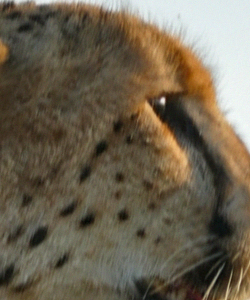}
        \put(3,3){\begin{color}{white}DeblurGANv2-inception\end{color}}
      \end{overpic}      
     \begin{overpic}[width=.245\linewidth]{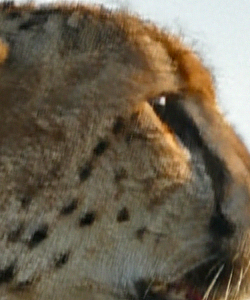}
        \put(3,3){\begin{color}{white}DeblurGANv2-mobilenet\end{color}}
      \end{overpic}     
     \begin{overpic}[width=.245\linewidth]{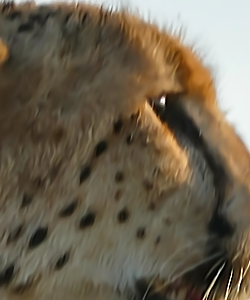}
        \put(3,3){\begin{color}{white}GLAS\end{color}}
      \end{overpic}\vspace{.15em}
      
     \begin{overpic}[width=.245\linewidth]{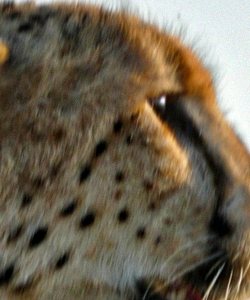}
        \put(3,3){\begin{color}{white}Guided Filter\end{color}}
      \end{overpic}                 
     \begin{overpic}[width=.245\linewidth]{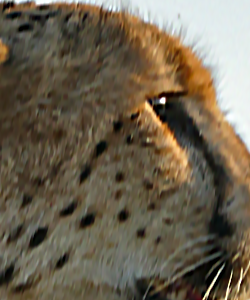}
        \put(3,3){\begin{color}{white}SparseDeblur\end{color}}
      \end{overpic}         
     \begin{overpic}[width=.245\linewidth]{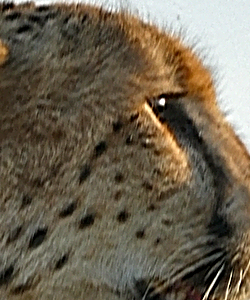}
        \put(3,3){\begin{color}{white}Spectral Irreg\end{color}}
      \end{overpic}   
     \begin{overpic}[width=.245\linewidth]{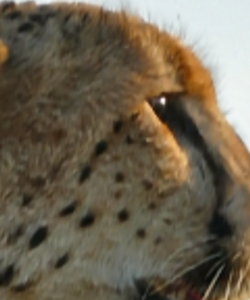}
        \put(3,3){\begin{color}{white}SRN-Deblur\end{color}}
      \end{overpic}\vspace{.15em}
      
     \begin{overpic}[width=.245\linewidth]{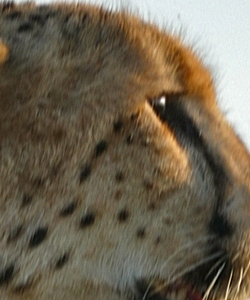}
        \put(3,3){\begin{color}{white}Polyblur-it1\end{color}}
      \end{overpic}         
     \begin{overpic}[width=.245\linewidth]{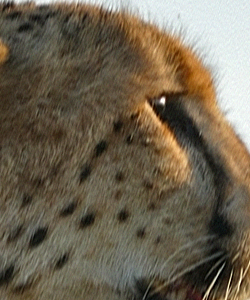}
        \put(3,3){\begin{color}{white}Polyblur-it2\end{color}}
      \end{overpic}   
     \begin{overpic}[width=.245\linewidth]{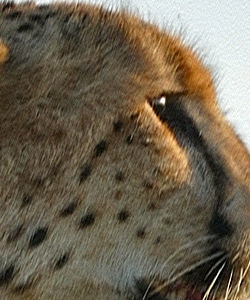}
        \put(3,3){\begin{color}{white}Polyblur-it3\end{color}}
      \end{overpic}   
     \begin{overpic}[width=.245\linewidth]{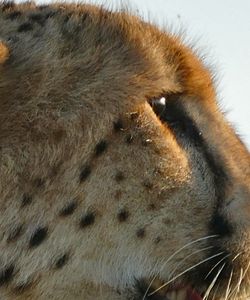}
        \put(3,3){\begin{color}{white}Reference\end{color}}
      \end{overpic}         
    \caption{Example of comparison on one image from the DIV2K validation dataset with synthetic blur.}
    \label{fig:synt_results5}
\end{figure*}

\begin{figure*}
\footnotesize
    \centering
     \begin{tikzpicture}
      \node[anchor=north west,inner sep=0] (image) at (0,0) {\includegraphics[clip, trim=25 0 25 0, width=.245\linewidth]{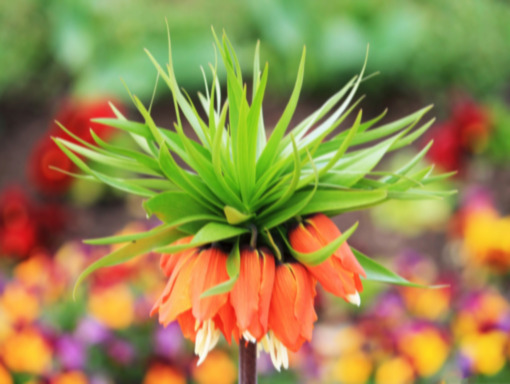}};
      \begin{scope}[x={(image.north east)},y={(image.south west)}]
        \draw[yellow1, very thick] (0.5272, 0.5488) rectangle (0.6902, 0.7116);
         \node[] at (0.24,0.93) {\begin{color}{white}Low-resolution\end{color}};
       \end{scope}
     \end{tikzpicture}    
      \begin{overpic}[width=.245\linewidth]{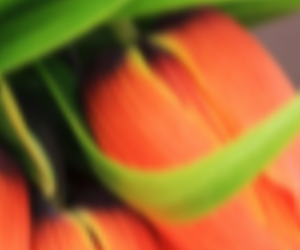}
        \put(3,3){\begin{color}{white}Bicubic $\times 4$\end{color}}
      \end{overpic}
      \begin{overpic}[width=.245\linewidth]{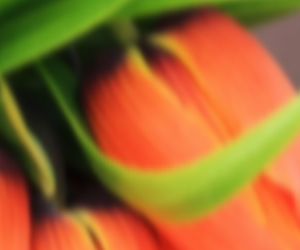}
        \put(3,3){\begin{color}{white}\textsc{edsr}\end{color}}
      \end{overpic}
     \begin{overpic}[width=.245\linewidth]{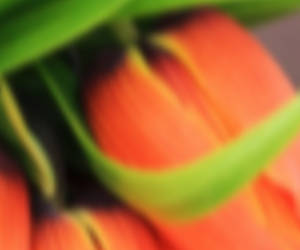}
        \put(3,3){\begin{color}{white}KernelGAN + \textsc{zssr}\end{color}}
      \end{overpic}\vspace{.15em} 

     \begin{overpic}[width=.245\linewidth]{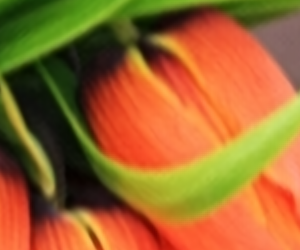}
        \put(3,3){\begin{color}{white}Polyblur-it1 + \textsc{edsr}\end{color}}
      \end{overpic}      
     \begin{overpic}[width=.245\linewidth]{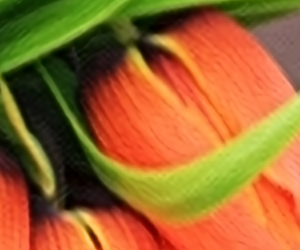}
        \put(3,3){\begin{color}{white}Polyblur-it2 + \textsc{edsr}\end{color}}
      \end{overpic}     
     \begin{overpic}[width=.245\linewidth]{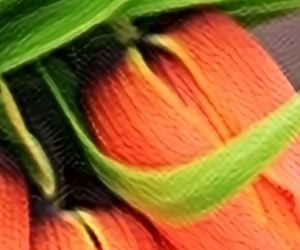}
        \put(3,3){\begin{color}{white}Polyblur-it3 + \textsc{edsr}\end{color}}
      \end{overpic}     
     \begin{overpic}[width=.245\linewidth]{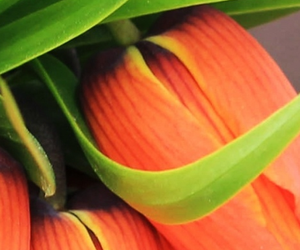}
        \put(3,3){\begin{color}{white}Groundtruth\end{color}}
      \end{overpic}     
      
      \vspace{.25em}
      \begin{tikzpicture}
      \node[anchor=north west,inner sep=0] (image) at (0,0) {\includegraphics[clip, trim=25 0 25 0, width=.245\linewidth]{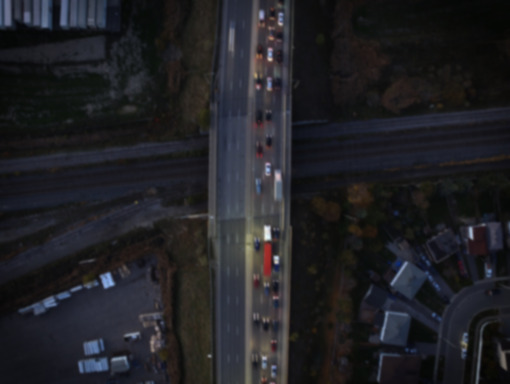}};
      \begin{scope}[x={(image.north east)},y={(image.south west)}]
         \draw[red1, very thick] (0.4712, 0.5618) rectangle (0.6342, 0.7246);
         \node[] at (0.24,0.93) {\begin{color}{white}Low-resolution\end{color}};
       \end{scope}
     \end{tikzpicture}    
      \begin{overpic}[width=.245\linewidth]{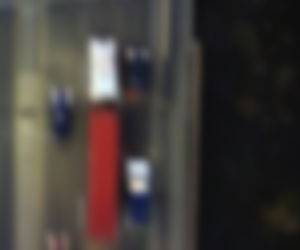}
        \put(3,3){\begin{color}{white}Bicubic $\times 4$\end{color}}
      \end{overpic}
      \begin{overpic}[width=.245\linewidth]{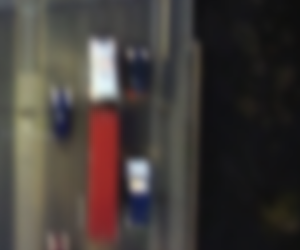}
        \put(3,3){\begin{color}{white}\textsc{edsr}\end{color}}
      \end{overpic}
     \begin{overpic}[width=.245\linewidth]{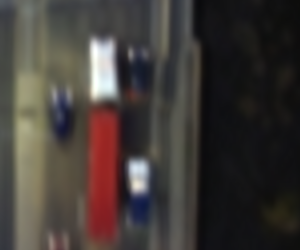}
        \put(3,3){\begin{color}{white}KernelGAN + \textsc{zssr}\end{color}}
      \end{overpic}\vspace{.15em}

     \begin{overpic}[width=.245\linewidth]{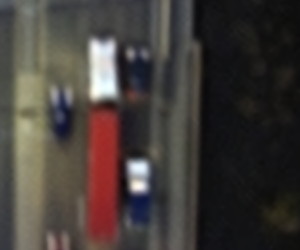}
        \put(3,3){\begin{color}{white}Polyblur-it1 + \textsc{edsr}\end{color}}
      \end{overpic}      
     \begin{overpic}[width=.245\linewidth]{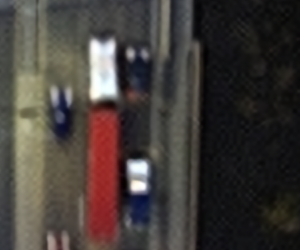}
        \put(3,3){\begin{color}{white}Polyblur-it2 + \textsc{edsr}\end{color}}
      \end{overpic}     
     \begin{overpic}[width=.245\linewidth]{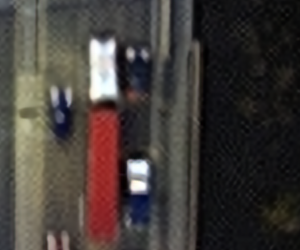}
        \put(3,3){\begin{color}{white}Polyblur-it3 + \textsc{edsr}\end{color}}
      \end{overpic}     
     \begin{overpic}[width=.245\linewidth]{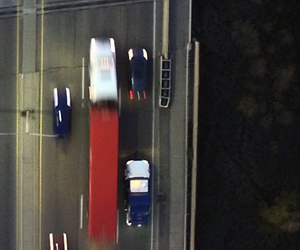}
        \put(3,3){\begin{color}{white}Groundtruth\end{color}}
      \end{overpic}     
      
   \vspace{.25em}
      \begin{tikzpicture}
      \node[anchor=north west,inner sep=0] (image) at (0,0) {\includegraphics[clip, trim=25 0 25 0, width=.245\linewidth]{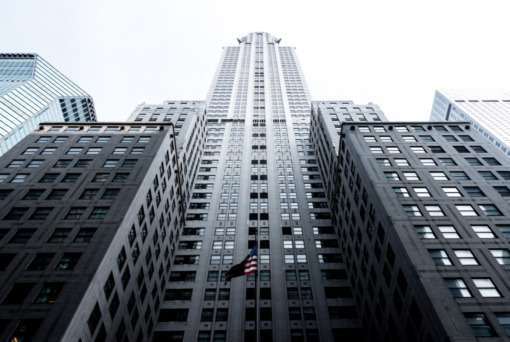}};
      \begin{scope}[x={(image.north east)},y={(image.south west)}]
         \draw[red1, very thick] (0.5272, 0.5548) rectangle (0.6902, 0.7376);
         \node[] at (0.24,0.93) {\begin{color}{white}Low-resolution\end{color}};
       \end{scope}
     \end{tikzpicture}   
      \begin{overpic}[width=.245\linewidth]{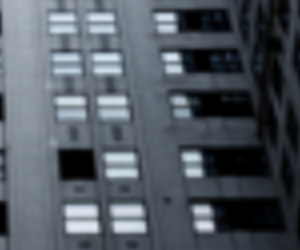}
        \put(3,3){\begin{color}{white}Bicubic $\times 4$\end{color}}
      \end{overpic}
      \begin{overpic}[width=.245\linewidth]{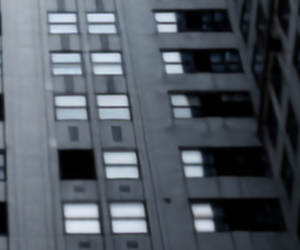}
        \put(3,3){\begin{color}{white}\textsc{edsr}\end{color}}
      \end{overpic}
     \begin{overpic}[width=.245\linewidth]{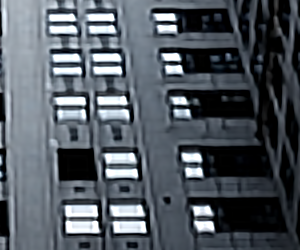}
        \put(3,3){\begin{color}{white}KernelGAN + \textsc{zssr}\end{color}}
      \end{overpic}\vspace{.15em} 

     \begin{overpic}[width=.245\linewidth]{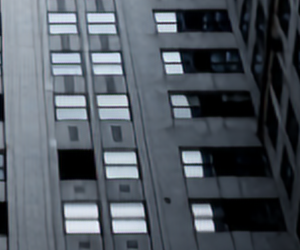}
        \put(3,3){\begin{color}{white}Polyblur-it1 + \textsc{edsr}\end{color}}
      \end{overpic}      
     \begin{overpic}[width=.245\linewidth]{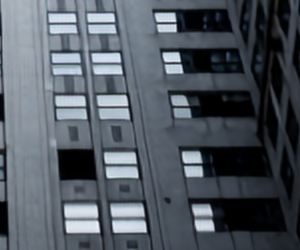}
        \put(3,3){\begin{color}{white}Polyblur-it2 + \textsc{edsr}\end{color}}
      \end{overpic}     
     \begin{overpic}[width=.245\linewidth]{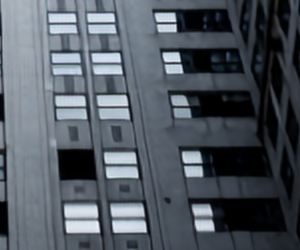}
        \put(3,3){\begin{color}{white}Polyblur-it3 + \textsc{edsr}\end{color}}
      \end{overpic} 
     \begin{overpic}[width=.245\linewidth]{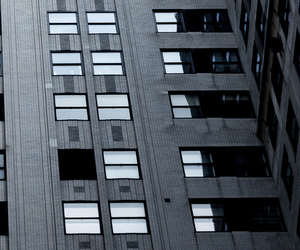}
        \put(3,3){\begin{color}{white}Groundtruth\end{color}}
      \end{overpic}

    \caption{DIV2KRK~\cite{bell2019blind} $4\times$ upscaling with \emph{unknown} kernel and out-of-the-shelf super-resolution model.}
    \label{fig:sr4x_results1}
\end{figure*}

\begin{figure*}
\footnotesize
    \centering
      \begin{tikzpicture}
      \node[anchor=north west,inner sep=0] (image) at (0,0) {\includegraphics[clip, trim=50 0 50 0, width=.245\linewidth]{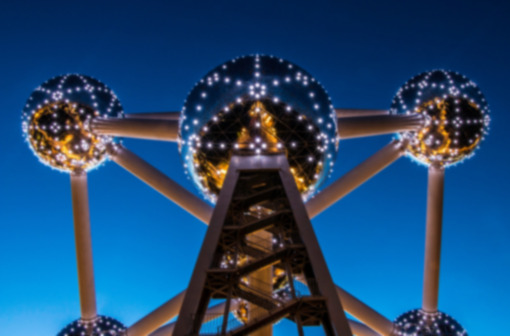}};
      \begin{scope}[x={(image.north east)},y={(image.south west)}]
         \draw[red1, very thick] (0.5305, 0.5558) rectangle (0.7134, 0.7418);
         \node[] at (0.24,0.93) {\begin{color}{white}Low-resolution\end{color}};
       \end{scope}
     \end{tikzpicture}   
      \begin{overpic}[width=.245\linewidth]{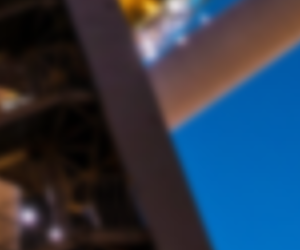}
        \put(3,3){\begin{color}{white}Bicubic $\times 4$\end{color}}
      \end{overpic}
      \begin{overpic}[width=.245\linewidth]{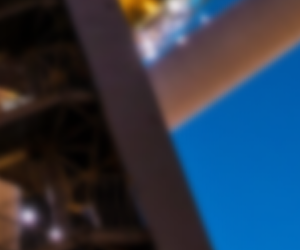}
        \put(3,3){\begin{color}{white}\textsc{edsr}\end{color}}
      \end{overpic}
     \begin{overpic}[width=.245\linewidth]{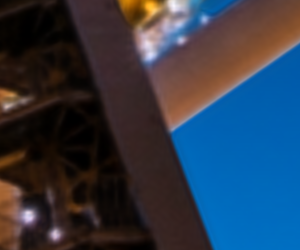}
        \put(3,3){\begin{color}{white}KernelGAN + \textsc{zssr}\end{color}}
      \end{overpic}\vspace{.15em} 

     \begin{overpic}[width=.245\linewidth]{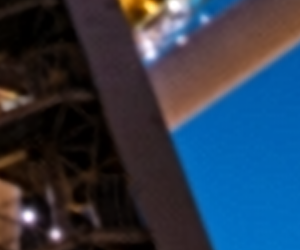}
        \put(3,3){\begin{color}{white}Polyblur-it1 + \textsc{edsr}\end{color}}
      \end{overpic}      
     \begin{overpic}[width=.245\linewidth]{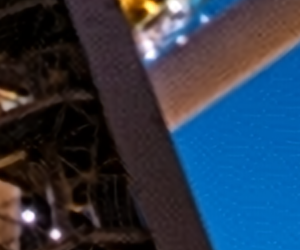}
        \put(3,3){\begin{color}{white}Polyblur-it2 + \textsc{edsr}\end{color}}
      \end{overpic}     
     \begin{overpic}[width=.245\linewidth]{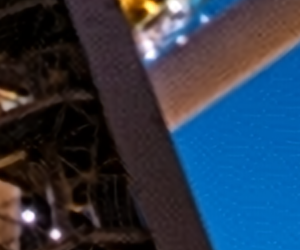}
        \put(3,3){\begin{color}{white}Polyblur-it3 + \textsc{edsr}\end{color}}
      \end{overpic}     
     \begin{overpic}[width=.245\linewidth]{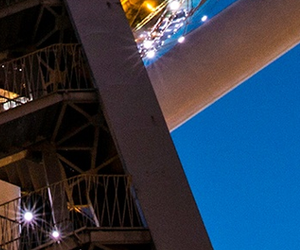}
        \put(3,3){\begin{color}{white}Groundtruth\end{color}}
      \end{overpic}     
      
      \vspace{.25em}
      \begin{tikzpicture}
      \node[anchor=north west,inner sep=0] (image) at (0,0) {\includegraphics[clip, trim=0 32 0 50, width=.245\linewidth]{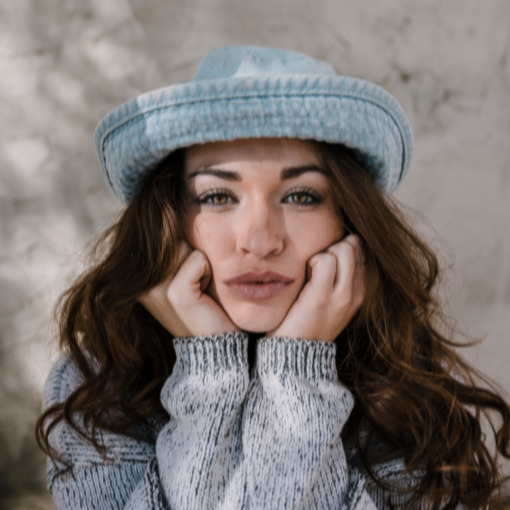}};
      \begin{scope}[x={(image.north east)},y={(image.south west)}]
         \draw[red1, very thick] (0.4755, 0.4644) rectangle (0.6225, 0.6104);
         \node[] at (0.24,0.93) {\begin{color}{white}Low-resolution\end{color}};
       \end{scope}
     \end{tikzpicture}    
      \begin{overpic}[width=.245\linewidth]{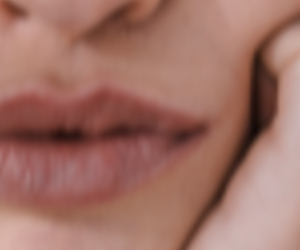}
        \put(3,3){\begin{color}{white}Bicubic $\times 4$\end{color}}
      \end{overpic}
      \begin{overpic}[width=.245\linewidth]{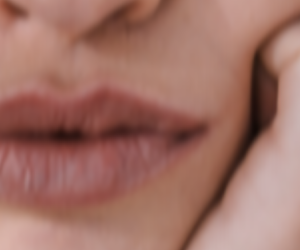}
        \put(3,3){\begin{color}{white}\textsc{edsr}\end{color}}
      \end{overpic}
     \begin{overpic}[width=.245\linewidth]{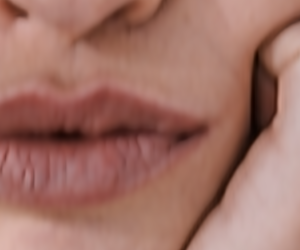}
        \put(3,3){\begin{color}{white}KernelGAN + \textsc{zssr}\end{color}}
      \end{overpic}\vspace{.15em} 

     \begin{overpic}[width=.245\linewidth]{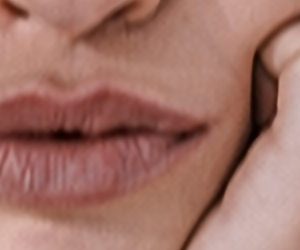}
        \put(3,3){\begin{color}{white}Polyblur-it1 + \textsc{edsr}\end{color}}
      \end{overpic}      
     \begin{overpic}[width=.245\linewidth]{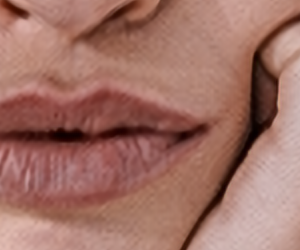}
        \put(3,3){\begin{color}{white}Polyblur-it2 + \textsc{edsr}\end{color}}
      \end{overpic}     
     \begin{overpic}[width=.245\linewidth]{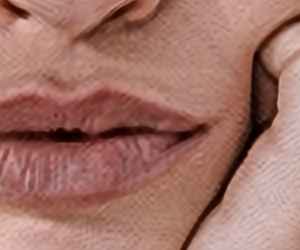}
        \put(3,3){\begin{color}{white}Polyblur-it3 + \textsc{edsr}\end{color}}
      \end{overpic}     
     \begin{overpic}[width=.245\linewidth]{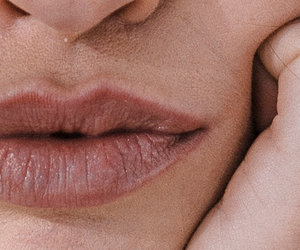}
        \put(3,3){\begin{color}{white}Groundtruth\end{color}}
      \end{overpic}     
      
   \vspace{.25em}
      \begin{tikzpicture}
      \node[anchor=north west,inner sep=0] (image) at (0,0) {\includegraphics[clip, trim=50 0 50 0, width=.245\linewidth]{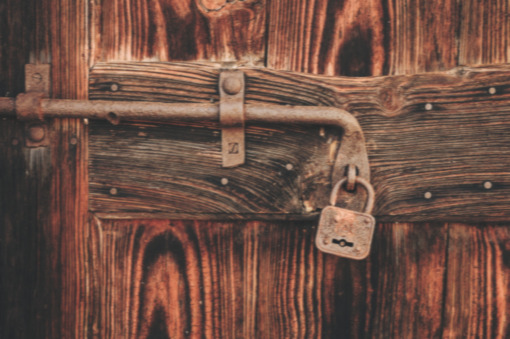}};
      \begin{scope}[x={(image.north east)},y={(image.south west)}]
         \draw[red1, very thick] (0.5305, 0.5553) rectangle (0.7134, 0.7397);
         \node[] at (0.24,0.93) {\begin{color}{white}Low-resolution\end{color}};
       \end{scope}
     \end{tikzpicture}     
      \begin{overpic}[width=.245\linewidth]{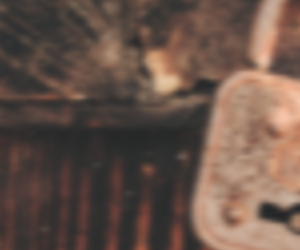}
        \put(3,3){\begin{color}{white}Bicubic $\times 4$\end{color}}
      \end{overpic}
      \begin{overpic}[width=.245\linewidth]{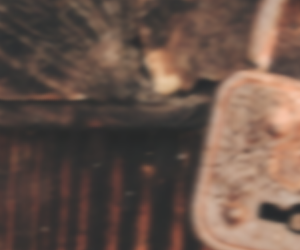}
        \put(3,3){\begin{color}{white}\textsc{edsr}\end{color}}
      \end{overpic}
     \begin{overpic}[width=.245\linewidth]{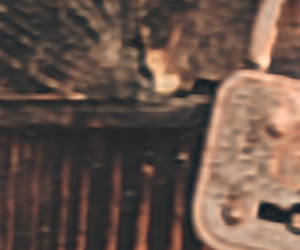}
        \put(3,3){\begin{color}{white}KernelGAN + \textsc{zssr}\end{color}}
      \end{overpic}\vspace{.15em} 

     \begin{overpic}[width=.245\linewidth]{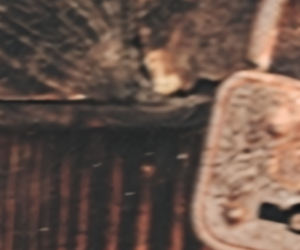}
        \put(3,3){\begin{color}{white}Polyblur-it1 + \textsc{edsr}\end{color}}
      \end{overpic}      
     \begin{overpic}[width=.245\linewidth]{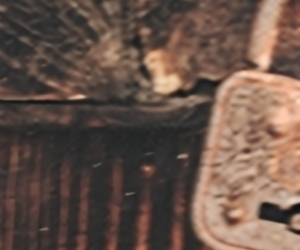}
        \put(3,3){\begin{color}{white}Polyblur-it2 + \textsc{edsr}\end{color}}
      \end{overpic}     
     \begin{overpic}[width=.245\linewidth]{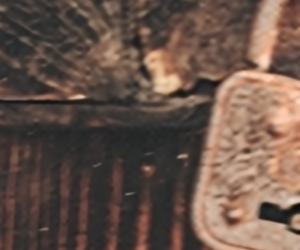}
        \put(3,3){\begin{color}{white}Polyblur-it3 + \textsc{edsr}\end{color}}
      \end{overpic}     
     \begin{overpic}[width=.245\linewidth]{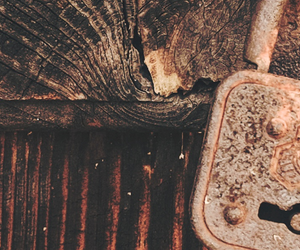}
        \put(3,3){\begin{color}{white}Groundtruth\end{color}}
      \end{overpic}

    \caption{DIV2KRK~\cite{bell2019blind} $4\times$ upscaling with \emph{unknown} kernel and out-of-the-shelf super-resolution model.}
    \label{fig:sr4x_results2}
\end{figure*}

\vspace{.5em}
\noindent \textbf{Deblurring before super-resolution.} Figures~\ref{fig:sr4x_results1} and \ref{fig:sr4x_results2} present additional results to Section~6 on deblurring before super-resolution. We apply Polyblur as a pre-step before using an off-the-shelf deep network for doing $4\times$ image upscaling. We trained from scratch an EDSR~\cite{lim2017enhanced} network with 32 layers and 64 filters using DIV2K training dataset. Polyblur produces the best quantitative and qualitative results. The evaluation is done on the DIV2KRK dataset introduced in~\cite{bell2019blind}.

\definecolor{red1}{rgb}{0.8431    0.1882    0.1529}
\definecolor{orange1}{rgb}{0.9882    0.5529    0.3490}
\definecolor{yellow1}{rgb}{0.9961    0.8784    0.5451}
\definecolor{blue1}{rgb}{0.3000    0.3000    0.8490}
\definecolor{green2}{rgb}{0.2    0.8    0.3}
\definecolor{green1}{rgb}{0.1020    0.5961    0.3137}
\definecolor{red2}{rgb}{0.8431    0.1882    0.1529}

\setlength\fboxsep{0pt}
\setlength\fboxrule{1.4pt}

\begin{figure*}
\footnotesize
    \centering
     \begin{tikzpicture}
      \node[anchor=north west,inner sep=0] (image) at (0,0) {\includegraphics[clip, trim=50 320 0 300, width=.39\linewidth]{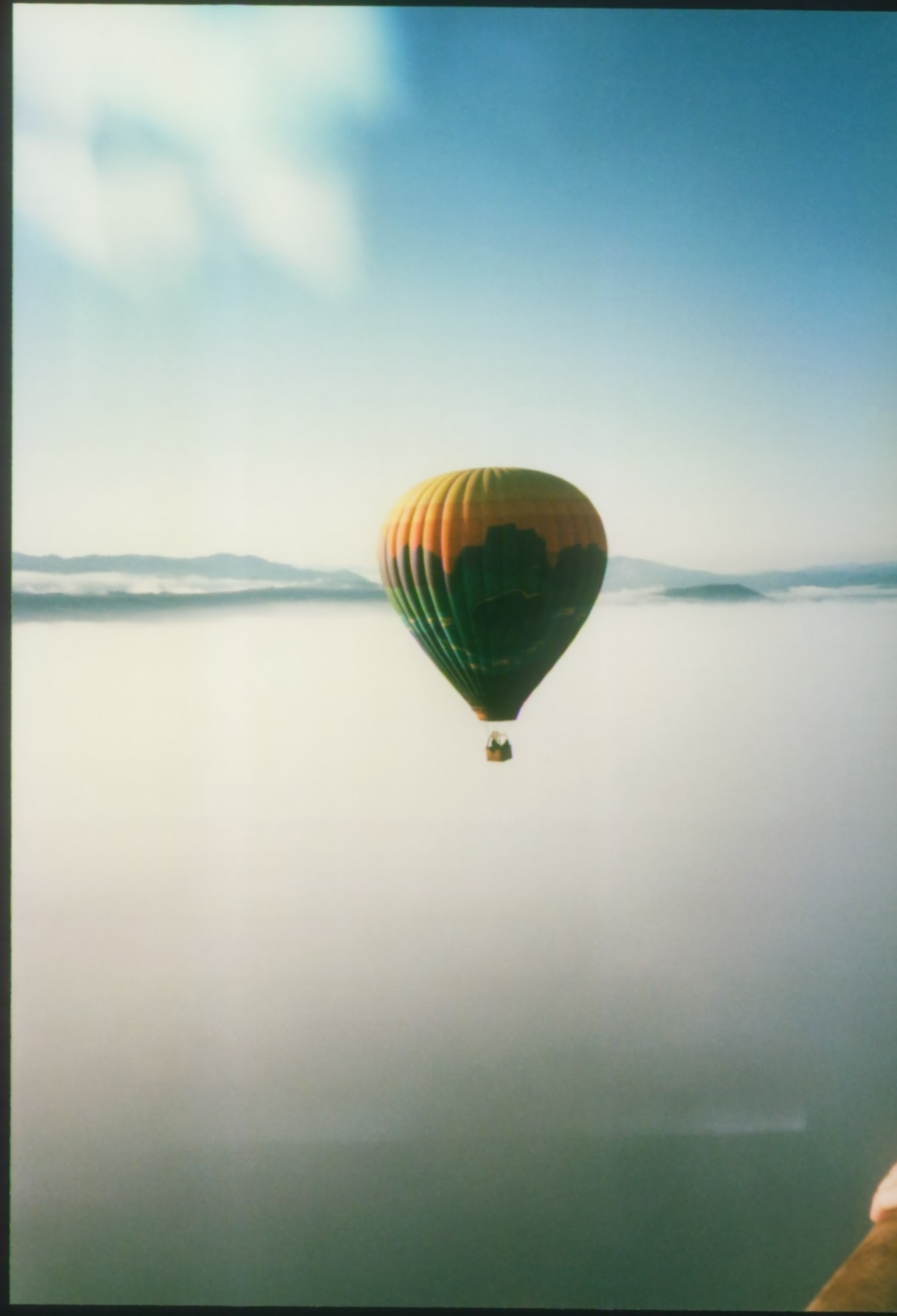}};
      \begin{scope}[x={(image.north east)},y={(image.south west)}]
         \draw[red1, very thick] (0.4358, 0.2942) rectangle (0.6070, 0.6372);
         \node[] at (0.07,0.96) {\begin{color}{arsenic}Input\end{color}};
       \end{scope}
     \end{tikzpicture}     
      \begin{overpic}[width=.195\linewidth]{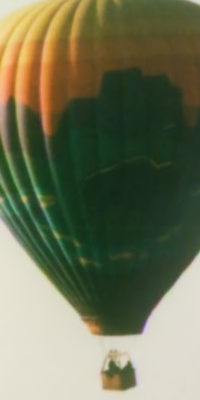}
        \put(3,3){\begin{color}{arsenic}Input\end{color}}
      \end{overpic}
     \begin{overpic}[width=.195\linewidth]{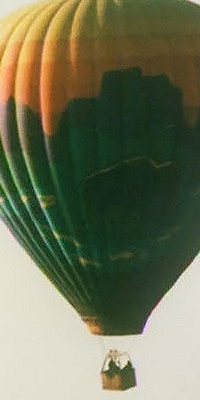}
        \put(3,3){\begin{color}{arsenic}Conv. Deblurring\end{color}}
      \end{overpic}
     \begin{overpic}[width=.195\linewidth]{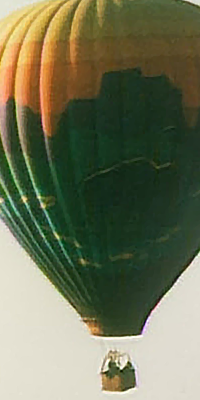}
        \put(3,3){\begin{color}{arsenic}L0-Deblur\end{color}}
      \end{overpic}\vspace{.15em}
      
     \begin{overpic}[width=.195\linewidth]{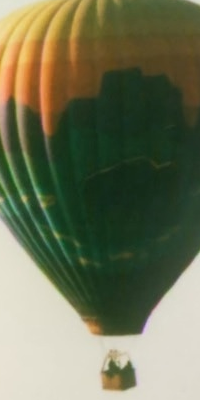}
        \put(3,3){\begin{color}{arsenic}DeblurGANv2-inception\end{color}}
      \end{overpic}      
     \begin{overpic}[width=.195\linewidth]{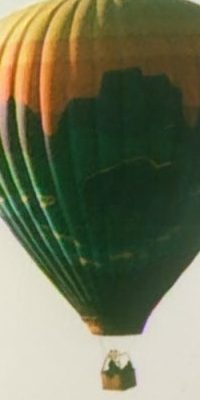}
        \put(3,3){\begin{color}{arsenic}DeblurGANv2-mobilenet\end{color}}
      \end{overpic}     
     \begin{overpic}[width=.195\linewidth]{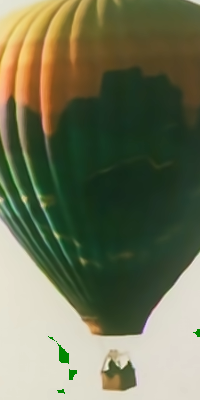}
        \put(3,3){\begin{color}{arsenic}GLAS\end{color}}
      \end{overpic}     
     \begin{overpic}[width=.195\linewidth]{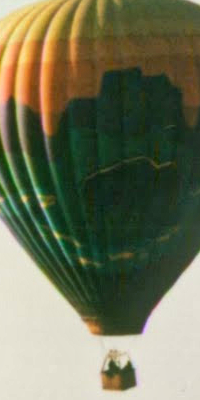}
        \put(3,3){\begin{color}{arsenic}Guided Filter\end{color}}
      \end{overpic}                 
     \begin{overpic}[width=.195\linewidth]{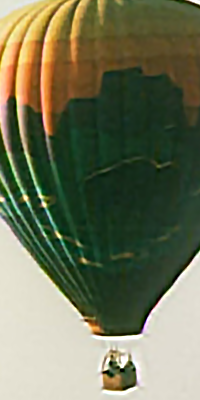}
        \put(3,3){\begin{color}{arsenic}SparseDeblur\end{color}}
      \end{overpic}\vspace{.15em}
      
     \begin{overpic}[width=.195\linewidth]{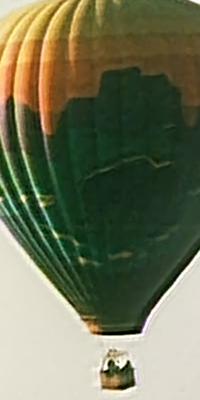}
        \put(3,3){\begin{color}{arsenic}Spectral Irreg\end{color}}
      \end{overpic}   
     \begin{overpic}[width=.195\linewidth]{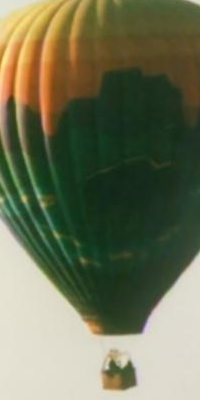}
        \put(3,3){\begin{color}{arsenic}SRN-Deblur\end{color}}
      \end{overpic}  
     \begin{overpic}[width=.195\linewidth]{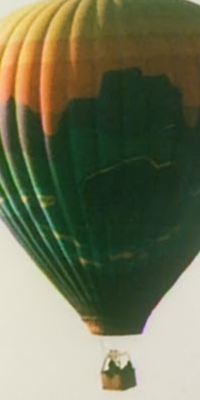}
        \put(3,3){\begin{color}{arsenic}Polyblur-it1\end{color}}
      \end{overpic}         
     \begin{overpic}[width=.195\linewidth]{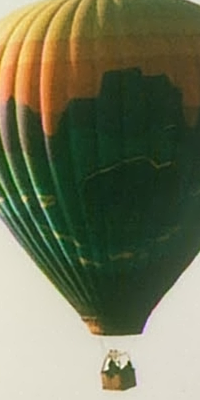}
        \put(3,3){\begin{color}{arsenic}Polyblur-it2\end{color}}
      \end{overpic}   
     \begin{overpic}[width=.195\linewidth]{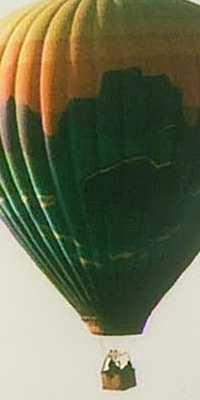}
        \put(3,3){\begin{color}{arsenic}Polyblur-it3\end{color}}
      \end{overpic}   
    \caption{Example of removing mild blur \emph{in the wild}.}
    \label{fig:wild1}
\end{figure*}

\begin{figure*}
\footnotesize
    \centering
     \begin{tikzpicture}
      \node[anchor=north west,inner sep=0] (image) at (0,0) {\includegraphics[clip, trim=80 0 500 20, width=.39\linewidth]{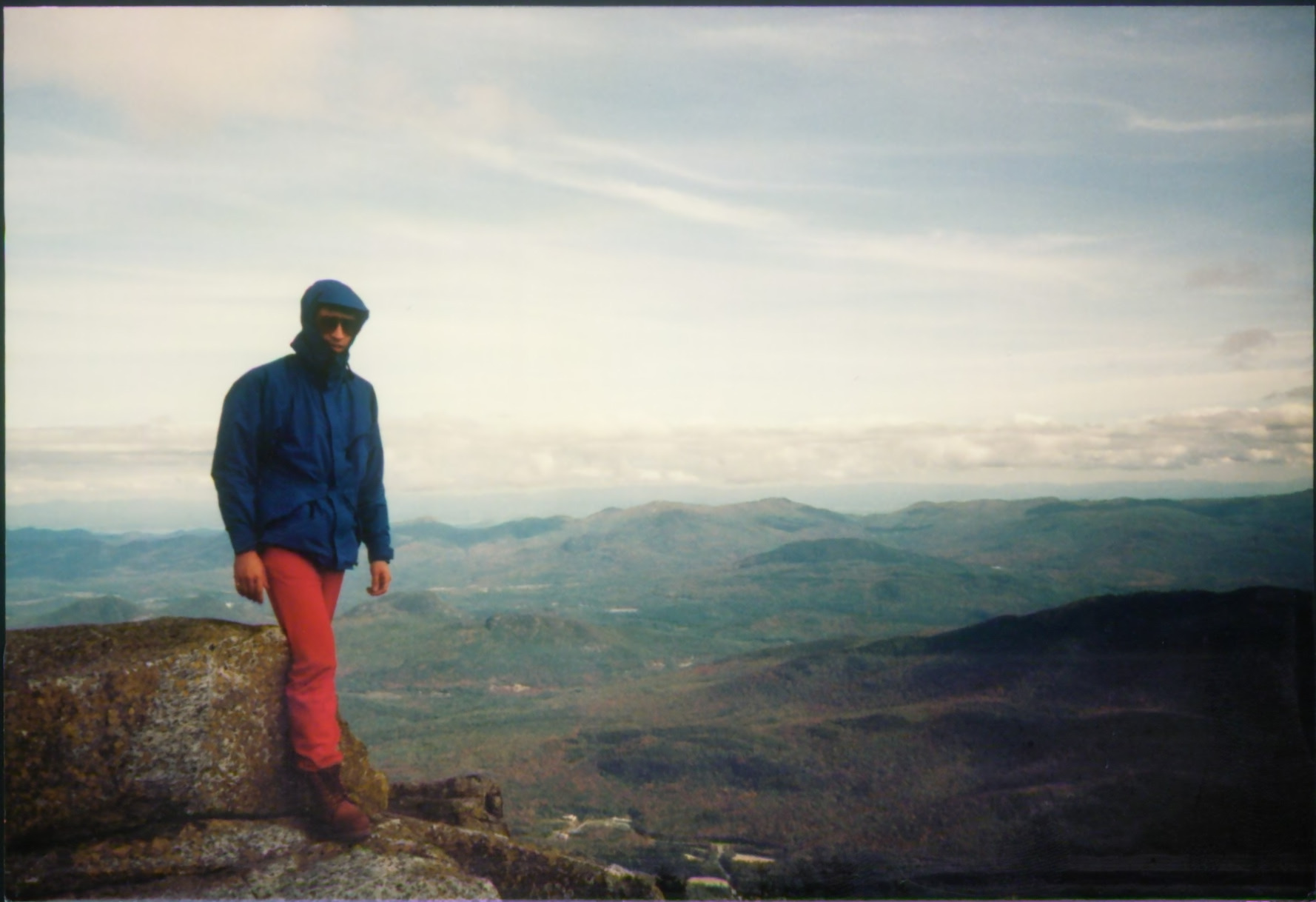}};
      \begin{scope}[x={(image.north east)},y={(image.south west)}]
         \draw[red1, very thick]  (0.1937, 0.2888) rectangle (0.4023, 0.7062);
         \node[] at (0.07,0.96) {\begin{color}{white}Input\end{color}};
       \end{scope}
     \end{tikzpicture}     
      \begin{overpic}[width=.19\linewidth]{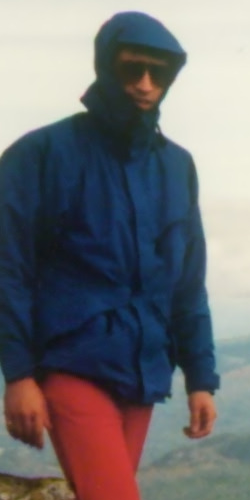}
        \put(3,3){\begin{color}{white}Input\end{color}}
      \end{overpic}
     \begin{overpic}[width=.19\linewidth]{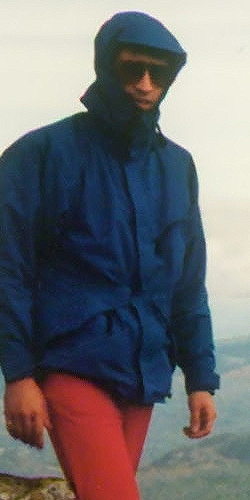}
        \put(3,3){\begin{color}{white}Conv. Deblurring\end{color}}
      \end{overpic}
     \begin{overpic}[width=.19\linewidth]{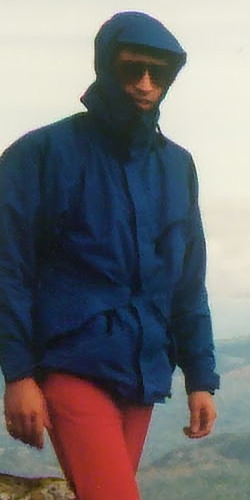}
        \put(3,3){\begin{color}{white}L0-Deblur\end{color}}
      \end{overpic}\vspace{.15em}
      
     \begin{overpic}[width=.19\linewidth]{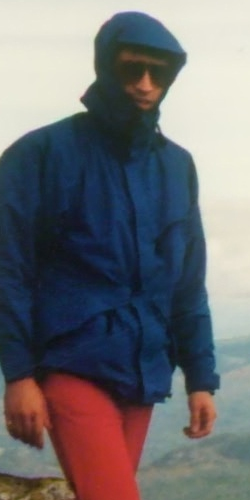}
        \put(3,3){\begin{color}{white}DeblurGANv2-inception\end{color}}
      \end{overpic}      
     \begin{overpic}[width=.19\linewidth]{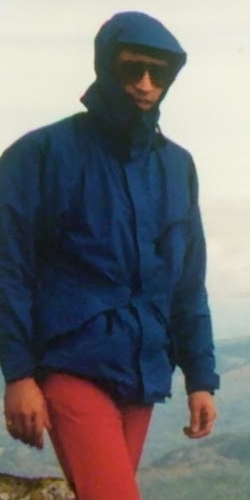}
        \put(3,3){\begin{color}{white}DeblurGANv2-mobilenet\end{color}}
      \end{overpic}     
     \begin{overpic}[width=.19\linewidth]{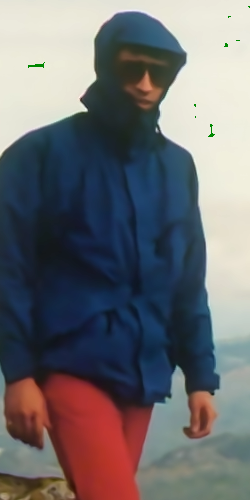}
        \put(3,3){\begin{color}{white}GLAS\end{color}}
      \end{overpic}     
     \begin{overpic}[width=.19\linewidth]{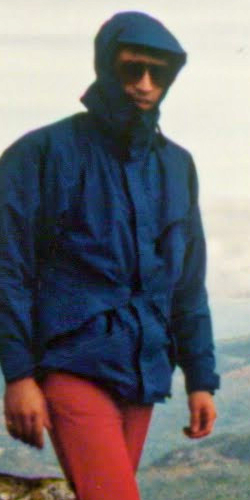}
        \put(3,3){\begin{color}{white}Guided Filter\end{color}}
      \end{overpic}                 
     \begin{overpic}[width=.19\linewidth]{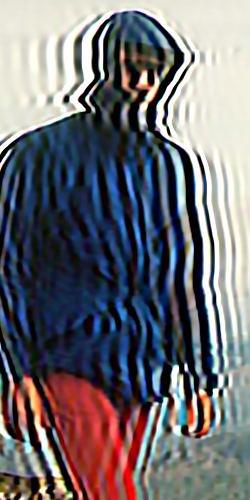}
        \put(3,3){\begin{color}{white}SparseDeblur\end{color}}
      \end{overpic}\vspace{.15em}
      
     \begin{overpic}[width=.19\linewidth]{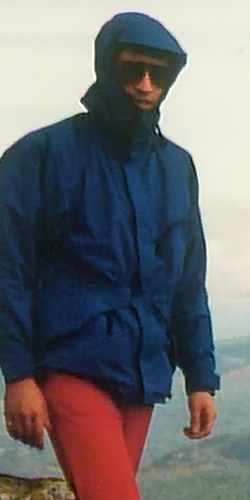}
        \put(3,3){\begin{color}{white}Spectral Irreg\end{color}}
      \end{overpic}   
     \begin{overpic}[width=.19\linewidth]{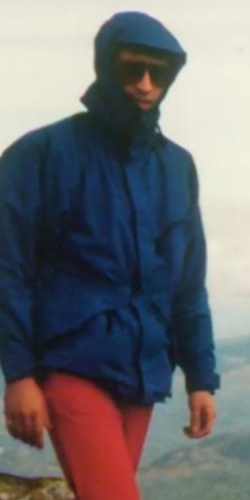}
        \put(3,3){\begin{color}{white}SRN-Deblur\end{color}}
      \end{overpic}  
     \begin{overpic}[width=.19\linewidth]{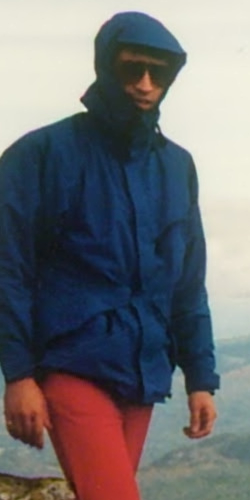}
        \put(3,3){\begin{color}{white}Polyblur-it1\end{color}}
      \end{overpic}         
     \begin{overpic}[width=.19\linewidth]{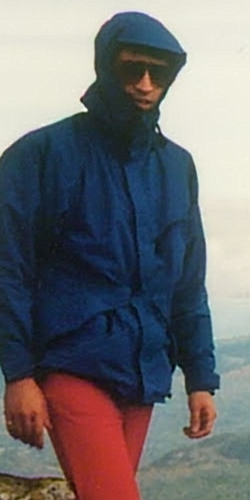}
        \put(3,3){\begin{color}{white}Polyblur-it2\end{color}}
      \end{overpic}   
     \begin{overpic}[width=.19\linewidth]{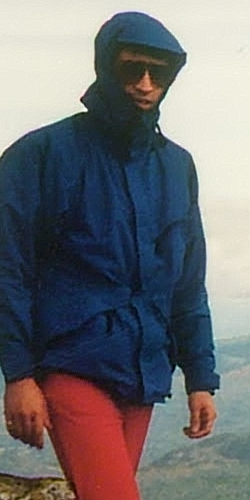}
        \put(3,3){\begin{color}{white}Polyblur-it3\end{color}}
      \end{overpic}   
    \caption{Example of removing mild blur \emph{in the wild}.}
    \label{fig:wild2}
\end{figure*}

\begin{figure*}
\footnotesize
    \centering
     \begin{tikzpicture}
      \node[anchor=north west,inner sep=0] (image) at (0,0) {\includegraphics[clip, trim=0 0 0 340, width=.39\linewidth]{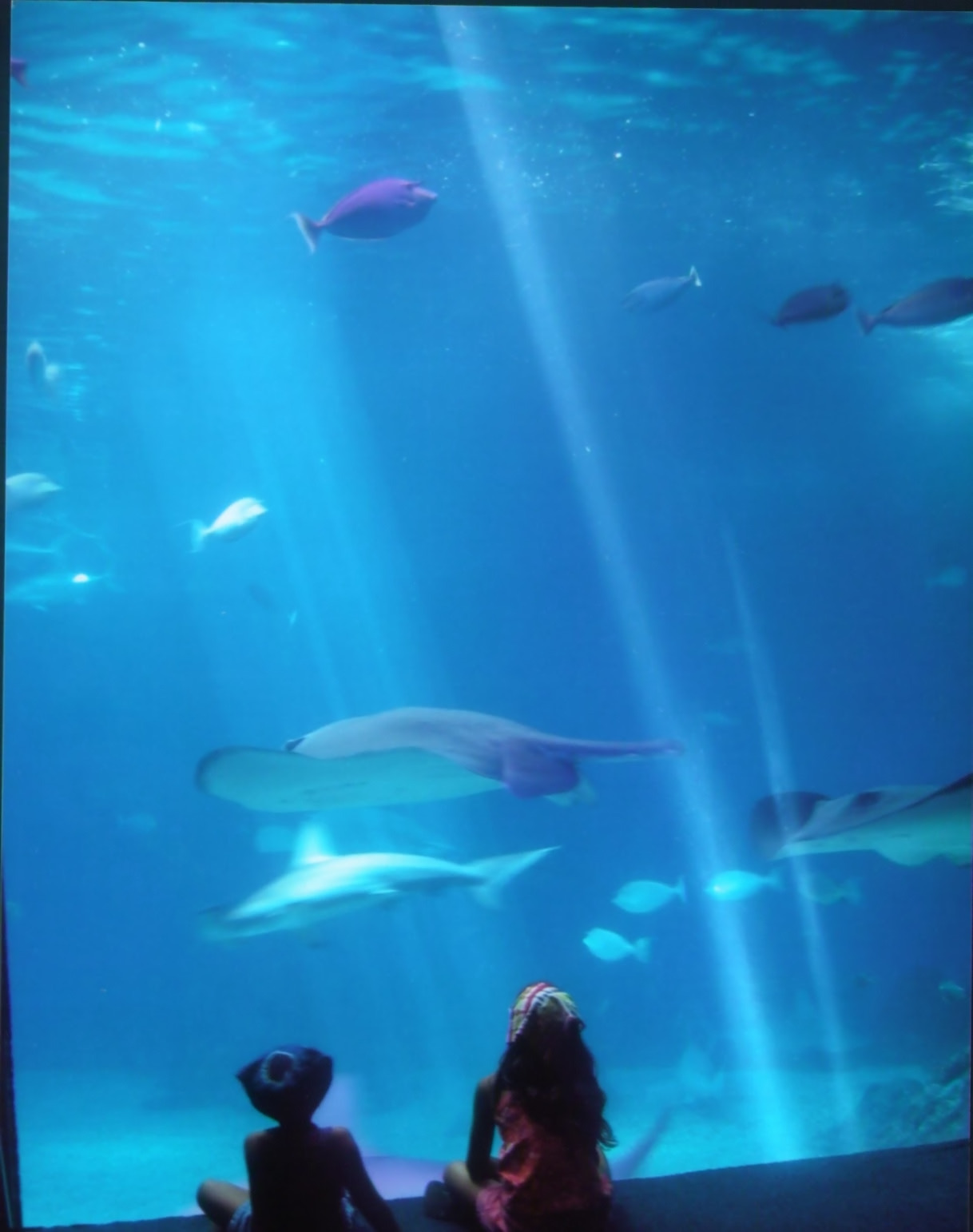}};
      \begin{scope}[x={(image.north east)},y={(image.south west)}]
         \draw[red1, very thick] (0.4392, 0.4859) rectangle (0.6445, 0.9018);
         \node[] at (0.07,0.96) {\begin{color}{white}Input\end{color}};
       \end{scope}
     \end{tikzpicture}  
      \begin{overpic}[width=.19\linewidth]{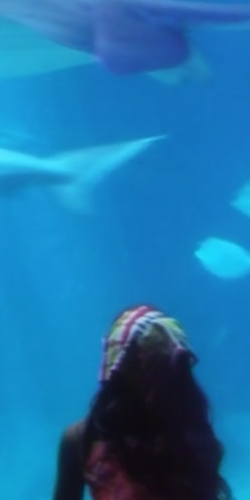}
        \put(3,3){\begin{color}{white}Input\end{color}}
      \end{overpic}
     \begin{overpic}[width=.19\linewidth]{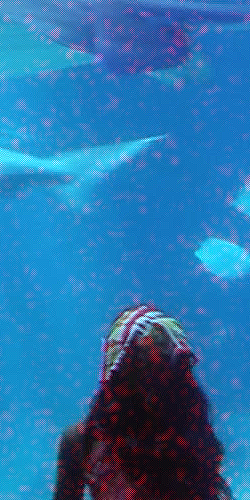}
        \put(3,3){\begin{color}{white}Conv. Deblurring\end{color}}
      \end{overpic}
     \begin{overpic}[width=.19\linewidth]{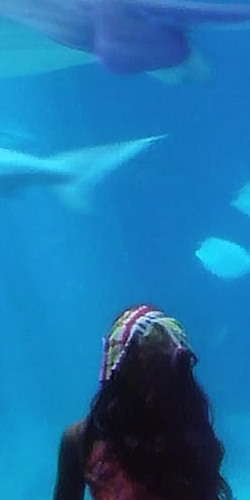}
        \put(3,3){\begin{color}{white}L0-Deblur\end{color}}
      \end{overpic}\vspace{.15em}
      
     \begin{overpic}[width=.19\linewidth]{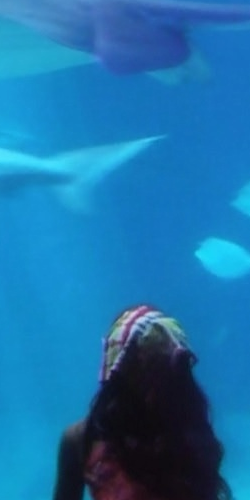}
        \put(3,3){\begin{color}{white}DeblurGANv2-inception\end{color}}
      \end{overpic}      
     \begin{overpic}[width=.19\linewidth]{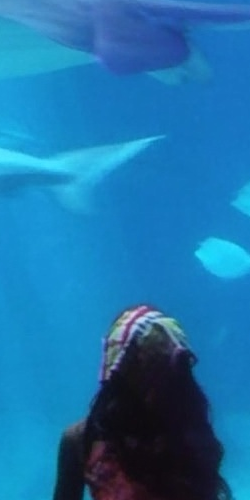}
        \put(3,3){\begin{color}{white}DeblurGANv2-mobilenet\end{color}}
      \end{overpic}     
     \begin{overpic}[width=.19\linewidth]{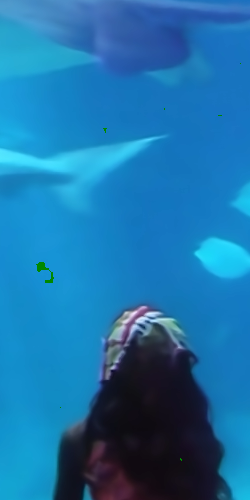}
        \put(3,3){\begin{color}{white}GLAS\end{color}}
      \end{overpic}     
     \begin{overpic}[width=.19\linewidth]{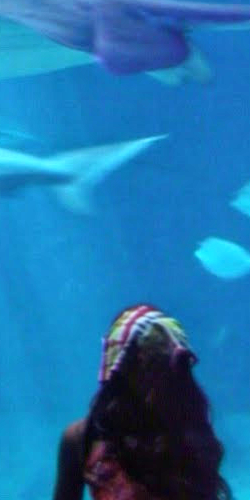}
        \put(3,3){\begin{color}{white}Guided Filter\end{color}}
      \end{overpic}                 
     \begin{overpic}[width=.19\linewidth]{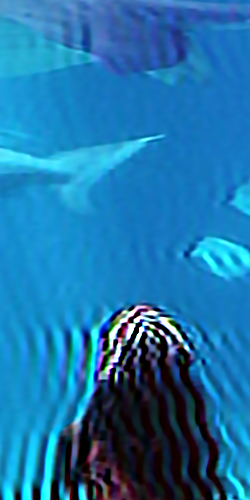}
        \put(3,3){\begin{color}{white}SparseDeblur\end{color}}
      \end{overpic}\vspace{.15em}
      
     \begin{overpic}[width=.19\linewidth]{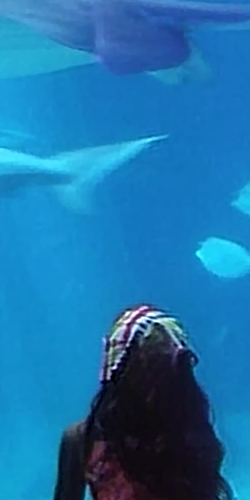}
        \put(3,3){\begin{color}{white}Spectral Irreg\end{color}}
      \end{overpic}   
     \begin{overpic}[width=.19\linewidth]{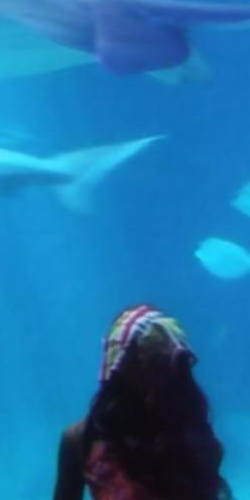}
        \put(3,3){\begin{color}{white}SRN-Deblur\end{color}}
      \end{overpic}  
     \begin{overpic}[width=.19\linewidth]{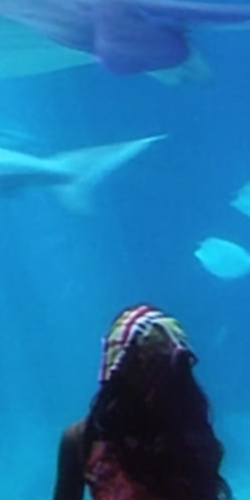}
        \put(3,3){\begin{color}{white}Polyblur-it1\end{color}}
      \end{overpic}         
     \begin{overpic}[width=.19\linewidth]{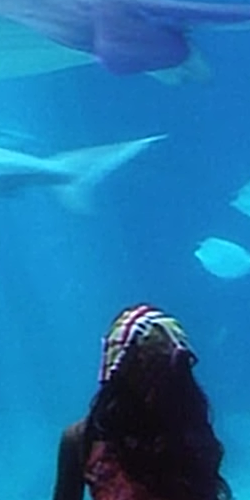}
        \put(3,3){\begin{color}{white}Polyblur-it2\end{color}}
      \end{overpic}   
     \begin{overpic}[width=.19\linewidth]{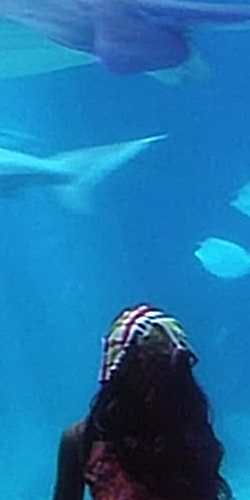}
        \put(3,3){\begin{color}{white}Polyblur-it3\end{color}}
      \end{overpic}   
    \caption{Example of removing mild blur \emph{in the wild}.}
    \label{fig:wild3}
\end{figure*}

\begin{figure*}
\footnotesize
    \centering
     \begin{tikzpicture}
      \node[anchor=north west,inner sep=0] (image) at (0,0) {\includegraphics[clip, trim=800 0 600 150, width=.39\linewidth]{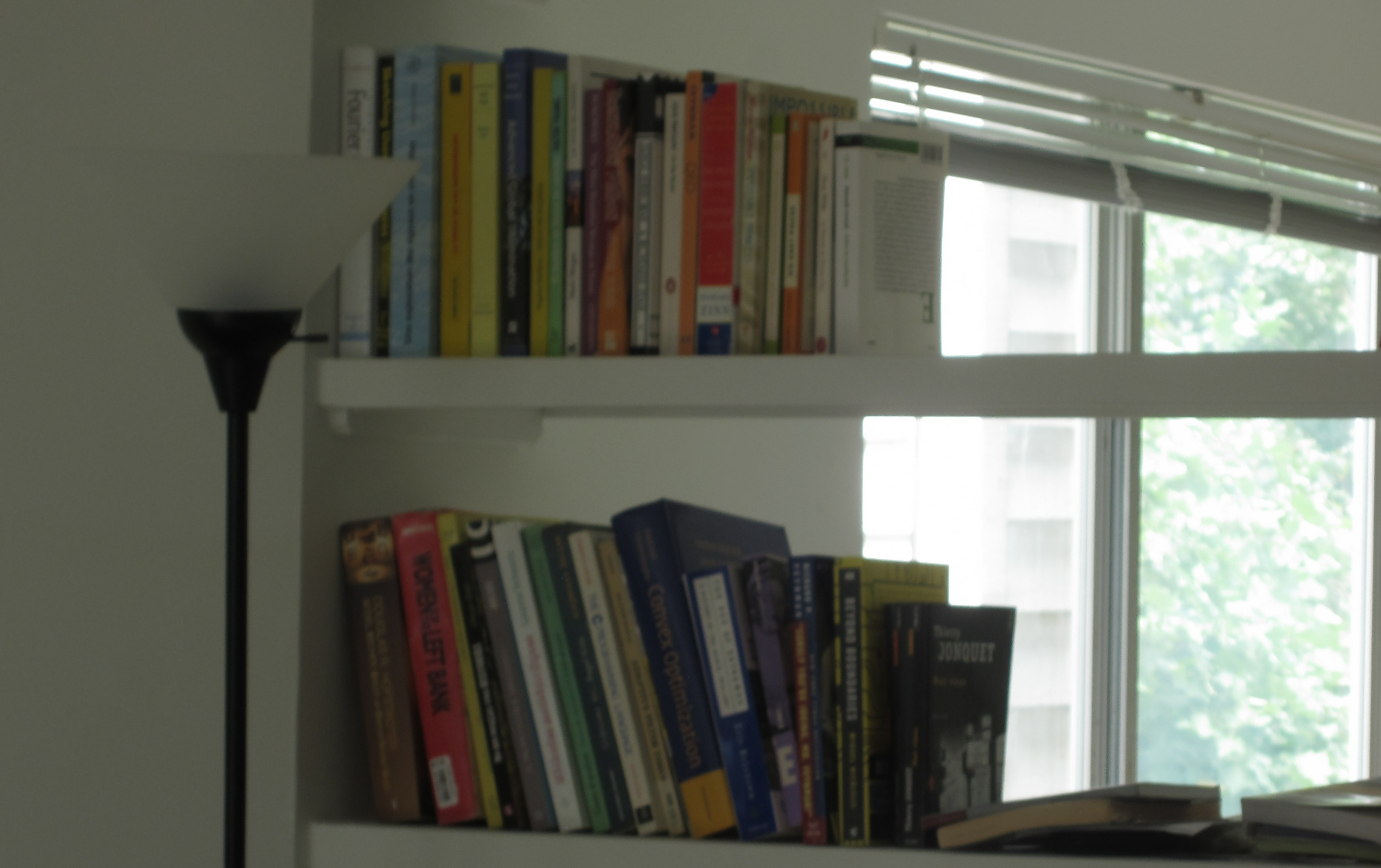}};
      \begin{scope}[x={(image.north east)},y={(image.south west)}]
         \draw[red1, very thick](0.6787, 0.6815) rectangle (0.8215, 0.9541);
         \node[] at (0.07,0.96) {\begin{color}{white}Input\end{color}};
       \end{scope}
     \end{tikzpicture}  
      \begin{overpic}[width=.19\linewidth]{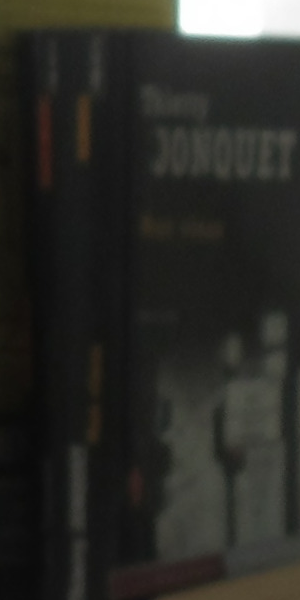}
        \put(3,3){\begin{color}{white}Input\end{color}}
      \end{overpic}
     \begin{overpic}[width=.19\linewidth]{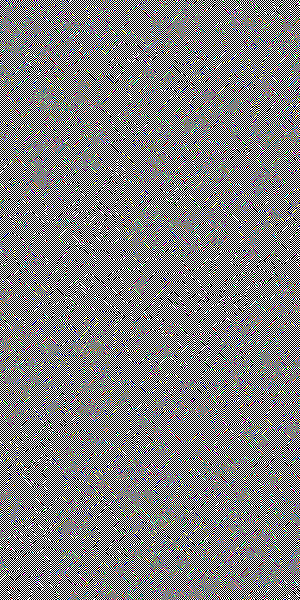}
        \put(3,3){\begin{color}{white}Conv. Deblurring\end{color}}
      \end{overpic}
     \begin{overpic}[width=.19\linewidth]{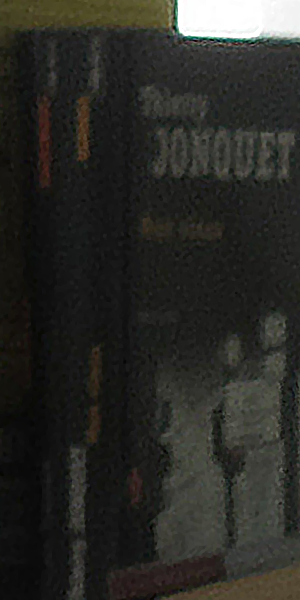}
        \put(3,3){\begin{color}{white}L0-Deblur\end{color}}
      \end{overpic}\vspace{.15em}
      
     \begin{overpic}[width=.19\linewidth]{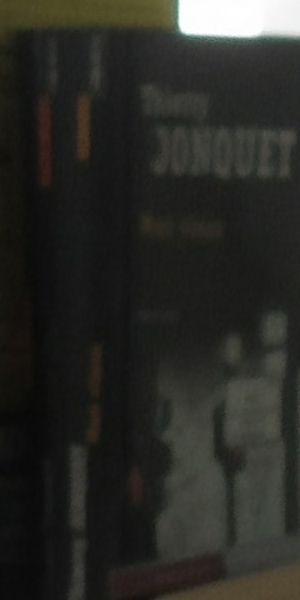}
        \put(3,3){\begin{color}{white}DeblurGANv2-inception\end{color}}
      \end{overpic}      
     \begin{overpic}[width=.19\linewidth]{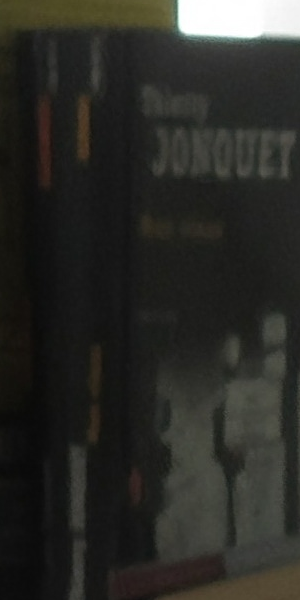}
        \put(3,3){\begin{color}{white}DeblurGANv2-mobilenet\end{color}}
      \end{overpic}     
     \begin{overpic}[width=.19\linewidth]{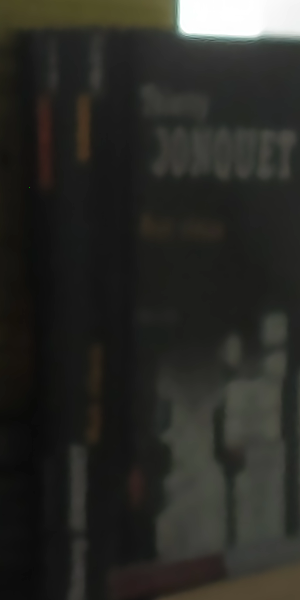}
        \put(3,3){\begin{color}{white}GLAS\end{color}}
      \end{overpic}     
     \begin{overpic}[width=.19\linewidth]{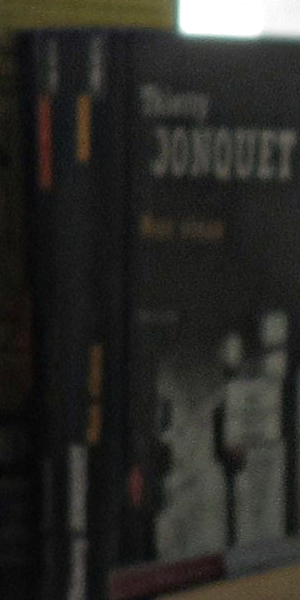}
        \put(3,3){\begin{color}{white}Guided Filter\end{color}}
      \end{overpic}                 
     \begin{overpic}[width=.19\linewidth]{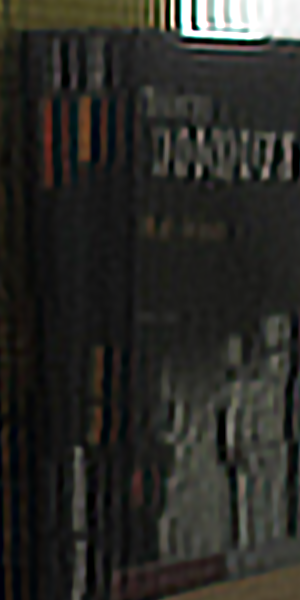}
        \put(3,3){\begin{color}{white}SparseDeblur\end{color}}
      \end{overpic}\vspace{.15em}
      
     \begin{overpic}[width=.19\linewidth]{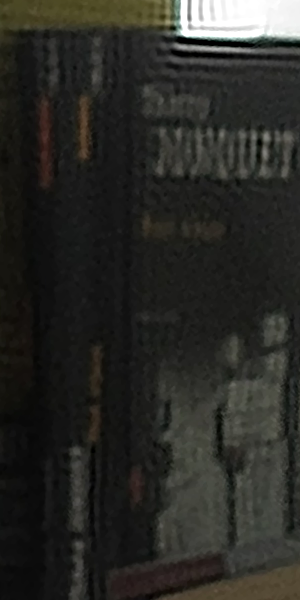}
        \put(3,3){\begin{color}{white}Spectral Irreg\end{color}}
      \end{overpic}   
     \begin{overpic}[width=.19\linewidth]{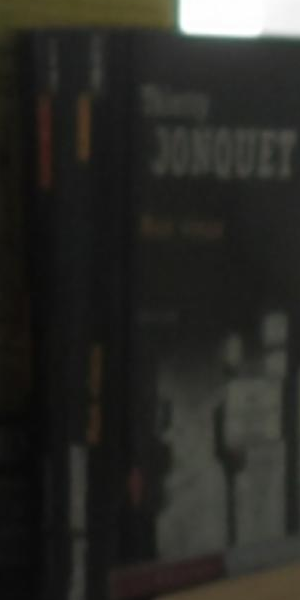}
        \put(3,3){\begin{color}{white}SRN-Deblur\end{color}}
      \end{overpic}  
     \begin{overpic}[width=.19\linewidth]{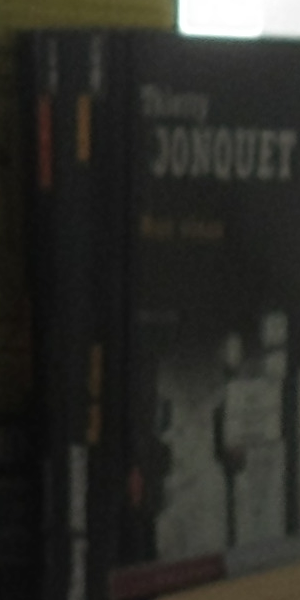}
        \put(3,3){\begin{color}{white}Polyblur-it1\end{color}}
      \end{overpic}         
     \begin{overpic}[width=.19\linewidth]{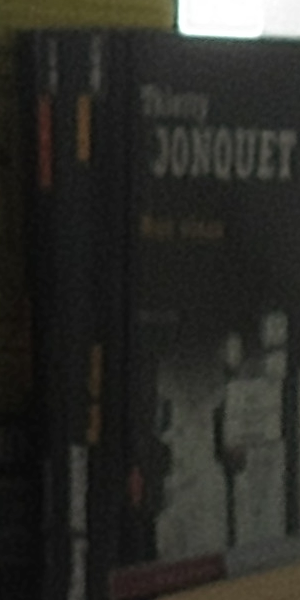}
        \put(3,3){\begin{color}{white}Polyblur-it2\end{color}}
      \end{overpic}   
     \begin{overpic}[width=.19\linewidth]{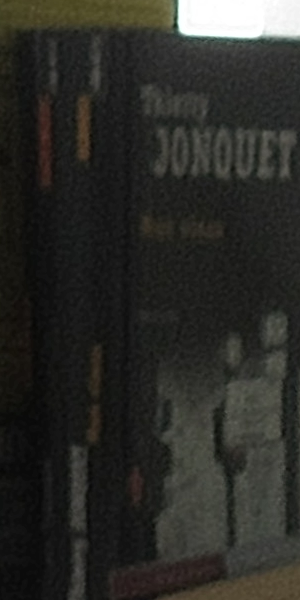}
        \put(3,3){\begin{color}{white}Polyblur-it3\end{color}}
      \end{overpic}   
    \caption{Example of removing mild blur \emph{in the wild}.}
    \label{fig:wild4}
\end{figure*}

\begin{figure*}
\footnotesize
    \centering
    
      \begin{tikzpicture}
      \node[anchor=north west,inner sep=0] (image) at (0,0) {\includegraphics[clip, trim=1400 0 600 400, width=.385\linewidth]{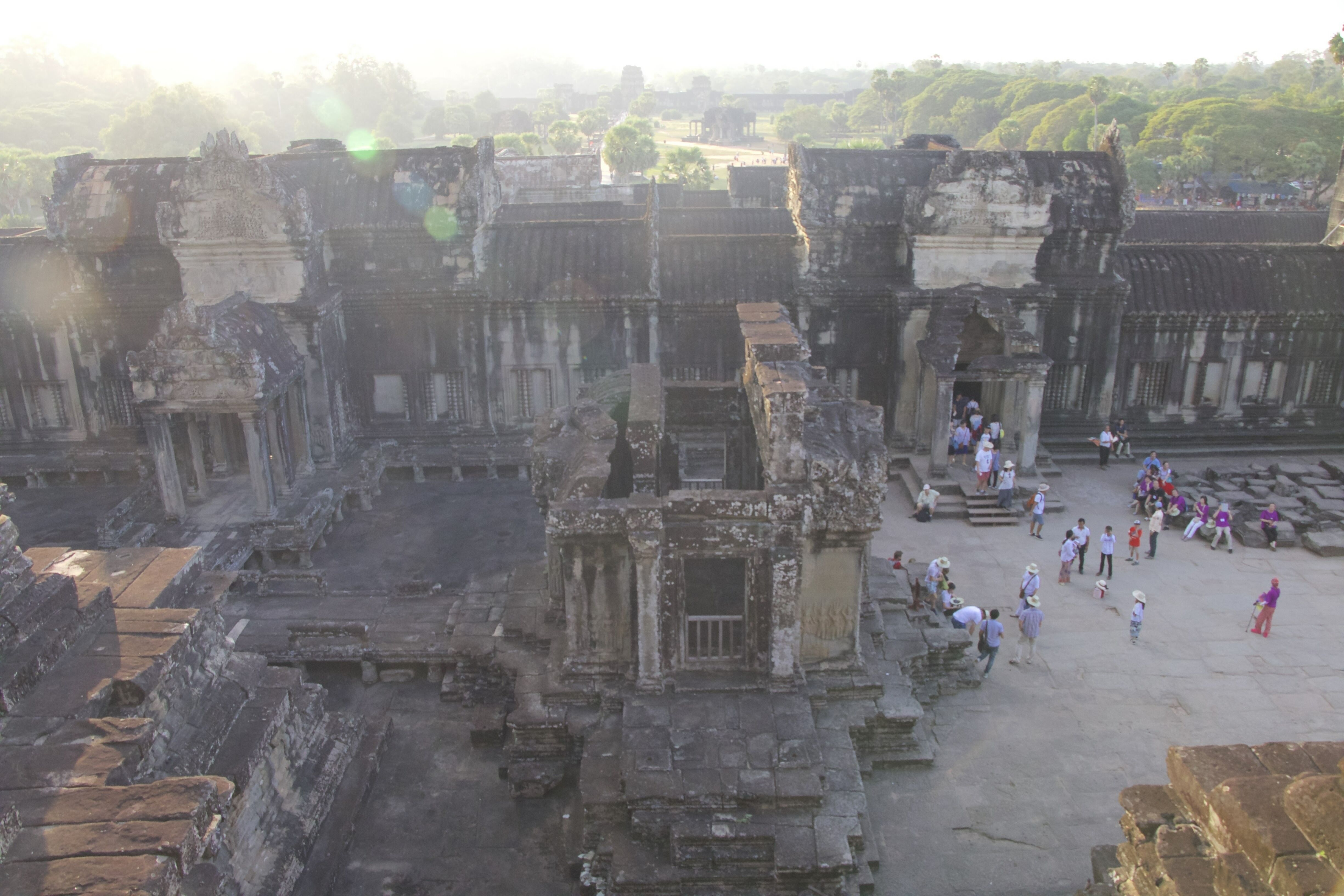}};
      \begin{scope}[x={(image.north east)},y={(image.south west)}]
         \draw[red1, very thick](0.6901, 0.3002) rectangle (0.8109, 0.5445);
         \node[] at (0.07,0.96) {\begin{color}{white}Input\end{color}};
       \end{scope}
     \end{tikzpicture}  %
      \begin{overpic}[width=.19\linewidth]{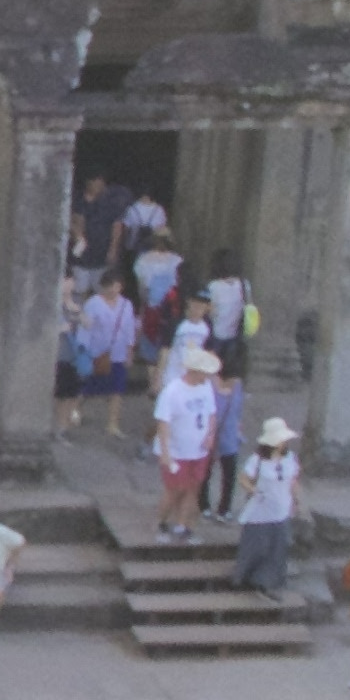}
        \put(3,3){\begin{color}{white}Input\end{color}}
      \end{overpic} 
     \begin{overpic}[width=.19\linewidth]{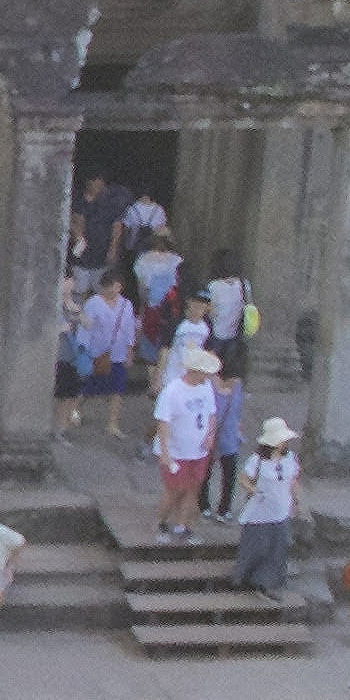}
        \put(3,3){\begin{color}{white}Conv. Deblurring\end{color}}
      \end{overpic} 
     \begin{overpic}[width=.19\linewidth]{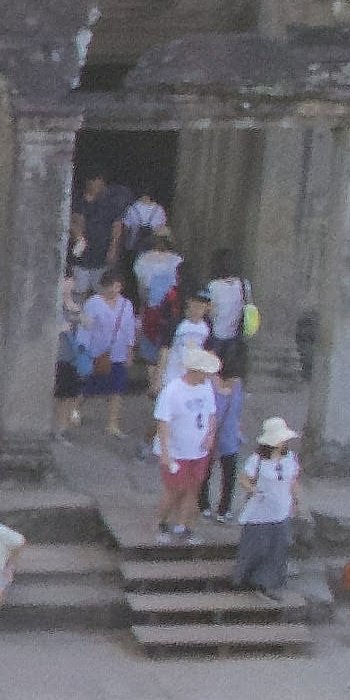}
        \put(3,3){\begin{color}{white}L0-Deblur\end{color}}
      \end{overpic}\vspace{.15em} 
      
     \begin{overpic}[width=.19\linewidth]{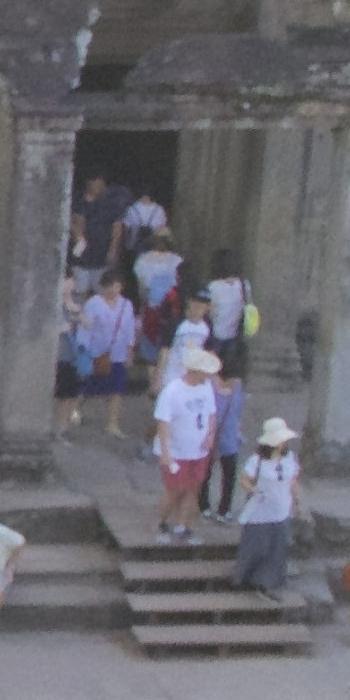}
        \put(3,3){\begin{color}{white}DeblurGANv2-inception\end{color}}
      \end{overpic}      
     \begin{overpic}[width=.19\linewidth]{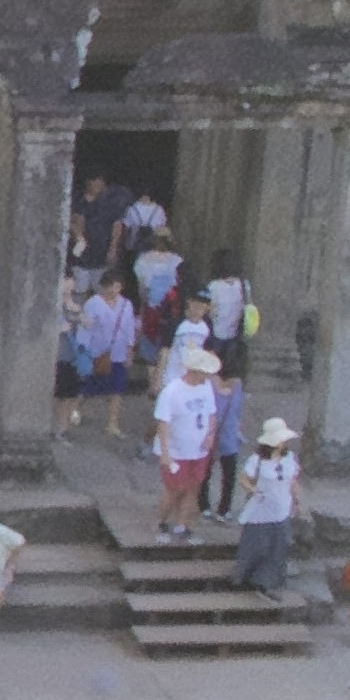}
        \put(3,3){\begin{color}{white}DeblurGANv2-mobilenet\end{color}}
      \end{overpic}     
     \begin{overpic}[width=.19\linewidth]{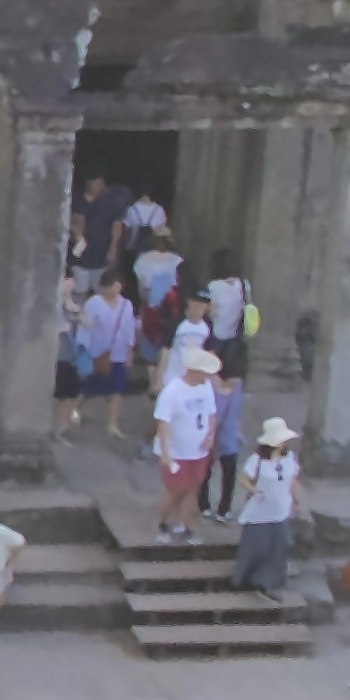}
        \put(3,3){\begin{color}{white}GLAS\end{color}}
      \end{overpic}
     \begin{overpic}[width=.19\linewidth]{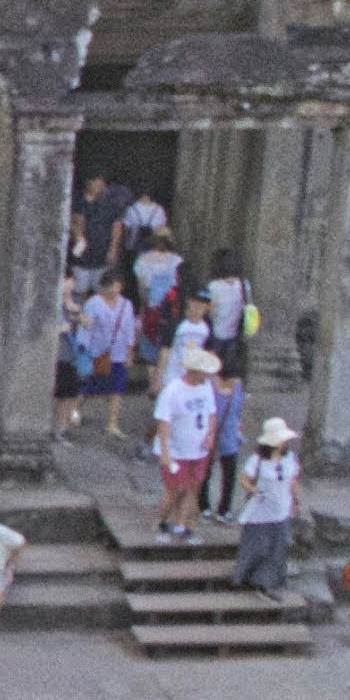}
        \put(3,3){\begin{color}{white}Guided Filter\end{color}}
      \end{overpic}             
     \begin{overpic}[width=.19\linewidth]{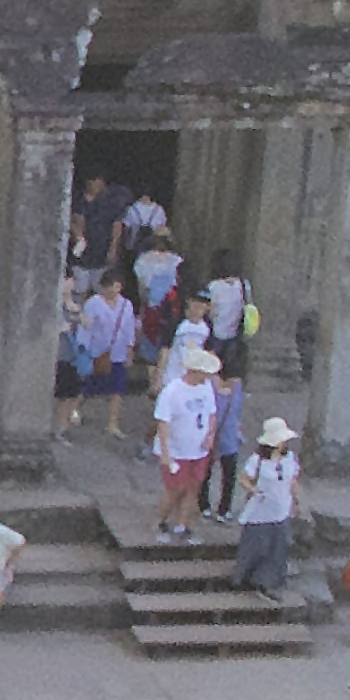}
        \put(3,3){\begin{color}{white}SparseDeblur\end{color}}
      \end{overpic}\vspace{.15em} 
      
     \begin{overpic}[width=.19\linewidth]{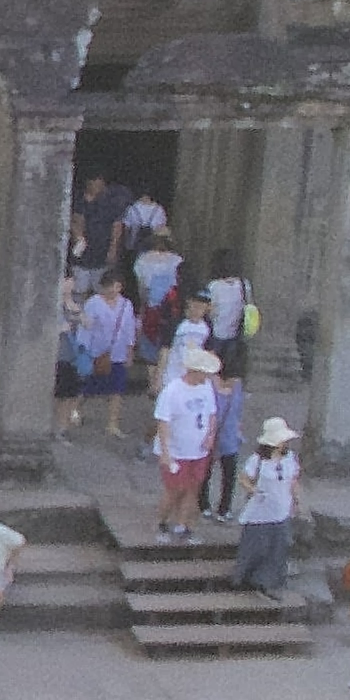}
        \put(3,3){\begin{color}{white}Spectral Irreg\end{color}}
      \end{overpic}  
     \begin{overpic}[width=.19\linewidth]{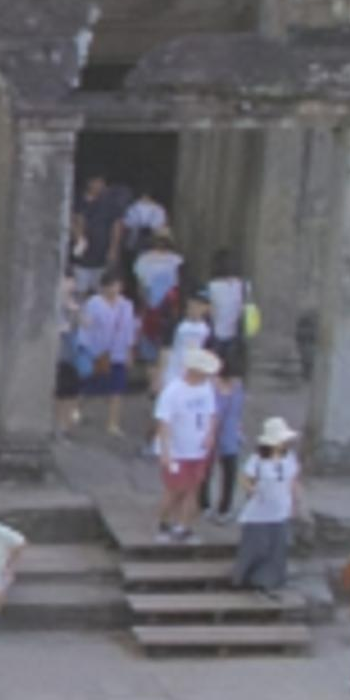}
        \put(3,3){\begin{color}{white}SRN-Deblur\end{color}}
      \end{overpic} 
     \begin{overpic}[width=.19\linewidth]{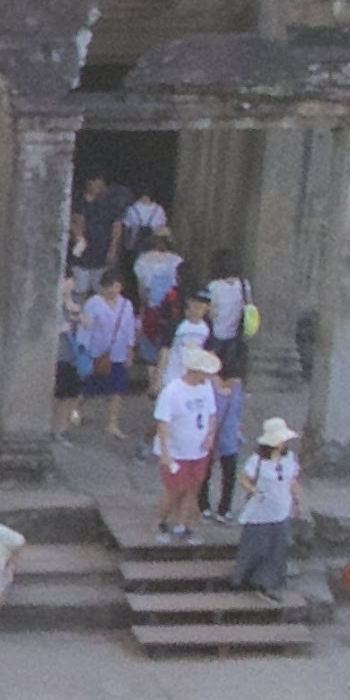}
        \put(3,3){\begin{color}{white}Polyblur-it1\end{color}}
      \end{overpic}        
     \begin{overpic}[width=.19\linewidth]{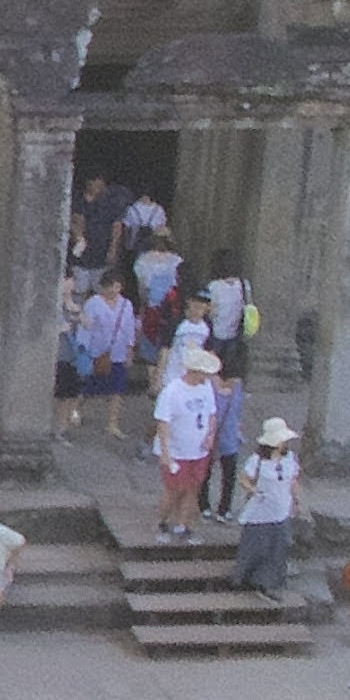}
        \put(3,3){\begin{color}{white}Polyblur-it2\end{color}}
      \end{overpic}  
     \begin{overpic}[width=.19\linewidth]{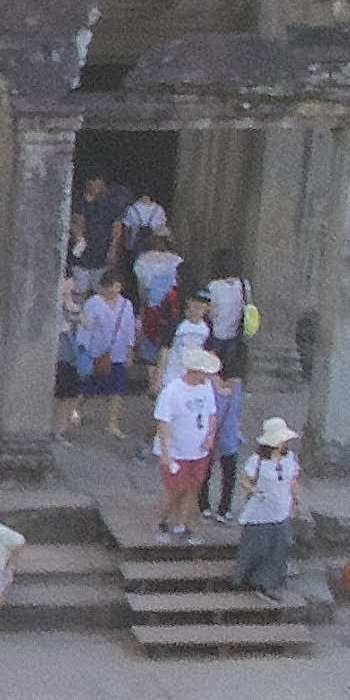}
        \put(3,3){\begin{color}{white}Polyblur-it3\end{color}}
      \end{overpic}   
    \caption{Example of removing mild blur \emph{in the wild}.}
    \label{fig:wild5}
\end{figure*}

\vspace{.5em}
\noindent \textbf{Comparison of images \emph{in the wild}.}  In Figures~\ref{fig:wild1} to \ref{fig:wild5} we present additional results of Polyblur applied to some images in the wild (Section~6). We compare Polyblur (one, two and three iterations) to the following adaptive sharpening or deblurring methods: SRN-Deblur~\cite{tao2018scale}, Sparse Deblurring~\cite{zhang2013multi}, DeblurGANv2 (inception and mobilenet architectures)~\cite{kupyn2019deblurgan}, Spectral Irregularities~\cite{goldstein2012blur}, L0-Deblur~\cite{pan2016l0}, GLAS~\cite{zhu2011restoration}, Guided Filter~\cite{he2012guided}, Convolutional Deblurring~\cite{hosseini2019convolutional}. Polyblur manages to remove mild blur, as the one present in most images, without introducing any new artifacts. 

\ifCLASSOPTIONcaptionsoff
  \newpage
\fi

\bibliographystyle{IEEEtran}
\bibliography{references}

\end{document}

%% file: md_math.tex
\DeclareMathOperator*{\argmin}{arg\,min}

\def\bx{\mathbf{x}}

\newtheorem{thm}{Theorem}
\newtheorem{lemma}[thm]{Lemma}

\newtheorem{assumption}{Assumption}